\documentclass{article}
\usepackage{ijcai24}

\usepackage{times}
\usepackage{soul}
\usepackage{url}
\usepackage[hidelinks]{hyperref}
\usepackage[utf8]{inputenc}
\usepackage[small]{caption}
\usepackage{graphicx}
\usepackage{amsmath}
\usepackage{amsthm}
\usepackage{booktabs}
\usepackage{algorithm}
\usepackage{algorithmic}
\usepackage[switch]{lineno}
\usepackage{tikz}
\usetikzlibrary{hobby}
\usetikzlibrary{shapes}
\usepackage{stmaryrd}
\usetikzlibrary{positioning}
\usepackage{multirow}
\usepackage{pgfplots}
\usepackage{algorithm}
\usepackage{algorithmic}
\usepackage{booktabs}
\usepackage{amsmath} 
\usepackage{amsfonts} 
\usepackage{listings}
\usepackage{newfloat}

\usepackage{bibentry}
\newcommand{\codesize}{\fontsize{8pt}{9pt}\ttfamily}

\usepackage[most]{tcolorbox}
\newtcblisting{mycode}{
    arc=2mm, outer arc=2mm,
  listing only, 
  listing options={
    basicstyle=\codesize,
    numbers=left,
    numberstyle=\scriptsize,
     numbersep=5pt,
    breaklines=true
  },
  hbox
}

\newtheorem{explanation}{Explanation}

\DeclareCaptionStyle{ruled}{labelfont=normalfont,labelsep=colon,strut=off} 
\floatstyle{ruled}
\newfloat{listing}{tb}{lst}{}
\floatname{listing}{Listing}
\newcommand{\li}{\textsc{LINT}}

\linenumbers

\title {Assessing the Interpretability of Programmatic Policies \\ with Large Language Models
}

\author{
\normalsize{Zahra Bashir, Michael Bowling, Levi H. S. Lelis}
\affiliations
\normalsize{Department of Computing Science, University of Alberta, Canada}
\affiliations
\normalsize{Alberta Machine Intelligence Institute(Amii)}
}

\begin{document}
\nolinenumbers
\maketitle

\begin{abstract}
Although the synthesis of programs encoding policies often carries the promise of interpretability, systematic evaluations were never performed to assess the interpretability of these policies, likely because of the complexity of such an evaluation. In this paper, we introduce a novel metric that uses large-language models (LLM) to assess the interpretability of programmatic policies. For our metric, an LLM is given both a program and a description of its associated programming language. The LLM then formulates a natural language explanation of the program. This explanation is subsequently fed into a second LLM, which tries to reconstruct the program from the natural-language explanation. Our metric then measures the behavioral similarity between the reconstructed program and the original. 
We validate our approach with synthesized and human-crafted programmatic policies for playing a real-time strategy game, comparing the interpretability scores of these programmatic policies to obfuscated versions of the same programs. Our LLM-based interpretability score consistently ranks less interpretable programs lower and more interpretable ones higher. These findings suggest that our metric could serve as a reliable and inexpensive tool for evaluating the interpretability of programmatic policies.

\end{abstract}

\section{Introduction}

There is a growing interest in the use of programmatic representations of policies to solve sequential decision-making problems, both in single-agent~\cite{pmlr-v80-verma18a,qiu2022programmatic} and multi-agent settings~\cite{MarinoMoraesToledoLelis,medeiros2022can}. This interest is justified as one can provide strong inductive bias to the learning process through the domain-specific language defining the space of programs. This bias can allow programmatic policies to generalize more easily to unseen settings~\cite{Inala2020Synthesizing} and make them more amenable to verification~\cite{viper}. 

Previous work on programmatic policies also often emphasizes interpretability. However, systematic studies that assessed the interpretability of these policies were never performed. A common method is to present specific programs and claim their interpretability~\cite{pmlr-v80-verma18a,aleixo23}. The scarcity of comprehensive evaluations could be attributed to the fact that such studies are time consuming and costly, mainly because they would involve human programmers. This lack of a thorough analysis hinders our understanding of what precisely makes a programmatic policy interpretable. For instance, neural networks can be viewed as programs written in a domain-specific language that allows the addition of layers and nodes to the neural architecture---clearly, the programmatic framing for policies does not guarantee interpretability. So, what are the properties that make a programmatic policy interpretable? 

Any viable approach to addressing this question is likely to involve evaluating the interpretability of programmatic policies. In this paper, we introduce a simple and cost-effective methodology to assess program interpretability and demonstrate its application to programmatic policies. Our methodology uses large language models (LLMs)~\cite{brown2020llm} to assign an interpretability score to a program. We call this score the LLM-based INTerpretability (\li) score. In our methodology, we use an instance of an LLM to generate a natural-language explanation of a program. This explanation is given as input to another instance of an LLM, which is asked to reconstruct the program described in the explanation. A third instance of an LLM verifies that the explanation is in natural language and does not provide step-by-step programming instructions on how to write the program. The \li\ score is the value of a metric comparing the behavior of the original and reconstructed programs. We introduce general behavior metrics for sequential decision-making problems.

The evaluation of our methodology is based on methods from the program obfuscation literature~\cite{collberg2009surreptitious}. Obfuscated programs are designed to be non-interpretable, and some obfuscation techniques allow us to construct programs with different levels of obfuscation. Assuming that obfuscation can be used as a proxy for interpretability, we hypothesize that the \li\ scores negatively correlate with the degree of obfuscation we apply to the programs. 
Our methodology also includes the use of programmatic policies written by humans. Our premise is that since these policies are human-written, they should be inherently interpretable, and thus be scored as such in our metric. 




Empirical results on classical programming problems and programmatic policies for playing MicroRTS~\cite{ontanon2018first} show that the \li\ scores strongly and negatively correlate with the level of obfuscation of the programs evaluated. 
Although user studies should still be the gold standard for evaluating interpretability, our results suggest that \li\ can be used as a reliable and inexpensive tool to help drive research in interpretable programmatic policies.

\section{\li: LLM-based Interpretability Score}

\begin{figure}
    \centering
    \begin{tikzpicture}[node distance=3cm, auto]
        \tikzstyle{inputBox} = [rectangle, draw, fill=blue!10, 
                                text width=2em, text centered, 
                                minimum height=2em]
        \tikzstyle{largeBox} = [rectangle, draw, fill=red!10, 
                                text width=6em, text centered, 
                                minimum height=2em]
        \tikzstyle{scoreBox} = [rectangle, draw, fill=yellow!10, 
                                text width=2em, text centered, 
                                minimum height=2em]
        \tikzstyle{arrow} = [thick,->,>=stealth]
    
        \node [inputBox] (constraints) {\( C \)};
        \node [largeBox, right=1.0cm of constraints] (explainer) {Explainer}; 
        \node [inputBox, right=0.5cm of explainer] (program) {\( \pi \)};
        \node [inputBox, below=0.5cm of constraints] (dsl) {DSL};
        \node [largeBox, above=0.5cm of explainer] (verifier) {Verifier};
        \node [largeBox, below=0.5cm of explainer] (reconstructor) {Reconstructor};
        \node [inputBox, right=0.5cm of reconstructor] (programPrime) {\( \pi' \)};
        \node [scoreBox, below right=-0.1cm and 0.5cm of program] (score) {\( B \)};
    
        \draw [arrow] (program.west) -- (explainer.east);
        \draw [arrow] (dsl.east) -- (explainer.west);
        \draw [arrow] (dsl.east) -- (reconstructor.west);
        \draw [arrow] (constraints.east) -- (explainer.west);
        \draw [arrow] (explainer) -- (verifier);
        \draw [arrow] (verifier) -- (explainer);
        \draw [arrow] (explainer) -- (reconstructor);
        \draw [arrow] (reconstructor) -- (programPrime);
        \draw [arrow] (program.east) -| (score.north); 
        \draw [arrow] (programPrime.east) -| (score.south);
        \draw [arrow] (score.east) -- ([xshift=0.5cm]score.east) node[above] {LINT};
    \end{tikzpicture}

    \caption{General overview of \li. The Explainer receives a program $\pi$, a set of constraints $C$, and a description of the DSL in which $\pi$ was written; it produces a natural language explanation of $\pi$, which is checked by the Verifier. The explanation is provided as input, along with the description of the DSL, to the Reconstructor, which attempts to reconstruct $\pi$ from the explanation, thus producing $\pi'$. $B$ scores the similarity (or dissimilarity) of $\pi$ and $\pi'$.}
    \label{fig:lis}
\end{figure}
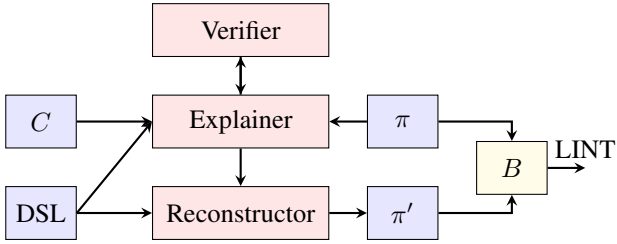

\setlength{\abovecaptionskip}{2pt} 



We define the function $B(\pi_1, \pi_2)$ as a similarity metric for the behavior of two programs. We consider functions $B$ that return a number between $0$ and $1$, where the value of $0$ represents the most dissimilar behavior for the two programs and $1$ represents identical behavior for the programs.\footnote{In our experiments, we also consider a dissimilarity metric, where $0$ represents the most similar and $1$ the most dissimilar behavior.} We denote by $\mathcal{L}_e(\pi, G, C)$ an LLM that receives a program $\pi$, a domain-specific language (DSL) $G$, a set of constraints $C$, and returns a natural language explanation of $\pi$. We refer to this LLM as \textbf{explainer}. We denote by $\mathcal{L}_r(e, G)$ an LLM that receives a natural language explanation $e$ of a program and a DSL $G$, and returns a program accepted by the language $G$ that exhibits the behavior described in $e$. We refer to this LLM as the \textbf{reconstructor}. Both the explainer and the reconstructor receive a natural language description of $G$ with a context-free grammar that specifies the programs $G$ accepts. 

Given a set $\Pi$ with $n$ programs and a behavior metric $B$, the \li\ score is computed as 
\begin{equation}
\li(\Pi, B) = \frac{1}{n} \sum_{\pi \in \Pi} B(\pi, \mathcal{L}_r(\mathcal{L}_e(\pi, G, C), G))\,.
\label{eq:li}
\end{equation}
The \li\ score of set $\Pi$ is the average value of how similar the programs in $\Pi$ are from the reconstructed ones. We define the \li\ score over a set of programs to measure the interpretability of the programs a system generates. However, in our experiments we also consider the case where $|\Pi| = 1$. 

Figure~\ref{fig:lis} shows a schematic view of how the \li\ score is computed for a program $\pi$.

\subsection{Set of Constraints for Explanation} 

The above formulation considers a set of constraints $C$ to generate the explanation of a program. $C$ prevents the LLM from generating the explanation of the program with non-interpretable elements that communicate the program to the other LLM. The constraints are instructions in the LLM prompt. We include the constraints shown in the list below. 
\begin{enumerate}
\item \emph{Try to understand what is happening in the code and explain it in natural language to someone who wants to learn about this program.}
\item \emph{Write a high-level explanation and do not explain the code line-by-line, but it is fine to include numbers in your natural language explanation.}
\item \emph{You must not use programming language jargon as people not familiar with programming might not understand the explanation.}
\end{enumerate}
Without these constraints, the LLM could generate line-by-line instructions of how to reconstruct the program. For example, even if the program was an implementation of the neural network, the LLM could provide instructions on how to implement the architecture and copy the weights of the model. Although this explanation could allow the second LLM to reconstruct the program, the original program might not be interpretable. Even with these constraints, the LLM occasionally generates explanations that use programming instructions such as \emph{``[...] after a nested for-loop [...]''}. 
We use a third LLM, the \textbf{verifier}, to partially check for the constraints. Specifically, we ask it to verify whether the explanation uses computer programming jargon and/or keywords of the DSL. If the verifier answers `yes' to the use of jargons, then we sample another explanation from the explainer. 

\subsection{Multiple Trials}

Due to the stochastic nature of how the LLMs generate the explanations and programs, we repeat $k$ times the computation of $B(\pi, \mathcal{L}_r(\mathcal{L}_e(\pi, G, C), G))$ in Equation~\ref{eq:li} and use in the summation the best $B$-value of the $k$ trials. Trials are carried out by generating one explanation for each program, and each explanation is used to generate $k$ programs. The value of $k$ should be large enough to account for the variance of the LLM generation and small enough to prevent the LLM from reconstructing the original program by chance. Since the program space is vast, as we evaluate empirically, it is safe to use a few trials to compute the \li\ score without allowing the LLM to reconstruct the correct program by chance. 

\section{Caveats of \li\ Score}

When assessing the interpretability of programs, we assume a level of knowledge of the person interpreting them. \li\ assumes the knowledge of an LLM, which may not reflect reality due to a mismatch of knowledge between the LLM and the target audience of the program. For example, if the goal of having interpretable programs is to teach people strategies for playing a real-time strategy game, then the LLM might have deeper knowledge of this genre of game than rookie players trying to learn strategies from the programs. As a result, a policy that is ``interpretable'' for the LLM is not necessarily interpretable to the target audience. Conversely, if the program requires knowledge that the LLM does not possess (e.g., $\pi$ is written in a DSL different from the languages with which the LLM is trained), \li\ can produce false negatives. 

Similarly to the BLEU score~\cite{papineni2002bleu}, \li\ should not be used as an objective function. Using \li\ as such could cause the system to disregard $C$, and the explainer could generate non-interpretable explanations. Instead of using it as a target, \li\ can be used as a tool to assess the interpretability of computer-written programs, to bias design decisions made during the development cycle of synthesizers. 

\section{Empirical Methodology}

The primary objective of our evaluation is to check whether the \li\ scores correlate with the interpretability of a given set of programs. We rely on methods from the static obfuscation literature~\cite{collberg2009surreptitious} to generate programs with different levels of interpretability. Static obfuscation algorithms have the goal of transforming a program before it starts running into less interpretable programs, with the goal of making it harder for adversarial agents to gain knowledge of the program by reading its implementation. For that, we consider semantics-preserving obfuscation transformations, where we can control the degree to which a program is obfuscated. We hypothesize that \li\ scores correlate with the degree of obfuscation of a set of programs. 

We consider two instances of \li: one for evaluating the interpretability of programs that encode solutions to programming tasks; and another for evaluating programmatic policies~\cite{MarinoMoraesToledoLelis} for playing MicroRTS, a real-time strategy game~\cite{ontanon2018first}. \textbf{We provide the complete set of prompts used in our experiments in the Supplementary Materials.} All experiments used GPT-4~\cite{openai2023gpt4}. We use $k = 5$ in all our experiments.



\begin{figure}[htbp]
\centering 
\begin{mycode}
void subsets(char *av[], int c, int n, 
char *sbset[], int sz) {
    if (c == n) {
        if (sz < n) {
            for (int i = 0; i < sz; i++) 
                printf(sbset[i]);
            printf("----------");
        }
        return;
    }
    sbsets(av, c+1, n, sbset, sz);
    sbset[sz] = av[c];
    sbsets(av, c+1, n, sbset, sz+1);}

main(Q,O)char**O;{if(--Q){main(Q,O);O[Q]
[0]^=0X80;for(O[0][0]=0;O[++O[0][0]]!=0;)
if(O[O[0][0]][0]>0)puts(O[O[0][0]]);
puts("----------");main(Q,O);}}
\end{mycode}
  \caption{Non-obfuscated code for computing proper subsets (lines 1--13); an obfuscated program for the same problem (line 15).}
  \label{fig:obfuscated_example} 
\end{figure}

\subsection{Classical Programming Problems}

We consider $10$ programs written in C for solving the following problems: computation of factorials, addition of two numbers, conversion of byte to binary, computation of all proper subsets of a set of arguments, of the value of $\pi$, of $\ln(n)$ for any $n$, of the smallest 100 prime numbers, of the square root of a number, sorting elements, and a program to play tic-tac-toe. The obfuscated versions of these programs were designed so that they would be as non-interpretable as possible, since all obfuscated programs we use are winning entries of the International Obfuscated C Code Contest~\cite{ioccc}. The obfuscated programs were constructed using different techniques, such as replacing sequences of instructions with equivalents that are less interpretable~\cite{cohen1993osprotection}. Figure~\ref{fig:obfuscated_example} shows an example of the programs used in our experiment, where the first function is a non-obfuscated implementation for computing the proper subsets of a set of numbers, while the second is an obfuscated implementation to solve the same problem. The proper subsets of a set $I$ include all subsets except $I$. The complete set of programs is provided in the Supplementary Materials. 


The function $B$ we consider in this experiment measures the number of input values that the reconstructed program correctly maps to their corresponding output value. A $B$-value of $1.0$ indicates that the reconstructed program mapped all inputs to the correct output; a value of $0.0$ indicates that the reconstructed program failed on all inputs. 

\subsection{Programmatic Policies}

We also used programmatic policies for MicroRTS. These programs are categorized into two types: ``synthesized'' policies, written by a computer program in the domain-specific language known as the Microlanguage~\cite{MarinoMoraesToledoLelis}, and ``human-crafted'' policies, written in Java by human programmers. Both types of programs receive a state of the game and return the action the agent performs in that state.


We consider the two-player version of MicroRTS, where each player controls a number of units to collect resources, build structures, and train other units that will eventually battle the opponent. Programmatic policies are the current state of the art in this domain, with programmatic policies winning the last three competitions.\footnote{\url{https://sites.google.com/site/micrortsaicompetition}} MicroRTS has the following types of unit: Worker, Light, Ranged, Heavy, Base, and Barracks. The first four types can move around a gridded map where the game is played and attack opponent units; Workers can collect resources and build Bases and Barracks; Bases can train more Worker units and store resources, while Barracks can train non-Worker units. Units differ in how much damage they can inflict on opponent units and in how much damage they can suffer before being removed from the game. 

\begin{figure}[htbp]
\centering 
\begin{mycode}
for(Unit u)
    for(Unit u)
        u.train(Worker,Up,2)
    u.attack_if_in_range()
    u.train(Heavy,EnemyDir,8)
for(Unit u)
    u.train(Light,Left,100)
    u.build(Barracks,EnemyDir,1)
    u.harvest(25)
    u.attack(Closest)
\end{mycode}
  \caption{Policy written in the Microlanguage.} 
  \label{fig:microrts_example} 
\end{figure}

\vspace{0.05in}
\noindent
\textbf{Microlanguage}

\noindent
The Microlanguage allows programs to iterate through all units the player controls, so it assigns an action to each of the units. The language also supports if-then-else structures. Figure~\ref{fig:microrts_example} shows an example of a program synthesized with Local Learner (2L), a self-play algorithm~\cite{moraes2023choosing}. The loops allow for an action prioritization scheme. This is because once an action is assigned to a unit, it cannot be overwritten by another action, so the instructions in the earlier for-loops will be assigned first. In the program shown in Figure~\ref{fig:microrts_example}, training Worker units has the highest priority because the instruction for training these units is in the first nested for-loop to be executed (lines 2 and 3), which iterates through all units until it eventually finds a Base that will train them. Other actions that require the use of resources (e.g., constructing a Barracks in line 10), will be executed only if the player has enough resources after training Worker units. 

\vspace{0.05in}
\noindent
\textbf{Java}

\noindent
The human-crafted policies are written in Java and follow standard Java principles. This allows for the representation of more complex policies, but lacks the Microlanguage's domain-specific approach. Figure~\ref{fig:microrts_java_example} shows an example.



\begin{figure}[htbp]
\centering 
\begin{mycode}
for (Unit u : pgs.getUnits()) 
    if (u.getType() == barracks 
    && u.getPlayer()== player 
    && gs.getActionAssignment(u) == null) 
        if (p.getResources() >= light.cost)
            train(u, light);
\end{mycode}
  \caption{Policy written in Java by human programmer.}
  \label{fig:microrts_java_example} 
\end{figure}


\vspace{0.05in}
\noindent
\textbf{Obfuscating Programmatic Policies}

\noindent
In the experiment with programmatic policies, we modified the programs to create different levels of interpretability, to verify whether the \li\ scores correlate with these levels. We achieve this using the obfuscation technique of adding useless snippets to the programs, which is a known program obfuscation technique~\cite{cohen1993osprotection}.
We consider two levels of obfuscation: level 1, where we add a few lines of code that do not change the behavior of policy, and level 2, where we add a greater number of such lines compared to level 1. For the synthesized set, we add 10 and 23 lines for levels 1 and 2, respectively; for the human-crafted set, we add 38 and 71 lines for levels 1 and 2, respectively. Under the assumption that programs with longer useless snippets are less interpretable than programs with shorter snippets, we hypothesize that \li\ assigns higher scores to non-obfuscated programs, lower scores to level 1, and the lowest scores to level 2.  

Figure~\ref{fig:level_1} shows a sample of a snippet that we add to the programmatic policies used in our experiments for level 1 of obfuscation; all snippets are shown in the Supplementary Materials, including level 2 snippets. The snippet in Figure~\ref{fig:level_1} does not change the behavior of the policy because the only unit that can harvest resources is a builder, so line 6 does not change the behavior of the policy. 


\begin{figure}[htbp]
\centering 
\begin{mycode}
if (u.canHarvest()):
    for (unit u)
        if (u.isBuilder()):
            pass
        else:
            u.harvest(50)
\end{mycode}
  \caption{Sample of useless code snippet used in level 1.} 
  \label{fig:level_1} 
\end{figure}

\vspace{0.05in}
\noindent
\textbf{Set of Policies Evaluated}

\noindent
For the synthesized set $\Pi$, we selected a subset of size 20 programs from the totally ordered set with approximately $1,000$ programmatic policies 2L synthesized for the BaseWorkers-16$\times$16A map. Two adjacent policies in the ordered set are likely to be similar to each other due to the process in which 2L synthesizes them. We select 20 uniformly spaced policies from the ordered set to obtain a more diverse subset. That is, given that we have $m$ policies in the ordered set, we select the policies with indices $\big \lfloor \frac{i \times (m - 1)}{(19)} \big \rfloor$ with $i = 0, \cdots, 19$. For the human-crafted set, we used 10 programs selected from a collection available on GitHub.\footnote{\url{https://github.com/rubensolv/SCV/tree/master/pvai}} We present all the programs used in our study in the Supplementary Materials.


\vspace{0.05in}
\noindent
\textbf{Behavior Metrics}

\noindent
We used three behavior metrics $B$ for programmatic policies. For all metrics, we consider a set of $10$ policies, which are chosen from a totally ordered set of programmatic policies 2L synthesized; we refer to this set as the set of opponents $\mathcal{O}$. Although the policies evaluated and the set of opponents are selected from the same pool of programs, there is no overlap between the two sets. We ensure that our metric results are not skewed by having overly weak or overly strong opponents. This is achieved by, while sampling policies for $\mathcal{O}$, rejecting those that win or lose all matches against the set of 20 policies we evaluate in our experiment. 
Let $S_{\pi, o}$ be the set of states in which the policy $\pi$ is to act in a match played with the opponent $o$ in $\mathcal{O}$. Also, let $S_{\pi} = \bigcup_{o \in \mathcal{O}} S_{\pi, o}$ be the union of the states of all matches played with the opponents. 

The first metric, which can be applied to any sequential decision-making problem with discrete action spaces, is the fraction in which the actions chosen by the reconstructed program $\pi'$ match the actions chosen by the original program $\pi$ for states in the set $S_{\pi}$: $|S_{\pi}|^{-1} \times \sum_{s \in S_{\pi}} \mathbf{1}[\pi(s) = \pi'(s)]$, where $\mathbf{1}[\cdot]$ is the indicator function. We refer to this metric as the \textbf{action metric}. If the reconstructed program is equivalent to the original, then the action metric is $1.0$. 

The second metric, which can be applied to any zero-sum game, compares the signature of wins, draws, and losses of the reconstructed policy with the signature of the original policy. The signature $a_\pi$ of a policy $\pi$ is a vector of size $|\mathcal{O}|$ where each entry $i$ assumes the values of $1$, $0$, or $-1$, representing the result of a win, draw, or loss, respectively, of a match played between $\pi$ and the $i$-th opponent in $\mathcal{O}$. This metric computes $|\mathcal{O}|^{-1} \times \sum_{i = 1}^{|\mathcal{O}|} \mathbf{1}[a_\pi[i] = a_{\pi'}[i]]$, where $a_\pi[i]$ represents the $i$-th entry of $a_\pi$. We call this metric the \textbf{outcome metric}. Similarly to the action metric, if $\pi'$ is equivalent to $\pi$, then the outcome metric value is $1.0$.

The third metric compares the set of features observed in matches between the reconstructed program and $\mathcal{O}$ with the features observed in matches between the original program and $\mathcal{O}$. Let $F(\pi, o)$ be a vector of features observed in a match between $\pi$ and $o$. We use the seven features of \citeauthor{aleixo23}~\shortcite{aleixo23}, where each feature is the sum of the number of units of a given type that the player trained (or built) in all states of the match; the types can be Worker, Light, Heavy, Ranged, Base, or Barracks. A last feature sums up the amount of resources collected in the match. This metric measures the average normalized L1 norm between the feature vector of the original and reconstructed programs. We refer to this metric as \textbf{feature metric}. If the reconstructed program is equivalent to the original, then this metric is $0.0$. While other metrics measure similarity, the feature metric measures dissimilarity. 

We use these three metrics because each of them individually has weaknesses; together, they offer a more reliable summary of the behavior of a policy. Many of the actions in a MicroRTS match are related to Worker units collecting resources, so while two policies might encode totally different strategies, due to the large number of Worker units collecting resources, the policies could have a large action metric value. The outcome metric can also be misleading if almost any policy defeats the set of opponents or is defeated by the set of opponents (i.e., the opponents are too weak and/or too strong). Finally, the feature metric simply counts the number of units and resources, without measuring their behavior. 

\vspace{0.05in}
\noindent
\textbf{Baselines for Reconstructed Programs}

\noindent
We consider a number of programs as baselines for the programs \li\ reconstructs. Namely, for each program $\pi$ in $\Pi$, we compare the behavior metric values for the reconstructed program $\pi'$ of $\pi$ with a randomly selected program from $\Pi$ that is different from $\pi$; we call this baseline \textbf{Rand}. 
In the experiment with the synthesized set, 
since all programs in $\Pi$ were generated in a single run of 2L and for a fixed map, the programs Rand selects can be similar to $\pi$. 


In another baseline, where we select a random program from the pool of programs 2L synthesizes for a different map; we use programs synthesized for the BaseWorkers-8$\times$8 map with this baseline. Since the strategy for playing the game can change drastically from map to map, this randomly selected program is likely to be less similar to $\pi$ than the programs Rand selects. We call this baseline \textbf{Rand-Other}. 

Another baseline we consider selects the policy from the set of evaluated policies $\Pi$ that is different from $\pi$ but is most similar to $\pi$ with respect to its syntax. We treat each line of a program as an element of a set. The program most similar to $\pi$ is the one whose intersection with the syntax set of $\pi$ is the largest. We refer to this baseline as \textbf{Closest-Syntax}. 

The last baseline we consider returns the policy in $\Pi$ different from the evaluated program $\pi$ with the smallest L2 norm for the difference between its feature vector and the feature vector of $\pi$. We refer to this baseline as \textbf{Closest-Feature}. 


\vspace{0.05in}
\noindent
\textbf{Baseline for the Reconstruction System}

\noindent
The Reconstructor relies on the natural language explanation of the original program that the Explainer provides. To ensure that the program cannot be reconstructed by sampling programs without the explanation, we consider a baseline where we ask an LLM to generate a program for playing on a given map of MicroRTS. We provide a natural language description of the BaseWorkers-16$\times$16A map and ask the model to generate a program written in the Microlanguage that can ``play well'' the game. We repeat this process $k$ times, exactly how it is implemented in the reconstruction system for \li, and we return the program with the best behavior score out of the $k$ evaluated. We refer to this baseline as the \textbf{$k$-Shot LLM}. 


\section{Empirical Results: Programming Problems}

The system was able to reconstruct all non-obfuscated programs to solve the programming problems and achieved the score of $1.0$. The system was unable to reconstruct any of the obfuscated programs, resulting in a score of $0.0$. As a representative example, Explanation~\ref{explanation:subset_non} shows the Explainer's output for the non-obfuscated program shown in Figure~\ref{fig:obfuscated_example}. 


\begin{explanation}[Proper Subset - Non-Obfuscated]
Imagine you have a list of items, and you want to figure out all the different ways you can select some of these items without taking all of them at once. When the program starts, it first checks if you've given it any items to work with. If not, it asks for some. Once it has the list of items, it begins to explore all the possible selections. It does this by starting with the first item and deciding whether to include it in the current selection or not. If it decides not to include the first item, it moves to the next item and makes the same decision. It continues doing this until it has considered every item in the list. After considering all the items, if it has left out at least one item from the current selection, it displays that selection for you to see. Then, it goes back and tries a different combination by including some items that it previously left out and leaving out some that it previously included. It continues this process until it has explored and shown you all possible selections of items from the list, always making sure it never shows you a selection that includes all the items at once. In the end, you're left with a display of all the different ways you can select some of the items without taking all of them.
\label{explanation:subset_non}
\end{explanation}

The explanation of the program is high-level and makes an analogy with a list of items, and the process of computing the proper subsets is described as a process of ``selecting'' and ``displaying'' the items, where all possible combinations of subsets are selected: \emph{``It does this by starting with the first item and deciding whether to include it in the current selection or not''}. The Reconstructor can produce a correct implementation of a program that computes the proper subsets from this explanation. Explanation~\ref{explanation:subset_obs} shows the explanation \li\ generated for the obfuscated program in Figure~\ref{fig:obfuscated_example}. 

\begin{explanation}[Proper Subset - Obfuscated]
Imagine you have a list of items. Initially, all items are marked as ``normal.''
The program goes through the following steps:
\begin{enumerate}
    \item It starts at the end of the list and works its way to the beginning. For each item, the program does two actions:First, it keeps the item as ``normal'' and then displays all the items marked as ``greater than normal.'' After displaying these items, it separates them with a line of dashes. Second, it switches the item from ``normal'' to ``greater than normal'' and repeats the display process.
    \item After dealing with an item, the program moves to the next item closer to the beginning of the list and repeats step 2.
\end{enumerate}
This process continues until the program has considered all items in the list.
\label{explanation:subset_obs}
\end{explanation} 

This explanation is well-structured and includes the steps that supposedly need to be performed. However, the description is not clear in some parts. For example, it is not clear what ``greater than normal'' means. 



\section{Empirical Results: Programmatic Policies}

\begin{table}
\footnotesize
\setlength{\tabcolsep}{2.8pt}
\begin{tabular}{lrrr}
\toprule
& \multicolumn{3}{c}{Metrics $B$}
\\\cmidrule(lr){2-4}
& \multicolumn{1}{c}{Action ($\uparrow$)}   & \multicolumn{1}{c}{Outcome ($\uparrow$)} & \multicolumn{1}{c}{Feature ($\downarrow$)} \\
\midrule
 \li & $0.940 \pm 0.010$ & $0.840 \pm 0.042$ &$0.133 \pm 0.020$ \\
 Rand & $0.733 \pm 0.015$ & $0.615\pm 0.057$ & $0.418 \pm 0.018$\\
 Rand-Other & $0.564 \pm 0.025$ & $0.470 \pm 0.058 $ & $0.486 \pm 0.012$ \\
 Closest-Syntax & $0.799 \pm 0.015$ & $0.600 \pm 0.056$ &$0.403 \pm 0.018$ \\
 Closest-Feature & $0.823 \pm 0.013$ & $0.770 \pm 0.049$ & $0.189 \pm 0.014$ \\
 $k$-Shot LLM & $0.343 \pm 0.010$ & $0.420 \pm 0.057$ & $0.441 \pm 0.009$ \\
 \bottomrule
\end{tabular}
\caption{Value of the behavior metrics for programmatic policies. Action and Outcome metrics are similarity metrics, so higher values are better ($\uparrow$), while Feature is a metric of dissimilarity, so lower values are better ($\downarrow$). The cells show the metric values and the 95\% confidence interval.}
\label{tab:similarity_metrics}
\end{table}


\begin{table}
\footnotesize
\begin{tabular}{lrrr}
\toprule
\multicolumn{1}{c}{\multirow{2}{*}{Metric}} & \multicolumn{1}{c}{\multirow{2}{1cm}{Original \\ Program}} & \multicolumn{2}{c}{Obfuscation Level}
\\\cmidrule(lr){3-4}
&  & \multicolumn{1}{c}{Level 1} & \multicolumn{1}{c}{Level 2} \\
\midrule
Action & 
0.945 $ \pm 0.010$ & 
0.866 $ \pm 0.014$ & 
0.732 $ \pm 0.012$ \\

Outcome & 
0.840 $ \pm 0.042$& 
0.705 $ \pm 0.053$& 
0.490 $ \pm 0.058$  \\

Feature & 
0.133 $ \pm0.020$ &
0.272 $ \pm0.024$ & 
0.418 $ \pm 0.018$ \\
\bottomrule
\end{tabular}
\caption{Average values of the behavior metrics for the original program and for the two levels of obfuscation; the cells also show the 95\% confidence interval.}
\label{tab:obfuscation_policies}
\end{table}







Table~\ref{tab:similarity_metrics} shows the average and 95\% confidence interval values for the 20 programs used in our experiment with the synthesized set. The \li\ row shows the metric values computed for the original programs and their reconstructions. The baseline values represent the measurements between the original programs and the baseline programs. The values of \li\ are the best according to all metrics, which shows that the reconstructed programs are more similar to the original than any of the baselines. Table~\ref{tab:similarity_metrics} also shows that the baselines that obtain values closer to \li\ are Rand, Closest-Syntax, and Closest-Feature. This is because these baselines select a program from the pool of programs 2L synthesized for the same map, and these programs tend to be similar to each other. Rand-Other obtained lower Action and Outcome values and a higher Feature value, demonstrating that the metrics can capture the expected differences between policies synthesized for playing in different maps. Finally, $k$-Shot LLM presents the lowest similarity values, demonstrating the importance of the explanation of the original program the Explainer generates. 

Table~\ref{tab:obfuscation_policies} presents the results that test our hypothesis that \li\ correlates with the degree of interpretability of the programs. 
The results indicate a higher similarity between the original and reconstructed programs for the policies 2L synthesizes than between the original obfuscated programs and their reconstructions. Also, \li\ provides higher similarity scores and a lower dissimilarity score for Level 1 than for Level 2 obfuscation. Table~\ref{tab:obfuscation_policies_human_set} presents the results for the human-crafted set. These results align with those of the synthesized set, where the \li-score decreases as we increase the level of obfuscation. Under the assumption that higher degrees of obfuscation result in less interpretable programs, the results support our hypothesis. 

\begin{table}
\footnotesize
\begin{tabular}{lrrr}
\toprule
\multicolumn{1}{c}{\multirow{2}{*}{Metric}} & \multicolumn{1}{c}{\multirow{2}{1cm}{Original \\ Program}} & \multicolumn{2}{c}{Obfuscation Level}
\\\cmidrule(lr){3-4}
&  & \multicolumn{1}{c}{Level 1} & \multicolumn{1}{c}{Level 2} \\
\midrule
Action & 
0.98 $ \pm 0.003$ & 
0.92 $ \pm 0.017$ & 
0.85 $ \pm 0.031$ \\

Outcome & 
0.95 $ \pm 0.036$& 
0.84 $ \pm 0.060$& 
0.80 $ \pm 0.066$  \\

Feature & 
0.07 $ \pm0.022$ &
0.19 $ \pm0.029$ & 
0.22 $ \pm 0.031$ \\
\bottomrule
\end{tabular}
\caption{Average values of the behavior metrics for the original program and for the two levels of obfuscation; the cells also show the 95\% confidence interval.}
\label{tab:obfuscation_policies_human_set}
\end{table}

\subsection{Representative Sample}

In this section, we present the natural language explanation the Explainer generated for the program in Figure~\ref{fig:microrts_example} (Explanation~\ref{explanation:policy}). Then, we present the program the Reconstructor generates from the explanation (Figure~\ref{fig:reconstructed_microrts}). This is a representative example of the programs used in the experiments. 

\begin{explanation}[Programmatic Policy]
The script outlines a two-tiered strategy in the game of MicroRTS. It focuses on the tactical management of different types of units, ranging from Workers, to Heavy units, and finally Light units. 
The first part of the strategy involves a layered assignment of tasks. Every unit is asked to train two Worker units if possible. These Workers can be seen as the backbone of your army since they are the only units that can build structures, which in turn produce other units. At the same time, if a unit can't train a Worker, it is assigned to remain idle, thus serving as a defensive guard to your base. In addition, every unit is asked to train up to 8 Heavy units, that typically are stronger and can inflict more damage, acting as the frontline offensive unit. This phase essentially sets up your base with a balance of productive Workers and Heavy units. 

In the second part of the strategy, each unit is tasked with producing as many Light units as possible, up to a whopping 100. Light units are generally quicker and can swiftly react to the changes in the battlefield. Additionally, every unit is tasked to build a Barracks near the enemy's direction, effectively establishing a forward base closer to the enemy territory. They are also asked to harvest resources, up to 25 units, ensuring a steady supply for creating more units and structures. Finally, units are directed to attack the closest enemy units, marking an aggressive stance towards the opponent.

The overall goal of this strategy is to ensure a robust base with a mixture of Worker and Heavy units, while also maintaining an aggressive stance with a large army of Light units and Barracks near the enemy territory. It aims at resource gathering for continued production of units and structures, and pushing the opponent back through relentless attack.
\label{explanation:policy}
\end{explanation}

Similarly to Explanation~\ref{explanation:subset_non}, the explanation describes the policy in a level of detail that allows for the reconstruction of a program that behaves identically to the original program according to our metrics. Figure~\ref{fig:reconstructed_microrts} shows the reconstructed program. 
The reconstructed program is not identical in terms of line-by-line syntax, since information regarding the syntax is lost in the process of translating the program into a natural language explanation and back to a program. A casual inspection might even suggest that the reconstructed program does not behave as the original. This is because the original program has an instruction for training Worker units inside a nested loop, thus giving it the highest priority. The reconstructed program has training Worker units instruction inside the main loop. This means that the program can use the player's resources to assign actions to other units (e.g., train Light units in line 7) and by the time $u$ is a Base in the outer loop, the player no longer has resources for training Worker units. However, a more careful inspection of the program reveals that, in the first states of the game, where the player trains Worker units, none of the actions that use resources can be assigned to a unit: the player cannot train Heavy and Light units (lines 5 and 7, respectively) because the player does not have a Barracks yet; the player cannot build a Barracks (line 8) because it does not have enough resources to do so. Thus, similarly to the original program, the reconstructed one prioritizes the training of Worker units. 

\begin{figure}
\centering
\begin{mycode}
for(Unit u):
    u.train(Worker, Up, 2)
    u.attack_if_in_range()
    for(Unit u):
        u.train(Heavy, EnemyDir, 8)
    for(Unit u):
        u.train(Light, Left, 100)
    u.build(Barracks, EnemyDir, 1)
    u.harvest(25)
    u.attack(Closest)
\end{mycode}
\caption{Reconstruction of the program shown in Figure~\ref{fig:microrts_example}.}
\label{fig:reconstructed_microrts}
\end{figure}

\section{Related Works}

In contrast to the literature on programmatic policies, it is common to find evaluations of the interpretability of models in the context of supervised learning~\cite{ribeiro2016lime,lundberg2017shap,fong2017interpretable}. 
We conjecture that the methodological difference between the programmatic policies and the supervised learning literature is due to the need of enlisting participants who understand both the application domain and computer programming for evaluating programmatic policies, while the latter only requires participants who understand the application domain. 

Previous work in programmatic policies, such as NDPS~\cite{pmlr-v80-verma18a}, Viper~\cite{viper}, Propel~\cite{propel}, and $\pi$-PRL~\cite{qiu2022programmatic}, describe systems that synthesize programmatic policies in the space of  oblique decision trees~\cite{OC1}. Such trees represent programs with if-then-else structures with linear transformations of the inputs. Oblique decision trees are often assumed to be interpretable, which is likely true for small trees and low-dimensional problems, but it is unlikely to hold true for deep trees and high-dimensional problems. 

Metrics to measure code understandability~\cite{Buse_10,Posnett,Daka,Scalabrino2016,Oliveira} from the Software Engineering literature attempt to solve a related but different problem from the interpretability of programmatic policies we tackle in this paper. Code understandability considers scenarios where people write computer code that is meant to be understandable by other people, but it may not be because the person reading the code is not familiar with the API being used, the API documentation is lacking~\cite{Scalabrino}, or because the code is too long and time-consuming for one to understand. 
%
This is in contrast to our experiments on the interpretability of policies, where we assume that one has access to the correct resources (e.g., API description) and enough time to interpret a policy. In the interpretability of policies, we are interested in evaluating programs generated by other programs with the goal of maximizing the agent's reward and not necessarily to be interpretable. 

Recent work showed that there is currently no effective metric to measure code understandability~\cite{Scalabrino}. This contrasts with our results, which suggest that \li\ can be used as a reliable metric to assess the interpretability of policies. Although the two problems have nuanced but important differences, our encouraging results on policy interpretability suggest that future research could investigate the use of LLMs to measure code understandability.

\section{Conclusions}

Programmatic policies are often synthesized with the expectation of interpretability. However, to our knowledge, there has not been a systematic evaluation of the interpretability of such policies, probably due to the cost associated with such an evaluation. Especially because the evaluation of programmatic policies might require human users proficient in computer programming. In this paper, we presented an inexpensive methodology based on LLMs to assess the interpretability of programmatic policies. Namely, we introduced the LLM-based Interpretability (\li) score for programs. The \li\ score of a program is computed by having an LLM generate a description of it in natural language, which is provided as input to another LLM. This second LLM tries to reconstruct the program from the natural language description. The \li\ score measures the similarity between the original and reconstructed programs. Our empirical evaluation of \li\ relied on the literature on program obfuscation and we assumed that obfuscated programs are less interpretable than non-obfuscated ones. Empirical results on programming problems and programmatic policies showed that the \li\ scores of the evaluated programs correlate with their interpretability. Our results suggest that \li\ can be used as a tool to assess the interpretability of programmatic policies.

\appendix

\bibliographystyle{named}
\bibliography{ijcai24}

\end{document}


\section*{Supplementary Materials: \\ Assessing the Interpretability of Programmatic Policies with Large Language Models}
\tableofcontents


\newpage

\section{Discussion}
The prompts for explanation, reconstruction, and verification of both \textit{C Programming Problems} and \textit{Programmatic Policies} (either Synthesized or Human-crafted) are detailed in the subsequent sections. The C Programming Problems and Human-crafted set of policies are developed in C and Java, respectively, whereas the Synthesized set is written in MicroLanguage

The distinction between the two groups is based on the language in which they're implemented. For \textit{C Programming Problems} and \textit{Human-crafted Java Policies}, there's no need to clarify any domain-specific language since the programs are written in a language that the LLM is already acquainted with its terminology. On the other hand, the \textit{Synthesized Policies} utilize a language unique to MicroRTS. As a result, it's necessary to define this language and its components to the LLM. Specifically, for the explanation prompt, we must elucidate the DSL to ensure the LLM comprehends the program it's about to explain. Similarly, the DSL must be defined during reconstruction to ensure the regenerated program follows the MicroRTS programming rules. The same applies to the verifier.

\section{MicroRTS Prompts}
\subsection{Synthesized Set}
\subsubsection{Explainer Prompt}

\begin{tcolorbox}[breakable, width=\textwidth+1cm]
MicroRTS (ONTANÓN, 2013) is an implementation of a real-time strategy game, played between two players. Each player controls a set of units of different types.

\begin{itemize}
    \item Worker units can:
    \begin{enumerate}
        \item collect resources
        \item build structures (Barracks and Bases)
        \item attack opponent units
    \end{enumerate}
    \item Barracks and Bases:
    \begin{enumerate}
        \item Barracks can train combat units. (they can produce Light, Heavy or Ranged units)
        \item Bases can train the Workers. (they can only produce Workers)
        \item They can neither attack opponents units nor move.
    \end{enumerate}
    \item Combat units:
    \begin{enumerate}
        \item can be of type Light, Heavy, or Ranged.
        \item These units differ in:
        \begin{itemize}
            \item how long they survive a battle
            \item how much damage they can inflict to opponent units
            \item how close they need to be from opponent units to attack them.
        \end{itemize}
        \item They can attack the oppponent units.
    \end{enumerate}
    \item Resource units:
    \begin{enumerate}
        \item The source of resources, doesn't belong to any player.
        \item These units cannot execute any actions.
        \item When the number of resources left reaches 0, the unit disappears. (It only has one parameter: resources left.)
    \end{enumerate}
\end{itemize}
Actions are deterministic and there is no hidden information in MicroRTS. A match is played on a map and each map might require a different strategy for defeating the opponent.

This CFG allows nested loops and conditionals. It contains several Boolean functions (B) and command-oriented functions (C) that provide either information about the current state of the game or commands for the ally units. The following describes a scripting language for playing MicroRTS.

The Boolean functions are described below:
\begin{enumerate}
    \item u.hasNumberOfUnits(T, N): Checks if the ally player has N units of type T.
    \item u.opponentHasNumberOfUnits(T, N): Checks if the opponent player has N units of type T.
    \item u.hasLessNumberOfUnits(T, N): Checks if the ally player has less than N units of type T.
    \item u.haveQtdUnitsAttacking(N): Checks if the ally player has N units attacking the opponent.
    \item u.hasUnitWithinDistanceFromOpponent(N): Checks if the ally player has a unit within a distance N from a opponent’s unit.
    \item u.hasNumberOfWorkersHarvesting(N): Checks if the ally player has N units of type Worker harvesting resources.
    \item u.is\_Type(T): Checks if a unit is an instance of type T.
    \item u.isBuilder(): Checks if a unit is of type Worker.
    \item u.canAttack(): Checks if a unit can attack.
    \item u.hasUnitThatKillsInOneAttack(): Checks if the ally player has a unit that kills an opponent’s unit with one attack action.
    \item u.opponentHasUnitThatKillsUnitInOneAttack(): Checks if the opponent player has a unit that kills an ally’s unit with one attack action.
    v u.hasUnitInOpponentRange(): Checks if an unit of the ally player is within attack range of an opponent’s unit.
    \item u.opponentHasUnitInPlayerRange(): Checks if an unit of the opponent player is within attack range of an ally’s unit.
    \item u.canHarvest(): Checks if a unit can harvest resources.
\end{enumerate}

The Command functions are described below. These functions assign actions to units.
\begin{enumerate}
\item  u.build(T, D, N): Builds N units of type T on a cell located on the D direction of the unit.
\item  u.train(T, D, N): Trains N units of type T on a cell located on the D direction of the structure responsible for training them.
\item  u.moveToUnit(T\_p, O\_p): Commands a unit to move towards the player T\_p following a criterion O\_p.
\item  u.attack(O\_p): Sends N Worker units to harvest resources.
\item  u.harvest(N): Sends N Worker units to harvest resources.
\item  u.idle(): Commands a unit to stay idle and attack if an opponent unit comes within its attack range.
\item  u.moveAway(): Commands a unit to move in the opposite direction of the player's base.
'T' represents the types a unit can assume. 'N' is a set of integers. 'D' represents the directions available used in action functions.
'O\_p' is a set of criteria to select an opponent unit based on their current state. 'T\_p' represents the set of target players.
'e' is an empty block which means doing nothing.
\end{enumerate}
The for loops in this scripting language iterate over all units and the instructions inside the for loops attempt to assign actions to each of these units. For example,
for (Unit u){
     u.build(Barracks, EnemyDir, 8)}
     
The snippet above will assign a build action to unit u. Note that the only unit that can build is Worker. In the for loop above, if u is Ranged, for example, then the instruction u.build(Barracks,EnemyDir,8) will be ignored. 
Once an action is assigned to a unit, it cannot be changed. That is why the for loops offer a priority scheme to the actions. The way that for loop are organized is perhaps the most important feature of this language. The program always has the form:
for (Unit u){
              ...}
With the possibility of adding nested for-loops that go through all units. 
The parameter in each instruction limits the amount of the thing that is trained or build. For example, u.build(Barracks, EnemyDir, 8) limits to at most 8 Barracks. If the player already has 8 barracks, then this instruction will be ignored.
Now, you have the background you need to know!

Now, given the above information, first try to understand the meanings of all the boolean (B) and command (C) functions from above and try to relate them in the context of microRTS playing strategies.

Alright, let's take a look at program P in which I want you to write an explanation for:

\textit{\\Program P\\}\\
The following 7 are some guidelines for writing an explanation for this strategy:
\begin{enumerate}

\item Please write a high-level explanation and do not explain the code line by line but try to include numbers and unit names in your natural language explanation.
\item Try to understand what is happening in the code and explain it in natural language for someone who wants to know how to play MicroRTS using this strategy.
\item Write the explanation inside `$<explanation></explanation>$' tag. 
\item DON'T USE any quotation marks in writing the explanation. 
\item At the end of the explanation, write the overall goal of the strategy as well. (include it inside the explanation tag) 
\item Don't forget to mention the numbers but in a natural language way. 
\item The for-loops offer a hierarchy that determines the priority of the actions. 
\end{enumerate}

The list below specifies the different priorities for actions one can obtain with the for-loops (from highest to lowest priority).
\begin{itemize}
\item Actions inserted in nested for-loops at the top of the program receives the highest priority.
\item Actions inserted in nested for-loops that appear later in the program have higher priority than actions that appear outside a nested for loop.
\item Actions outside the nested for-loops that appear earlier in the program have higher priority than actions outside for-loops that appear later in the program.
\end{itemize}
The most important part of the explanation is to be clear with respect to the priority of the actions. That is, what is the action with highest priority, which one follows that one, and so on. You MUST NOT talk in terms of for-loops and programming language jargon as people not familiar with programming will not understand the explanation. Also, you MUST not use any `nest' or `nested' words as they are related to programming elements. You should talk in terms of priorities of the actions, as you would explain the strategy to a gamer. 

For example, if an action is within a nested for loop, then you need to say that the priority to this action is the highest. If an action isn't within a nested for loop, then you must say that the priority isn't the highest. 

So, following the instructions, can you provide a high-level explanation of the provided program that another instance of LLM can rewrite this program from that summary? Remember to verify the priority of the actions, they should be well explained in your text (e.g., this action has the highest priority, this has the second highest and so on).

\end{tcolorbox}

\subsubsection{Reconstructor Prompt}
\begin{tcolorbox}[breakable, width=\textwidth+1cm]
MicroRTS (ONTANÓN, 2013) is an implementation of a real-time strategy game, played between two players. Each player controls a set of units of different types.
\begin{itemize}
    \item Worker units can:
    \begin{enumerate}
        \item collect resources
        \item build structures (Barracks and Bases)
        \item attack opponent units
    \end{enumerate}
    \item Barracks and Bases:
    \begin{enumerate}
        \item Barracks can train combat units. (they can produce Light, Heavy or Ranged units)
        \item Bases can train the Workers. (they can only produce Workers)
        \item They can neither attack opponents units nor move.
    \end{enumerate}
    \item Combat units:
    \begin{enumerate}
        \item can be of type Light, Heavy, or Ranged.
        \item These units differ in:
        \begin{itemize}
            \item how long they survive a battle
            \item how much damage they can inflict to opponent units
            \item how close they need to be from opponent units to attack them.
        \end{itemize}
        \item They can attack the oppponent units.
    \end{enumerate}
    \item Resource units:
    \begin{enumerate}
        \item The source of resources, doesn't belong to any player.
        \item These units cannot execute any actions.
        \item When the number of resources left reaches 0, the unit disappears. (It only has one parameter: resources left.)
    \end{enumerate}
\end{itemize}
Actions are deterministic and there is no hidden information in MicroRTS. A match is played on a map and each map might require a different strategy for defeating the opponent.

Here, I provided a context free grammar (CFG) of microRTS playing strategy inside the $<CFG>$$</CFG>$ tag written bellow: 
\begin{align*}
\text{$<CFG>$} \\
S &\to SS \mid \text{for(Unit u) S} \mid \text{if(B) then S} \\
&\quad \mid \text{if(B) then S else S} \mid C \mid \lambda \\
B &\to u.\text{hasNumberOfUnits}(T, N) \\
&\quad \mid u.\text{opponentHasNumberOfUnits}(T, N) \\
&\quad \mid u.\text{hasLessNumberOfUnits}(T, N) \\
&\quad \mid u.\text{haveQtdUnitsAttacking}(N) \\
&\quad \mid u.\text{hasUnitWithinDistanceFromOpponent}(N) \\
&\quad \mid u.\text{hasNumberOfWorkersHarvesting}(N) \\
&\quad \mid u.\text{canAttack()} \\
&\quad \mid u.\text{hasUnitThatKillsInOneAttack()} \\
&\quad \mid u.\text{opponentHasUnitThatKillsUnitInOneAttack()} \\
&\quad \mid u.\text{hasUnitInOpponentRange()} \\
&\quad \mid u.\text{opponentHasUnitInPlayerRange()} \\
&\quad \mid u.\text{canHarvest()}\\
C &\to u.\text{build}(T, D, N) \mid u.\text{train}(T, D, N) \mid u.\text{moveToUnit}(T_p, O_p) \\
&\quad \mid u.\text{attack}(O_p) \mid u.\text{harvest}(N) \\
&\quad \mid u.\text{idle()} \mid u.\text{moveAway()}\\
T &\to \text{Base} \mid \text{Barracks} \mid \text{Ranged} \mid \text{Heavy} \\
&\quad \mid \text{Light} \mid \text{Worker} \\
N &\to 0 \mid 1 \mid 2 \mid 3 \mid 4 \mid 5 \mid 6 \mid 7 \mid 8 \mid 9 \\
&\quad \mid 10 \mid 15 \mid 20 \mid 25 \mid 50 \mid 100\\
D &\to \text{EnemyDir} \mid \text{Up} \mid \text{Down} \mid \text{Right} \mid \text{Left}\\
O_p &\to \text{Strongest} \mid \text{Weakest} \mid \text{Closest} \mid \text{Farthest} \\
&\quad \mid \text{LessHealthy} \mid \text{MostHealthy} \mid \text{Random}\\
T_p &\to \text{Ally} \mid \text{Enemy}\\
\text{$</CFG>$}
\end{align*}

This CFG allows nested loops and conditionals. It contains several Boolean functions (B) and command-oriented functions (C) that provide either information about the current state of the game or commands for the ally units. The following describes a scripting language for playing MicroRTS.

The Boolean functions are described below:
\begin{enumerate}
    \item u.hasNumberOfUnits(T, N): Checks if the ally player has N units of type T.
    \item u.opponentHasNumberOfUnits(T, N): Checks if the opponent player has N units of type T.
    \item u.hasLessNumberOfUnits(T, N): Checks if the ally player has less than N units of type T.
    \item u.haveQtdUnitsAttacking(N): Checks if the ally player has N units attacking the opponent.
    \item u.hasUnitWithinDistanceFromOpponent(N): Checks if the ally player has a unit within a distance N from a opponent’s unit.
    \item u.hasNumberOfWorkersHarvesting(N): Checks if the ally player has N units of type Worker harvesting resources.
    \item u.is\_Type(T): Checks if a unit is an instance of type T.
    \item u.isBuilder(): Checks if a unit is of type Worker.
    \item u.canAttack(): Checks if a unit can attack.
    \item u.hasUnitThatKillsInOneAttack(): Checks if the ally player has a unit that kills an opponent’s unit with one attack action.
    \item u.opponentHasUnitThatKillsUnitInOneAttack(): Checks if the opponent player has a unit that kills an ally’s unit with one attack action.
    v u.hasUnitInOpponentRange(): Checks if an unit of the ally player is within attack range of an opponent’s unit.
    \item u.opponentHasUnitInPlayerRange(): Checks if an unit of the opponent player is within attack range of an ally’s unit.
    \item u.canHarvest(): Checks if a unit can harvest resources.
\end{enumerate}

The Command functions are described below. These functions assign actions to units.
\begin{enumerate}
\item  u.build(T, D, N): Builds N units of type T on a cell located on the D direction of the unit.
\item  u.train(T, D, N): Trains N units of type T on a cell located on the D direction of the structure responsible for training them.
\item  u.moveToUnit(T\_p, O\_p): Commands a unit to move towards the player T\_p following a criterion O\_p.
\item  u.attack(O\_p): Sends N Worker units to harvest resources.
\item  u.harvest(N): Sends N Worker units to harvest resources.
\item  u.idle(): Commands a unit to stay idle and attack if an opponent unit comes within its attack range.
\item  u.moveAway(): Commands a unit to move in the opposite direction of the player's base.
`T' represents the types a unit can assume. `N' is a set of integers. `D' represents the directions available used in action functions.
`O\_p' is a set of criteria to select an opponent unit based on their current state. `T\_p' represents the set of target players.
`e' is an empty block which means doing nothing.
\end{enumerate}
The for loops in this scripting language iterate over all units and the instructions inside the for loops attempt to assign actions to each of these units. For example,
for (Unit u){
     u.build(Barracks, EnemyDir, 8)}
     
The snippet above will assign a build action to unit u. Note that the only unit that can build is Worker. In the for loop above, if u is Ranged, for example, then the instruction u.build(Barracks,EnemyDir,8) will be ignored. 
Once an action is assigned to a unit, it cannot be changed. That is why the for loops offer a priority scheme to the actions. The way that for loop are organized is perhaps the most important feature of this language. The program always has the form:
for (Unit u){
              ...}
With the possibility of adding nested for-loops that go through all units. 
The parameter in each instruction limits the amount of the thing that is trained or build. For example, u.build(Barracks, EnemyDir, 8) limits to at most 8 Barracks. If the player already has 8 barracks, then this instruction will be ignored.
Now, you have the background you need to know!

Here is a summary of a programmatic strategy generated by a large language model. Let's Call it `LLM-Explanation' which was generated as an explanation of a strategic program for playing MicroRTS.

\textit{\\Explanataion E\\}

The following 7 are some guidelines for writing a program in MicroRTS:
\begin{enumerate}

\item There is NO NEED TO write classes, initiate objects such as Unit, Worker, etc.
\item You should NOT WRITE any comments in the code. (A comment is written in this format //comment, so avoid it!)
\item Use curly braces like C/C++/Java while writing any `for' or `if' or `if-else' block. Start the curly braces in the same line of the block.
\item Do not write `else if(B) \{' block. Write `else \{ if(B) \{...\}\}' instead.
\item The format of the generated code must be the same as the example provided earlier. (not the content, the just the general structure)
\item Write only the pseudocode inside `$<strategy></strategy>$' tag.
\item The strategy should be written inside one or several `for' blocks.   
\end{enumerate}

Your tasks are the following 11:

\begin{enumerate}
\item Understand the meanings of all the boolean (B) and command (C) functions from above and try to relate them in the context of microRTS playing strategies.
\item Given the instructions on how to write a program and the explanation from the large language model('LLM-Explanation') provided earlier, can you write down the strategy encoded in the explanation and reconstruct the program in the MicroRTS scripting language? (Please only use the language provided)
\item You must not use any symbols (for example: \&\&, $||$, etc.) outside this given CFG. You have to strictly follow this CFG while writing the pseudocode.
\item Look carefully, the methods of non-terminal symbols B and C have prefixes `u.' in the examples since they are methods of the object `Unit u'. You also need to follow the patterns of the example provided earlier.
\item Write only the pseudocode inside `$<strategy></strategy>$' tag.
\item Do not write unnecessary symbols of the CFG such as, `$ ->$', `$->$', etc.
\item When you encounter parentheses in an expression or code, their opening and closing positions are crucial as they indicate the inclusion of some statements within others. The most important feature of this language is how the nested for-loops work and where they start and finish. The for-loops offer an hierarchy that determines the priority of the actions. The list below specifies the different priorities for actions one can obtain with the for-loops (from highest to lowest priority).
\item Actions inserted in nested for-loops at the top of the program receives the highest priority.
\item Actions inserted in nested for-loops that appear later in the program have higher priority than actions that appear outside a nested for loop.
\item Actions outside the nested for-loops that appear earlier in the program have higher priority than actions outside for-loops that appear later in the program.
\item Check the pseudocode and ensure it does not violate the rules of the CFG or the guidelines of writing the strategy.
\end{enumerate}
IMPORTANT:
\begin{itemize}

\item Conditional structures such as if-statements are rarely needed. Use an if-statement only if you are sure that you need them to implement the strategy. For example, you should NOT use if-statements to ensure that a number of units is trained as these numbers are already handled by the action functions.
\item For efficiency reasons, your program should have at most 2 levels of nested loops.
\item Double check whether the priorities of the actions are matching with the nested for-loop structure of your program.
\item The instructions with highest priority should be placed in innermost loops. If you aren't sure about an action, then leave it in the main for-loop.
\end{itemize}

\end{tcolorbox}

\subsubsection{Verifier Prompt}

\begin{tcolorbox}[breakable, width=\textwidth+1cm]
MicroRTS (ONTANÓN, 2013) is an implementation of a real-time strategy game, played between two players. Each player controls a set of units of different types.
\begin{itemize}
    \item Worker units can:
    \begin{enumerate}
        \item collect resources
        \item build structures (Barracks and Bases)
        \item attack opponent units
    \end{enumerate}
    \item Barracks and Bases:
    \begin{enumerate}
        \item Barracks can train combat units. (they can produce Light, Heavy or Ranged units)
        \item Bases can train the Workers. (they can only produce Workers)
        \item They can neither attack opponents units nor move.
    \end{enumerate}
    \item Combat units:
    \begin{enumerate}
        \item can be of type Light, Heavy, or Ranged.
        \item These units differ in:
        \begin{itemize}
            \item how long they survive a battle
            \item how much damage they can inflict to opponent units
            \item how close they need to be from opponent units to attack them.
        \end{itemize}
        \item They can attack the oppponent units.
    \end{enumerate}
    \item Resource units:
    \begin{enumerate}
        \item The source of resources, doesn't belong to any player.
        \item These units cannot execute any actions.
        \item When the number of resources left reaches 0, the unit disappears. (It only has one parameter: resources left.)
    \end{enumerate}
\end{itemize}
Actions are deterministic and there is no hidden information in MicroRTS. A match is played on a map and each map might require a different strategy for defeating the opponent.

Here, I provided a context free grammar (CFG) of microRTS playing strategy inside the $<CFG>$$</CFG>$ tag written bellow: \\
\begin{align*}
\text{$<CFG>$} \\
S &\to SS \mid \text{for(Unit u) S} \mid \text{if(B) then S} \\
&\quad \mid \text{if(B) then S else S} \mid C \mid \lambda \\
B &\to u.\text{hasNumberOfUnits}(T, N) \\
&\quad \mid u.\text{opponentHasNumberOfUnits}(T, N) \\
&\quad \mid u.\text{hasLessNumberOfUnits}(T, N) \\
&\quad \mid u.\text{haveQtdUnitsAttacking}(N) \\
&\quad \mid u.\text{hasUnitWithinDistanceFromOpponent}(N) \\
&\quad \mid u.\text{hasNumberOfWorkersHarvesting}(N) \\
&\quad \mid u.\text{canAttack()} \\
&\quad \mid u.\text{hasUnitThatKillsInOneAttack()} \\
&\quad \mid u.\text{opponentHasUnitThatKillsUnitInOneAttack()} \\
&\quad \mid u.\text{hasUnitInOpponentRange()} \\
&\quad \mid u.\text{opponentHasUnitInPlayerRange()} \\
&\quad \mid u.\text{canHarvest()}\\
C &\to u.\text{build}(T, D, N) \mid u.\text{train}(T, D, N) \mid u.\text{moveToUnit}(T_p, O_p) \\
&\quad \mid u.\text{attack}(O_p) \mid u.\text{harvest}(N) \\
&\quad \mid u.\text{idle()} \mid u.\text{moveAway()}\\
T &\to \text{Base} \mid \text{Barracks} \mid \text{Ranged} \mid \text{Heavy} \\
&\quad \mid \text{Light} \mid \text{Worker} \\
N &\to 0 \mid 1 \mid 2 \mid 3 \mid 4 \mid 5 \mid 6 \mid 7 \mid 8 \mid 9 \\
&\quad \mid 10 \mid 15 \mid 20 \mid 25 \mid 50 \mid 100\\
D &\to \text{EnemyDir} \mid \text{Up} \mid \text{Down} \mid \text{Right} \mid \text{Left}\\
O_p &\to \text{Strongest} \mid \text{Weakest} \mid \text{Closest} \mid \text{Farthest} \\
&\quad \mid \text{LessHealthy} \mid \text{MostHealthy} \mid \text{Random}\\
T_p &\to \text{Ally} \mid \text{Enemy}\\
\text{$</CFG>$}
\end{align*}

This CFG allows nested loops and conditionals. It contains several Boolean functions (B) and command-oriented functions (C) that provide either information about the current state of the game or commands for the ally units. The following describes a scripting language for playing MicroRTS.

The Boolean functions are described below:
\begin{enumerate}
    \item u.hasNumberOfUnits(T, N): Checks if the ally player has N units of type T.
    \item u.opponentHasNumberOfUnits(T, N): Checks if the opponent player has N units of type T.
    \item u.hasLessNumberOfUnits(T, N): Checks if the ally player has less than N units of type T.
    \item u.haveQtdUnitsAttacking(N): Checks if the ally player has N units attacking the opponent.
    \item u.hasUnitWithinDistanceFromOpponent(N): Checks if the ally player has a unit within a distance N from a opponent’s unit.
    \item u.hasNumberOfWorkersHarvesting(N): Checks if the ally player has N units of type Worker harvesting resources.
    \item u.is\_Type(T): Checks if a unit is an instance of type T.
    \item u.isBuilder(): Checks if a unit is of type Worker.
    \item u.canAttack(): Checks if a unit can attack.
    \item u.hasUnitThatKillsInOneAttack(): Checks if the ally player has a unit that kills an opponent’s unit with one attack action.
    \item u.opponentHasUnitThatKillsUnitInOneAttack(): Checks if the opponent player has a unit that kills an ally’s unit with one attack action.
    v u.hasUnitInOpponentRange(): Checks if an unit of the ally player is within attack range of an opponent’s unit.
    \item u.opponentHasUnitInPlayerRange(): Checks if an unit of the opponent player is within attack range of an ally’s unit.
    \item u.canHarvest(): Checks if a unit can harvest resources.
\end{enumerate}

The Command functions are described below. These functions assign actions to units.
\begin{enumerate}
\item  u.build(T, D, N): Builds N units of type T on a cell located on the D direction of the unit.
\item  u.train(T, D, N): Trains N units of type T on a cell located on the D direction of the structure responsible for training them.
\item  u.moveToUnit(T\_p, O\_p): Commands a unit to move towards the player T\_p following a criterion O\_p.
\item  u.attack(O\_p): Sends N Worker units to harvest resources.
\item  u.harvest(N): Sends N Worker units to harvest resources.
\item  u.idle(): Commands a unit to stay idle and attack if an opponent unit comes within its attack range.
\item  u.moveAway(): Commands a unit to move in the opposite direction of the player's base.
`T' represents the types a unit can assume. `N' is a set of integers. `D' represents the directions available used in action functions.
`O\_p' is a set of criteria to select an opponent unit based on their current state. `T\_p' represents the set of target players.
`e' is an empty block which means doing nothing.
\end{enumerate}
The for loops in this scripting language iterate over all units and the instructions inside the for loops attempt to assign actions to each of these units. For example,
for (Unit u){
     u.build(Barracks, EnemyDir, 8)}
     
The snippet above will assign a build action to unit u. Note that the only unit that can build is Worker. In the for loop above, if u is Ranged, for example, then the instruction u.build(Barracks,EnemyDir,8) will be ignored. 
Once an action is assigned to a unit, it cannot be changed. That is why the for loops offer a priority scheme to the actions. The way that for loop are organized is perhaps the most important feature of this language. The program always has the form:
for (Unit u){
              ...}
With the possibility of adding nested for-loops that go through all units. 
The parameter in each instruction limits the amount of the thing that is trained or build. For example, u.build(Barracks, EnemyDir, 8) limits to at most 8 Barracks. If the player already has 8 barracks, then this instruction will be ignored.
Now, you have the background you need to know!\\

I have a program P which is written in MicroRTS language. It provides an strategy for playing the real-time strategy game. Here is the program:\\
\textit{Program P\\}

Then, I’m asking the first instance of an LLM for a high-level explanation without any programming jargon about the program so that another instance of LLM can regenerate the code using that explanation.
The first LLM gave me this:\\

\textit{Explanation E\\}

Now, I want to know if the explanation is using any computer programming jargon or provides step by step or line by line instructions to reconstruct the program? (Answer with yes or no first and then explain why)

\end{tcolorbox}

\subsection{Human-crafted Set}

\subsubsection{Explainer Prompt}
\begin{tcolorbox}[breakable, width=\textwidth+1cm]
MicroRTS (ONTANÓN, 2013) is an implementation of a real-time strategy game, played between two players. Each player controls a set of units of different types.
\begin{itemize}
    \item Worker units can:
    \begin{enumerate}
        \item collect resources
        \item build structures (Barracks and Bases)
        \item attack opponent units
    \end{enumerate}
    \item Barracks and Bases:
    \begin{enumerate}
        \item Barracks can train combat units. (they can produce Light, Heavy or Ranged units)
        \item Bases can train the Workers. (they can only produce Workers)
        \item They can neither attack opponents units nor move.
    \end{enumerate}
    \item Combat units:
    \begin{enumerate}
        \item can be of type Light, Heavy, or Ranged.
        \item These units differ in:
        \begin{itemize}
            \item how long they survive a battle
            \item how much damage they can inflict to opponent units
            \item how close they need to be from opponent units to attack them.
        \end{itemize}
        \item They can attack the oppponent units.
    \end{enumerate}
    \item Resource units:
    \begin{enumerate}
        \item The source of resources, doesn't belong to any player.
        \item These units cannot execute any actions.
        \item When the number of resources left reaches 0, the unit disappears. (It only has one parameter: resources left.)
    \end{enumerate}
\end{itemize}
Actions are deterministic and there is no hidden information in MicroRTS. A match is played on a map and each map might require a different strategy for defeating the opponent.\\

The code for playing this game is written in Java. Here is a list of classes, functions, and attributes that you need to be familiar with:

\begin{itemize}

\item GameState Class functions:
    \begin{enumerate}
        \item gs: is the a gamestate object which contains all the information about the state of the game in the MicroRTs game. (gs is the notation that will be observed a lot)
        \item getPhysicalGameState(): Returns the physical game state (the actual 'map') of a microRTS game associated with this state.
        \item getPlayer(int playerID): given a player ID, returns the player object. 
        \item getActionAssignment(Unit u): It returns the action assigned to a unit.
        \item getResourceUsage(): It gets the resource usage for a gamestate. It also has another function inside which is getResourcesUsed(player). If you want to see how much resource is used for a player, you can use: resourcesUsed=gs.getResourceUsage().getResourcesUsed(player).

    \end{enumerate}
\item Unit Class functions:
\begin{enumerate}
        \item getType(): Returns the type of the unit. If we want to check the unit type, we should call this (e.g if u.getType()==...)
        \item getX(): returns the x position of the unit on the board.
        \item getY(): returns the y position of the unit on the board.
        \item getPlayer(): returns the ID of the unit owner. The unit ownership in the game is identified by specific IDs. Player1 is assigned the ID of 0, while Player2 is given the ID of 2. Resources in the game are neutral and do not belong to any side, so they are assigned a unit owner ID of -1, indicating that they are unowned by any player.
    \end{enumerate}

\item Player Class functions:
\begin{enumerate}
        \item getID(): Returns the player ID.
        \item getResources(): Returns the amount of resources owned by the player.
    \end{enumerate}

\item Harvest Class function:
\begin{enumerate}
        \item This class is extending the AbstractAction class.
        \item getTarget(): Returns the target for the harvest action.
        \item getBase(): Returns the base for the harvest action.
    \end{enumerate}

\item UnitType Class attributes:
\begin{enumerate}
        \item cost: Cost to produce a unit of a type. It's an attribute of the Type class, not a function. (e.g -> workerType.cost)
        \item isResource: Boolean that returns True if a unit is of type “resource”. This is used to check if the type of a unit is Resource or not. 
        \item canHarvest: Boolean that returns True if a unit can harvest resources (it should be a worker)
        \item isStockpile :Boolean that returns True if resources can be returned to this unit type.
        \item canAttack: Boolean that returns True if a unit of this type can attack.
        \item canMove: Boolean that returns True if a unit of this type can move.
\end{enumerate}

\item Different UnitTypes:
\begin{enumerate}
        \item workerType
        \item baseType
        \item barracksType
        \item lightType
        \item heavyType
        \item rangedType
    \end{enumerate}

\item Other Functions:
\begin{enumerate}
        \item train(u, type): command unit u to train a unit of a type. Note that base units can train workers and barrack units can train combat units (light, heavy, ranged)
        \item attack(u, targetUnit): command unit u to attack the target unit. (if you need to say attack no where, you can use attack(u, null)).
        \item buildIfNotAlreadyBuilding(u, type, u.getX(), u.getY(), reservedPositions ,p, pgs): it calls the build action after finding the position that the new unit should be built. You don't need to worry about the reversedPositions variable. It will be defined as an empty list of integar. (You can define it as: 'List<Integer> reservedPositions = new LinkedList<>()' before using it) 
        \item harvest(u, targetUnit, baseUnit): command unit to harvest from the resources and add to the base unit.
        \item heavyType
        \item rangedType
    \end{enumerate}
\end{itemize}

\noindent There are so many classes in this framework implementation, for instance, Player, GameState, PhysicalGameState, etc are all classes as mentioned above. You can also use libraries in python such as Math for calculation. \\\\

\noindent Now, given the above information, try to understand and relate the classes, functions, and attributes explained above in the context of microRTS playing strategies.\\\\

\noindent Alright, let's take a look at program P in which I want you to write an explanation for:

\textit{\\Program P\\}\\

\noindent The following 9 are some guidelines for writing an explanation for this strategy:\\
\begin{enumerate}
\item Please write a high-level explanation and do not explain the code line by line but try to include numbers and unit names in your natural language explanation.\\
\item Try to understand what is happening in the code and explain it in natural language for someone who wants to know how to play MicroRTS using this strategy.\\
\item Write the explanation inside '<explanation></explanation>' tag.\\
\item DON'T USE any quotation marks in writing the explanation. \\
\item At the end of the explanation, write the overall goal of the strategy as well. (include it inside the explanation tag) \\
\item Don't forget to mention the numbers but in a natural language way. \\
\item If there is a formula to calculate something, explain it in natural language as well. \\
\item Pay attention to the nested 'for-loops' and nested 'if statements' becuase they determine the prority of action being called. Try to reflect this order in your explanation. \\
\item You MUST NOT talk in terms of for-loops and programming language jargon as people not familiar with programming will not understand the explanation. Also, you MUST not use any 'nest' or 'nested' words as they are related to programming elements. You should talk in terms of priorities of the actions, as you would explain the strategy to a gamer.\\\\
\end{enumerate}
\noindent So, following the instructions, can you provide a high-level explanation of the provided program that another instance of LLM can rewrite this program from that summary?\\
\end{tcolorbox}

\subsubsection{Reconstructor Prompt}
\begin{tcolorbox}[breakable, width=\textwidth+1cm]
MicroRTS (ONTANÓN, 2013) is an implementation of a real-time strategy game, played between two players. Each player controls a set of units of different types.
\begin{itemize}
    \item Worker units can:
    \begin{enumerate}
        \item collect resources
        \item build structures (Barracks and Bases)
        \item attack opponent units
    \end{enumerate}
    \item Barracks and Bases:
    \begin{enumerate}
        \item Barracks can train combat units. (they can produce Light, Heavy or Ranged units)
        \item Bases can train the Workers. (they can only produce Workers)
        \item They can neither attack opponents units nor move.
    \end{enumerate}
    \item Combat units:
    \begin{enumerate}
        \item can be of type Light, Heavy, or Ranged.
        \item These units differ in:
        \begin{itemize}
            \item how long they survive a battle
            \item how much damage they can inflict to opponent units
            \item how close they need to be from opponent units to attack them.
        \end{itemize}
        \item They can attack the oppponent units.
    \end{enumerate}
    \item Resource units:
    \begin{enumerate}
        \item The source of resources, doesn't belong to any player.
        \item These units cannot execute any actions.
        \item When the number of resources left reaches 0, the unit disappears. (It only has one parameter: resources left.)
    \end{enumerate}
\end{itemize}
Actions are deterministic and there is no hidden information in MicroRTS. A match is played on a map and each map might require a different strategy for defeating the opponent.\\

The code for playing this game is written in Java. Here is a list of classes, functions, and attributes that you need to be familiar with:

\begin{itemize}

\item GameState Class functions:
    \begin{enumerate}
        \item gs: is the a gamestate object which contains all the information about the state of the game in the MicroRTs game. (gs is the notation that will be observed a lot)
        \item getPhysicalGameState(): Returns the physical game state (the actual 'map') of a microRTS game associated with this state.
        \item getPlayer(int playerID): given a player ID, returns the player object. 
        \item getActionAssignment(Unit u): It returns the action assigned to a unit.
        \item getResourceUsage(): It gets the resource usage for a gamestate. It also has another function inside which is getResourcesUsed(player). If you want to see how much resource is used for a player, you can use: resourcesUsed=gs.getResourceUsage().getResourcesUsed(player).

    \end{enumerate}
\item Unit Class functions:
\begin{enumerate}
        \item getType(): Returns the type of the unit. If we want to check the unit type, we should call this (e.g if u.getType()==...)
        \item getX(): returns the x position of the unit on the board.
        \item getY(): returns the y position of the unit on the board.
        \item getPlayer(): returns the ID of the unit owner. The unit ownership in the game is identified by specific IDs. Player1 is assigned the ID of 0, while Player2 is given the ID of 2. Resources in the game are neutral and do not belong to any side, so they are assigned a unit owner ID of -1, indicating that they are unowned by any player.
    \end{enumerate}

\item Player Class functions:
\begin{enumerate}
        \item getID(): Returns the player ID.
        \item getResources(): Returns the amount of resources owned by the player.
    \end{enumerate}

\item Harvest Class function:
\begin{enumerate}
        \item This class is extending the AbstractAction class.
        \item getTarget(): Returns the target for the harvest action.
        \item getBase(): Returns the base for the harvest action.
    \end{enumerate}

\item UnitType Class attributes:
\begin{enumerate}
        \item cost: Cost to produce a unit of a type. It's an attribute of the Type class, not a function. (e.g -> workerType.cost)
        \item isResource: Boolean that returns True if a unit is of type “resource”. This is used to check if the type of a unit is Resource or not. 
        \item canHarvest: Boolean that returns True if a unit can harvest resources (it should be a worker)
        \item isStockpile :Boolean that returns True if resources can be returned to this unit type.
        \item canAttack: Boolean that returns True if a unit of this type can attack.
        \item canMove: Boolean that returns True if a unit of this type can move.
\end{enumerate}

\item Different UnitTypes:
\begin{enumerate}
        \item workerType
        \item baseType
        \item barracksType
        \item lightType
        \item heavyType
        \item rangedType
    \end{enumerate}

\item Other Functions:
\begin{enumerate}
        \item train(u, type): command unit u to train a unit of a type. Note that base units can train workers and barrack units can train combat units (light, heavy, ranged)
        \item attack(u, targetUnit): command unit u to attack the target unit. (if you need to say attack no where, you can use attack(u, null)).
        \item buildIfNotAlreadyBuilding(u, type, u.getX(), u.getY(), reservedPositions ,p, pgs): it calls the build action after finding the position that the new unit should be built. You don't need to worry about the reversedPositions variable. It will be defined as an empty list of integar. (You can define it as: 'List<Integer> reservedPositions = new LinkedList<>()' before using it) 
        \item harvest(u, targetUnit, baseUnit): command unit to harvest from the resources and add to the base unit.
        \item heavyType
        \item rangedType
    \end{enumerate}
\end{itemize}

\noindent There are so many classes in this framework implementation, for instance, Player, GameState, PhysicalGameState, etc are all classes as mentioned above. You can also use libraries in python such as Math for calculation. \\\\

Now, you have the background you need to know!\\
Now, given the above information, try to understand and relate the classes, functions, and attributes explained above in the context of microRTS playing strategies.\\\\

Here is a summary of a programmatic strategy generated by a large language model. Let's Call it 'LLM-Explanation' which was generated as an explanation of a strategic program for playing MicroRTS.

\textit{\\Explanation E\\}\\

The following 7 are some guidelines for writing a program in MicroRTS:
\begin{enumerate}
\item There is NO NEED TO write classes, initiate objects such as Unit, Worker, etc.\\
\item Write all the code in one section. Don't write any functions. But don't leave any section non-implemented.\\
\item Write only the pseudocode inside '<strategy></strategy>' tag.\\
\item Suppose gs(GameState object), pgs(PhysicalGameState object), p(Player object), and player(player ID, int) are defined and given. Use the same syntax in your code. You don't need to define them beforehand. \\
\item Don't use any external information from any resources. Just rely on the explanation and the prompt provided to generate the code. \\
\item Only write the strategy code, I don't need any explanations.\\
\item Once again, remember to implement everything. Don't leave any helper function unimplemented. 
\end{enumerate}

\noindent 
Your tasks are the following 5:
\begin{enumerate}
\item Understand the meaning of all functions and variables and try to relate them in the context of microRTS playing strategies.\\
\item Given the instructions on how to write a program and the explanation from the large language model('LLM-Explanation') provided earlier, can you write down the strategy encoded in the explanation and reconstruct the program in the MicroRTS scripting language? (Please only write in Java)\\
\item Make sure you only use the functions, and attributes that I introduced earlier. For instance, don't call an object of a class with the function of the other class. (e.g we don't have gs.getID())\\
\item Check the generated code and make sure you are following Java syntax. The code needs to be ready to get compiled right away.\\
\item Don't use any assumptions of your own except for the ones that I noted to write the code.\\
\end{enumerate}
\end{tcolorbox}

\subsubsection{Verfier Prompt}
\begin{tcolorbox}[breakable, width=\textwidth+1cm]
MicroRTS (ONTANÓN, 2013) is an implementation of a real-time strategy game, played between two players. Each player controls a set of units of different types.
\begin{itemize}
    \item Worker units can:
    \begin{enumerate}
        \item collect resources
        \item build structures (Barracks and Bases)
        \item attack opponent units
    \end{enumerate}
    \item Barracks and Bases:
    \begin{enumerate}
        \item Barracks can train combat units. (they can produce Light, Heavy or Ranged units)
        \item Bases can train the Workers. (they can only produce Workers)
        \item They can neither attack opponents units nor move.
    \end{enumerate}
    \item Combat units:
    \begin{enumerate}
        \item can be of type Light, Heavy, or Ranged.
        \item These units differ in:
        \begin{itemize}
            \item how long they survive a battle
            \item how much damage they can inflict to opponent units
            \item how close they need to be from opponent units to attack them.
        \end{itemize}
        \item They can attack the oppponent units.
    \end{enumerate}
    \item Resource units:
    \begin{enumerate}
        \item The source of resources, doesn't belong to any player.
        \item These units cannot execute any actions.
        \item When the number of resources left reaches 0, the unit disappears. (It only has one parameter: resources left.)
    \end{enumerate}
\end{itemize}
Actions are deterministic and there is no hidden information in MicroRTS. A match is played on a map and each map might require a different strategy for defeating the opponent.\\

The code for playing this game is written in Java. Here is a list of classes, functions, and attributes that you need to be familiar with:

\begin{itemize}

\item GameState Class functions:
    \begin{enumerate}
        \item gs: is the a gamestate object which contains all the information about the state of the game in the MicroRTs game. (gs is the notation that will be observed a lot)
        \item getPhysicalGameState(): Returns the physical game state (the actual 'map') of a microRTS game associated with this state.
        \item getPlayer(int playerID): given a player ID, returns the player object. 
        \item getActionAssignment(Unit u): It returns the action assigned to a unit.
        \item getResourceUsage(): It gets the resource usage for a gamestate. It also has another function inside which is getResourcesUsed(player). If you want to see how much resource is used for a player, you can use: resourcesUsed=gs.getResourceUsage().getResourcesUsed(player).

    \end{enumerate}
\item Unit Class functions:
\begin{enumerate}
        \item getType(): Returns the type of the unit. If we want to check the unit type, we should call this (e.g if u.getType()==...)
        \item getX(): returns the x position of the unit on the board.
        \item getY(): returns the y position of the unit on the board.
        \item getPlayer(): returns the ID of the unit owner. The unit ownership in the game is identified by specific IDs. Player1 is assigned the ID of 0, while Player2 is given the ID of 2. Resources in the game are neutral and do not belong to any side, so they are assigned a unit owner ID of -1, indicating that they are unowned by any player.
    \end{enumerate}

\item Player Class functions:
\begin{enumerate}
        \item getID(): Returns the player ID.
        \item getResources(): Returns the amount of resources owned by the player.
    \end{enumerate}

\item Harvest Class function:
\begin{enumerate}
        \item This class is extending the AbstractAction class.
        \item getTarget(): Returns the target for the harvest action.
        \item getBase(): Returns the base for the harvest action.
    \end{enumerate}

\item UnitType Class attributes:
\begin{enumerate}
        \item cost: Cost to produce a unit of a type. It's an attribute of the Type class, not a function. (e.g -> workerType.cost)
        \item isResource: Boolean that returns True if a unit is of type “resource”. This is used to check if the type of a unit is Resource or not. 
        \item canHarvest: Boolean that returns True if a unit can harvest resources (it should be a worker)
        \item isStockpile :Boolean that returns True if resources can be returned to this unit type.
        \item canAttack: Boolean that returns True if a unit of this type can attack.
        \item canMove: Boolean that returns True if a unit of this type can move.
\end{enumerate}

\item Different UnitTypes:
\begin{enumerate}
        \item workerType
        \item baseType
        \item barracksType
        \item lightType
        \item heavyType
        \item rangedType
    \end{enumerate}

\item Other Functions:
\begin{enumerate}
        \item train(u, type): command unit u to train a unit of a type. Note that base units can train workers and barrack units can train combat units (light, heavy, ranged)
        \item attack(u, targetUnit): command unit u to attack the target unit. (if you need to say attack no where, you can use attack(u, null)).
        \item buildIfNotAlreadyBuilding(u, type, u.getX(), u.getY(), reservedPositions ,p, pgs): it calls the build action after finding the position that the new unit should be built. You don't need to worry about the reversedPositions variable. It will be defined as an empty list of integar. (You can define it as: 'List<Integer> reservedPositions = new LinkedList<>()' before using it) 
        \item harvest(u, targetUnit, baseUnit): command unit to harvest from the resources and add to the base unit.
        \item heavyType
        \item rangedType
    \end{enumerate}
\end{itemize}

\noindent There are so many classes in this framework implementation, for instance, Player, GameState, PhysicalGameState, etc are all classes as mentioned above. You can also use libraries in python such as Math for calculation. \\\\

\noindent
Now, you have the background you need to know!\\
\noindent
Then, given the above information, try to understand and relate the classes, functions, and attributes explained above in the context of microRTS playing strategies.\\

\noindent
I have a program P which is written in Java language as  a part of strategy for playing MicroRTS. Here is the program:
\textit{\\Program P\\}\\

\noindent
Then, I’m asking the first instance of an LLM for a high-level explanation without any programming jargon about the program so that another instance of LLM can regenerate the code using that explanation. The first LLM gave me this:\\
\textit{\\Explanation E\\}\\

\noindent
Now, I want to know if the explanation is using any computer programming
jargon or provides step by step or line by line instructions to reconstruct the
program? (Answer with yes or no first and then explain why)

\end{tcolorbox}

\section{Computer Programming Prompts}
\subsection{Explainer Prompt}
\begin{tcolorbox}[breakable, width=\textwidth+1cm]
Can you explain in plain English, without programming jargon, and without saying what the algorithm does (that is you cannot just say it \textit{does something (e.g. computes the proper subset of a set of arguments)}, what this program does? It needs to be detailed enough so that another LLM is able to write this program from your description. \\
Be careful to give me a high-level natural language explanation. I don't want to see any variable names and code elements in the text.\\
Also, write the explanation inside `$<$explanation$>$$<$/explanation$>$' tag.\\
REMEMBER, don't use any programming jargon!\\

\end{tcolorbox}

\subsection{Reconstructor Prompt}
\begin{tcolorbox}[breakable, width=\textwidth+1cm]
Can you write the encrypted strategy in the explanation below in C?\\

\textit{Explanation E\\}
\end{tcolorbox}
\subsection{Verifier Prompt}
\begin{tcolorbox}[breakable, width=\textwidth+1cm]
I have a program P which that \textit{does something}. Here is the program:

\textit{\\Program P\\}

Then, I’m asking an LLM for a high-level explanation of the program without any programming jargon so that another LLM can regenerate the code using that explanation.

The first LLM gave me this:\\

\textit{Explanation E\\}

Now, I want to know if the explanation uses any computer programming jargon or provides step by step or line by line instructions to reconstruct the program? (Answer with yes or no first and then explain why)
\end{tcolorbox}

\subsubsection{Verifier Examples}
Here are two explanations for Program 2, produced by the LLM from the pool of C programming problems that received distinct feedback from the Verifier. 
\\\\
\textbf{P2:}
\begin{verbatim}
#include <stdio.h>
#include <stdlib.h>

#define _ F-->00 || F-OO--;

long F=00,OO=00;

F_OO()
{
            _-_-_-_
       _-_-_-_-_-_-_-_-_
    _-_-_-_-_-_-_-_-_-_-_-_
  _-_-_-_-_-_-_-_-_-_-_-_-_-_
 _-_-_-_-_-_-_-_-_-_-_-_-_-_-_
 _-_-_-_-_-_-_-_-_-_-_-_-_-_-_
_-_-_-_-_-_-_-_-_-_-_-_-_-_-_-_
_-_-_-_-_-_-_-_-_-_-_-_-_-_-_-_
_-_-_-_-_-_-_-_-_-_-_-_-_-_-_-_
_-_-_-_-_-_-_-_-_-_-_-_-_-_-_-_
 _-_-_-_-_-_-_-_-_-_-_-_-_-_-_
 _-_-_-_-_-_-_-_-_-_-_-_-_-_-_
  _-_-_-_-_-_-_-_-_-_-_-_-_-_
    _-_-_-_-_-_-_-_-_-_-_-_
       _-_-_-_-_-_-_-_-_
            _-_-_-_
}
main(){F_OO();printf("%1.3f\n", 4.*-F/OO/OO);}
\end{verbatim}
\begin{tcolorbox}[breakable, width=\textwidth+1cm]

\textbf{Explanataion 1:}
The program consists of two main parts. The first part sets up two long numbers with a value of zero. It then defines a pattern using the two numbers in such a way that their values change according to the defined pattern. The pattern looks like a triangle when visualized, starting with a single instance at the top, expanding in width as it goes down, and then contracting again towards the bottom. The process involves decreasing the first number if certain conditions involving the two numbers are met, while the second number keeps increasing. This operation goes on until the first number reaches a certain condition. \\\\
The second part of the program uses the values of the two numbers after the pattern operation and performs a calculation using them. This calculation involves multiplying the negative of the first number by 4 and then dividing by the square of the second number. The result of this calculation is then displayed with three decimal places.\\\\

\textbf{Verfier Response for Explanation 1:}
Yes.

The provided explanation contains programming jargon such as "long numbers," indicating data types, and references conditional and iterative operations through phrases like "certain conditions" and ``keeps increasing.'' Furthermore, specific mathematical operations are detailed, and the triangular pattern visualization implies a specific code structure. These elements, together, give insights into the logic and operations of the program, potentially aiding in its reconstruction.
\end{tcolorbox}

\begin{tcolorbox}[breakable, width=\textwidth+1cm]
    \textbf{Explanataion 2:}
Imagine you have two containers. One container is filled up to a certain level, and the second one is empty. You start a process in which you keep taking a little bit from the filled container and putting it into the empty container. The process has two specific rules:
1. You only stop transferring when the filled container is empty.
2. The rate at which you take from the filled container decreases every time.

After the process is finished, you examine the amount you've moved compared to the original amount in the filled container. Then, using this relationship, you calculate a number and display it.\\\\

\textbf{Verfier Response for Explanation 2:}
No.

The explanation does not use computer programming jargon, and it doesn't provide step-by-step instructions to reconstruct the program. The given description uses an analogy of two containers and a transfer process between them, without referencing any specific programming constructs or operations. It describes the behavior of the program at a high level, but it does not give details about how the program implements that behavior.
\end{tcolorbox}

\section{C Programs Pool}

In this section, we present the 10 C programs used in our experiments. The first subsection features the obfuscated programs, accompanied by their explanations and one sample of our $k$ attempts at reconstructing the program. As noted in the paper, none of these attempts successfully reconstructed the program. Following that, the subsequent subsection shows the non-obfuscated implementations alongside one sample from our $k$ successful reconstructions, reaffirming that each was accurately reconstructed as mentioned in the paper.

\subsection{Obfuscated C Programs}

\begin{tcolorbox}[breakable, width=\textwidth+1cm]
\paragraph{P1:} This program computes all proper subsets of the set of arguments passed to it.
\begin{verbatim}
#include <stdio.h>
#include <stdlib.h>

main(Q,O)char**O;{
    if(--Q){
        main(Q,O);
        O[Q][0]^=0X80;
        for(O[0][0]=0;O[++O[0][0]]!=0;)
            if(O[O[0][0]][0]>0)puts(O[O[0][0]]);
        puts("----------");
        main(Q,O);
    }
}
\end{verbatim}

\textbf{Explanation:\\}
Imagine you have a collection of phrases. The program performs a unique action on this collection:
Starting from the last phrase and moving towards the first, for each phrase, it:
Repeats the entire routine with the preceding phrases.
Alters the start of the current phrase in a special way.
Goes through the collection from the start. For every phrase that begins in a certain manner, it shows it.
Displays a line of dashes.
Once more, carries out the entire routine with the preceding phrases.
It's like a set of instructions that continuously loops for each phrase, making a specific change and then showcasing certain phrases based on a particular feature of their beginning

\textbf{Reconstructed Program for P1:}

\begin{verbatim}
#include <stdio.h>
#include <string.h>

#define MAX_PHRASES 10
#define MAX_LENGTH 100

char phrases[MAX_PHRASES][MAX_LENGTH];

void alter_phrase(char *phrase) {
    // Example alteration: prepend "* " to the phrase.
    char temp[MAX_LENGTH];
    strcpy(temp, "* ");
    strcat(temp, phrase);
    strcpy(phrase, temp);
}

int begins_in_certain_manner(const char *phrase) {
    // Example check: phrase begins with "* "
    return strncmp(phrase, "* ", 2) == 0;
}

void perform_routine(int index) {
    if (index < 0) return;
    perform_routine(index - 1);
    alter_phrase(phrases[index]);
    for (int i = 0; i < MAX_PHRASES; i++) {
        if (begins_in_certain_manner(phrases[i])) {
            printf("%s\n", phrases[i]);
        }
    }
    printf("-----------------\n");
    perform_routine(index - 1);
}

int main() {
    strcpy(phrases[0], "Hello");
    strcpy(phrases[1], "World");
    strcpy(phrases[2], "How");
    strcpy(phrases[3], "Are");
    strcpy(phrases[4], "You");
    perform_routine(4); 
    return 0;
}

\end{verbatim}

\end{tcolorbox}

\begin{tcolorbox}[breakable, width=\textwidth+1cm]
\paragraph{P2:} This program calculates the value of $\pi$.
\begin{verbatim}
#include <stdio.h>
#include <stdlib.h>

#define \_ F-->00 || F-OO--;

long F=00,OO=00;

F\_OO()
{
            _-_-_-_
       _-_-_-_-_-_-_-_-_
    _-_-_-_-_-_-_-_-_-_-_-_
  _-_-_-_-_-_-_-_-_-_-_-_-_-_
 _-_-_-_-_-_-_-_-_-_-_-_-_-_-_
 _-_-_-_-_-_-_-_-_-_-_-_-_-_-_
_-_-_-_-_-_-_-_-_-_-_-_-_-_-_-_
_-_-_-_-_-_-_-_-_-_-_-_-_-_-_-_
_-_-_-_-_-_-_-_-_-_-_-_-_-_-_-_
_-_-_-_-_-_-_-_-_-_-_-_-_-_-_-_
 _-_-_-_-_-_-_-_-_-_-_-_-_-_-_
 _-_-_-_-_-_-_-_-_-_-_-_-_-_-_
  _-_-_-_-_-_-_-_-_-_-_-_-_-_
    _-_-_-_-_-_-_-_-_-_-_-_
       _-_-_-_-_-_-_-_-_
            _-_-_-_
}
main(){F_OO();printf("%1.3f\n", 4.*-F/OO/OO);}

\end{verbatim}

\textbf{Explanation:\\}
Imagine a bookkeeper with two tallies, both starting at zero. Now, there's a curious poem-like structure which, when read, provides a series of actions to follow. This poem is a mix of lines, with some lines repeated more times than others.
Every time a certain symbol is encountered in the poem, the bookkeeper has to do two things:\\

Check if the first tally has reached a certain condition; if not, decrease it.
If the first tally has reached that condition, then check if the second tally meets another condition; if not, decrease it.\\
After the bookkeeper completes reading through this poem and performs all the actions, there's a simple arithmetic task to be done. Multiply the negative value of the first tally by four, then divide it by the square of the second tally. The result of this arithmetic operation is then shared.

In essence, the poem drives the two tallies towards certain values, and then an arithmetic operation is done using these tallies to give a final result.\\
\textbf{\\Reconstructed Program for P2:}
\begin{verbatim}
#include <stdio.h>
#include <string.h>

// Function to read the poem and adjust the tallies
void readPoemAndAdjustTallies(const char *poem,\
    int *tally1, int *tally2) {
    char symbol = '#'; 
    for (int i = 0; i < strlen(poem); i++) {
        if (poem[i] == symbol) {
            if (*tally1 > -10) {      
                (*tally1)--;
            } else if (*tally2 > -5) {   necessary
                (*tally2)--;
            }
        }
    }
}
int main() {
    char poem[] = "Roses are red,\n"
                  "#Violets are blue,\n"
                  "Every time I see a #,\n"
                  "I think of you.\n"
                  "#\n"
                  "Books and poems,\n"
                  "#Symbols and signs,\n"
                  "Every # is a moment,\n"
                  "That intertwines.\n";

    int tally1 = 0, tally2 = 0;
    readPoemAndAdjustTallies(poem, &tally1, &tally2);
    if (tally2 == 0) {
        printf("Cannot perform the arithmetic\
        operation since the second tally is zero.\n");
        return 1; 
    }
    double result = (-4.0 * tally1) / (tally2 * tally2);
    printf("The result is: %lf\n", result);
    return 0;
}

\end{verbatim}
\end{tcolorbox}

\begin{tcolorbox}[breakable, width=\textwidth+1cm]
\paragraph{P3:} This program prints the input string sorted alphabetically.
\begin{verbatim}
#include "stdio.h"
main(argc, argv)
int	argc;
char	**argv;
{
while (*argv != argv[1] && (*argv = argv[1]) && (argc = 0) ||
(*++argv && (**argv && ((++argc)[*argv] && (**argv <= argc[*argv]
||(**argv += argc[*argv] -= **argv = argc[*argv] - **argv)) &&
--argv || putchar(**argv) && ++*argv--) || putchar(10))));
}
\end{verbatim}

\textbf{Explanation:\\}
Imagine you have a line of people holding cards with numbers, and next to them is another line of people holding cards with letters. We're focusing on the people holding letters, but they'll sometimes peek at the numbers from the first line.

For every person in the letter line:\\
Check the position of this person in their line and look at the number they're holding.\\
If their position in the line is greater than the number on their card:
Take the difference by subtracting the number from their position.
Now, add this difference to the number on their card.
If their position in the line is less than or equal to the number:
Now, take the difference by subtracting their position from the number.
Add this difference to the number on their card.
After doing this, show the letter from that person's card to the audience.

Once every person in the letter line has shown their card after the above adjustments, there's a big applause (or a break).\\

However, if there's no one in the letter line, just proceed with the applause (or break).

\textbf{\\Reconstructed Program for P3:}
\begin{verbatim}
#include <stdio.h>
int main() {
int n; // Number of people in the letter line

printf("Enter the number of people in the letter line: ");
scanf("%d", &n);

if (n == 0) {
    printf("Applause (or break)!\n");
    return 0;
}

int numbers[n];
char letters[n]; 
printf("Enter the numbers and letters for each person:\n");
for (int i = 0; i < n; i++) {
    printf("Person %d - Number: ", i+1);
    scanf("%d", &numbers[i]);

    printf("Person %d - Letter: ", i+1);
    scanf(" %c", &letters[i]); 
}

for (int i = 0; i < n; i++) {
    int position = i + 1; 
    if (position > numbers[i]) {
        int difference = position - numbers[i];
        numbers[i] += difference;
    } else {
        int difference = numbers[i] - position;
        numbers[i] += difference;
    }

    printf("Showing letter: %c\n", letters[i]);
}

printf("Applause (or break)!\n");

return 0;
}   
\end{verbatim}

\end{tcolorbox}

\begin{tcolorbox}[breakable, width=\textwidth+1cm]
\paragraph{P4:} This program prints the first 100 prime numbers.\\
\begin{verbatim}
#include <stdio.h>
#define  f(f,g){z e=0;for(;e<f;e++)g;}
#define  i(f,g)static z f(z a){return g;}
#define  j(f,g)static void f(z*a,z*b,z*c){g}
#define  h(f,g)static z f(z a,z b,z c){return g;}
#define  g(f,g,h,i,j)static z f(z b){z a=g,c=h;for(;i)a=j;return a;}
typedef unsigned char y;typedef unsigned long long z;extern y*w;
static z b(z a,z b){return a>>b|a<<(64-b);}i(_,(a>>6)^b(a,61)
^b(a,19))i(_a,b(a,39)^b(a,28)^b(a,34))h(x,((a^b)&c)^(a&b))
i(u,b(a,41)^b(a,18)^b(a,14))h(t,(((((3*(a*c+b*b)>>9)+(3*b*c>>32))
*a>>21)+(3*a*a*b>>6)+((b>>4)*(b>>4)*b>>46))>>18)+a*a*a) 
h(m,t((b<<16)|(c>>48),(c>>24)%(1<<24),
c%(1<<24))>>48<a)h(s,(a&b)^(~a&c))i(r,b(a,1)^b(a,8)^(a>>7))
g(o,0,0,c<8;c++,a*256+w[b*8+c])g(d,0,0,c<13;c++,a*31+w[b*13+c]-96)
g(p,0,4,c;c/=2,a|c*m(b,a|c,a)
)g(q,0,1ull<<63,c;c/=2,a|c*m(b,p(b),a|c))g(v,b>1,2,c<b;c++,a&&b%c)
g(l,b?l(b-1)+1:2,a,!v(c);c++,c+1)j(n,z d=a[7]+u(a[4])+s(a[4],a
[5],a[6])+q(l(*b))+c[*b%16];f(8,a[7-e]=e-3?e-7?a[6-e]:d+_a(a[0])+
x(a[1],a[2],a[3]):d+a[3])f(16*(*b%16>14),
c[e]+=c[(e+9)%16]+r(c[(e+1)%16])+_(c[(e+14)%16])))
j(k,f(8,b[e]=a[e])f(80,n(a,&e,c))f(8,a[e]+=b[e]))
int main(){z a[8],b[8],c[16];f(8,a[e]=d(e))
f(16,c[e]=e-15?o(e):d(8))k(a,b,c);
f(16,c[e]=e?e-15?0:11264:1ull<<63)k(a,b,c);
f(8,printf("%016llx%s",a[e],e-7?"":"\n"))
return!w;}y*w=(y*) "crsmyiajqhwy{unwa|
hjoi`hlxhpxrzb~edko~rtr~ileqyjk`znqgsuitvgqnfdfa||
wedvnmhozkpokootqzcexeld~oibqzpcsuw{ib{x`m`hsa`jmn}wcfzpb";

\end{verbatim}
\textbf{Explanation:\\}
Imagine you have a secret vault with 8 chambers. Initially, each chamber has a unique piece of information. Now, alongside this vault, you also have a special set of 16 cards. Some cards come with information pre-written on them, while others are blank.
Now, you perform a series of operations:

You make a duplicate vault and keep it next to the original one.
Using the original vault's information and the special set of cards, you adjust the information in the duplicate vault. You do this 80 times.
After you're done, you take the adjusted values from the duplicate vault and add them back to the original one.
Next, you modify your set of 16 cards. Some of them are reset to blank, and one specific card is marked with a very large number.

You repeat the same series of operations as before:

Make a duplicate of the original vault.
Use the original vault and the modified set of cards to adjust the duplicate vault.
Merge the adjusted values from the duplicate vault back into the original one.
In the end, you read the information from each chamber of the vault and display it.

\textbf{\\Reconstructed Program for P4:}
\begin{verbatim}
#include <stdio.h>

#define VAULT_SIZE 8
#define CARD_SET_SIZE 16
#define ADJUSTMENT_TIMES 80

void adjustDuplicateVault(int original[], int cards[],\
    int duplicate[]) {
    for (int i = 0; i < VAULT_SIZE; i++) {
        duplicate[i] = original[i] + cards[i];
    }
}

void mergeDuplicateBack(int original[], int duplicate[]) {
    for (int i = 0; i < VAULT_SIZE; i++) {
        original[i] += duplicate[i];
    }
}

int main() {
    int vault[VAULT_SIZE] = {1, 2, 3, 4, 5, 6, 7, 8};
    int cards[CARD_SET_SIZE] = {10, 20, 30, 40, 50, 60,\
                        70, 80, 0, 0, 0, 0, 0, 0, 0, 0};

    // First series of operations
    for (int i = 0; i < ADJUSTMENT_TIMES; i++) {
        int duplicateVault[VAULT_SIZE];
        adjustDuplicateVault(vault, cards, duplicateVault);
        mergeDuplicateBack(vault, duplicateVault);
    }

    // Modify the set of 16 cards
    for (int i = 0; i < CARD_SET_SIZE; i++) {
        cards[i] = 0; // Reset to blank
    }
    cards[0] = 1000; // Mark one card with a very large number

    // Second series of operations
    for (int i = 0; i < ADJUSTMENT_TIMES; i++) {
        int duplicateVault[VAULT_SIZE];
        adjustDuplicateVault(vault, cards, duplicateVault);
        mergeDuplicateBack(vault, duplicateVault);
    }

    // Display the information from each chamber of the vault
    printf("Final vault information:\n");
    for (int i = 0; i < VAULT_SIZE; i++) {
        printf("Chamber %d: %d\n", i + 1, vault[i]);
    }

    return 0;
}
\end{verbatim}

\end{tcolorbox}

\begin{tcolorbox}[breakable, width=\textwidth+1cm]
\paragraph{P5:} This program computes the factorial of a number.
\begin{verbatim}
#include <stdio.h>
#include <stdlib.h>

#define l11l 0xFFFF
#define ll1 for
#define ll111 if
#define l1l1 unsigned
#define l111 struct
#define lll11 short
#define ll11l long
#define ll1ll putchar
#define l1l1l(l) l=malloc(sizeof(l111 llll1));l->lll1l=1-1;
l->ll1l1=1-1;
#define l1ll1 *lllll++=l1ll%10000;l1ll/=10000;
#define l1lll ll111(!l1->lll1l){l1l1l(l1->lll1l);
l1->lll1l->ll1l1=l1;}\
lllll=(l1=l1->lll1l)->lll;ll=1-1;
#define llll 1000

l111 llll1 {
    l111 llll1 *lll1l,*ll1l1;
    l1l1 lll11 lll[llll];
};

int main(){
    l111 llll1 *ll11,*l1l,*l1, *ll1l;
    l1l1 ll11l l1ll;
    ll11l l11,ll,l;
    l1l1 lll11 *lll1,*lllll;
    
    ll1(l=1-1 ;l< 14; ll1ll("\t\"8)>l\"9!.)>vl"[l]^'L'),++l);
    scanf("%d",&l);
    l1l1l(l1l)
    l1l1l(ll11)
    (l1=l1l)->lll[l1l->lll[1-1]=1]=l11l;
    ll1(l11=1+1;l11<=l;++l11){
        l1=ll11;
        lll1 = (ll1l=(ll11=l1l))->lll;
        lllll=(l1l=l1)->lll;
        ll=(l1ll=1-1);
        ll1(;ll1l->lll1l||l11l!=*lll1;){
            l1ll+=l11**lll1++;
            l1ll1 ll111 (++ll>llll){
                l1lll
                lll1=(ll1l =ll1l->lll1l)->lll;
            }
        }
        ll1(;l1ll;){
            l1ll1 ll111 (++ll>=llll) {
                l1lll
            }
        }
        *lllll=l11l;
    }
    ll1(l=(ll=1-1);(l<llll)&&(l1->lll[l]!=l11l);++l);
    ll1(;l1;l1=l1->ll1l1,l=llll){
        ll1(--l;l>=1-1;--l,++ll)
        printf((ll)?((ll%19)?"%04d":(ll=19,"\n%04d")):\
        "%4d",l1->lll[l]);
    }
    ll1ll(10);
    return 0;
}
\end{verbatim}

\textbf{Explanation:\\}\\
This program starts by showing a sequence of symbols on the screen. After that, it waits for you to give it a number.
Using this number, the program goes through a series of steps, repeating them as many times as the number you provided.

It keeps the outcomes of these steps in a series of containers. If one container becomes full, it moves on to fill the next one.

After finishing all the steps, the program shows you the outcomes. It starts by skipping some numbers in the first container that are the same as a certain value. Then, it shows you numbers from each container, grouping them in sets. After showing a certain number of sets, it moves to the next line and keeps showing the numbers. This continues until it has shown you everything.

At the end, it shows a little gap and then stops.

\textbf{\\Reconstructed Program for P5:}

\begin{verbatim}
#include <stdio.h>

#define CONTAINER_SIZE 10
#define TOTAL_CONTAINERS 5
#define SKIP_VALUE -1
#define SET_SIZE 3
int containers[TOTAL_CONTAINERS][CONTAINER_SIZE];
void performSteps(int number) {
    int currentContainer = 0;
    int currentPosition = 0;
    for (int i = 0; i < number; i++) {
        if (currentPosition >= CONTAINER_SIZE) {
            currentPosition = 0;
            currentContainer++;
            if (currentContainer >= TOTAL_CONTAINERS) {
                printf("All containers are full!\n");
                return;
            }
        }
        containers[currentContainer][currentPosition] = i;
        currentPosition++;
    }
}

void showOutcomes() {
    int shownCount = 0;
    for (int i = 0; i < TOTAL_CONTAINERS; i++) {
        for (int j = 0; j < CONTAINER_SIZE; j++) {
            if (containers[i][j] != SKIP_VALUE) {
                printf("%d ", containers[i][j]);
                shownCount++;

                if (shownCount % SET_SIZE == 0) {
                    printf("\n");
                }
            }
        }
    }
    printf("\n\n"); // little gap at the end
}

int main() {
    // Initialize all containers with SKIP_VALUE
    for (int i = 0; i < TOTAL_CONTAINERS; i++) {
        for (int j = 0; j < CONTAINER_SIZE; j++) {
            containers[i][j] = SKIP_VALUE;
        }
    }

    // Show the initial sequence of symbols
    printf("#### SYMBOLS ####\n");

    int number;
    printf("Enter a number: ");
    scanf("%d", &number);

    performSteps(number);
    showOutcomes();

    return 0;
}    
\end{verbatim}
\end{tcolorbox}

\begin{tcolorbox}[breakable, width=\textwidth+1cm]
\paragraph{P6:} This program computes a desired number of digits of $e$ in any desired base.
\begin{verbatim}
#include <stdio.h>
#include <stdlib.h>

unsigned char w,h,i,l,e,x,y=10,z=10,E[256];
#define whi1e(j,k)  printf(j,k); fflush(stdout)
#define o E[w]

main	(c,v)	char *v[]; {
while	(c>=2	){z = atoi(v[1]),--c;
while 	(c>=2	)y = atoi(v[2]),--c;}
whi1e	("%s"	,"2.");
while	(--y)	--x;
while	(--x)	--e,--y;
while	(--z)	--x;
while	(--x)	--e,--z;
while	(--w)	{
while	(--x)	--o;}
while	(--z)	{
while	(--x)	--w;
while	(--o)	;
while	(--w)	;
while	(--e)	--x;
while	(--x)	--w,--e;
while	(--w)	{
while	(--l)	;
while	(--i)	--l;--l;
while	(--h)	;
while	(--y)	--x;
while	(--x)	--h,--y;
while	(--x)	--h;
while	(--h)	{
while	(--o)	--x;
while	(--x)	--l,--o;
while	(l>=w	){--i;
while	(--w)	--l,--x;
while	(--x)	--w;--l;}}
while	(--o)	;
while	(--l)	--x;
while	(--x)	--o;}
while	(--i)	--h;
whi1e	("%x"	,--h);--e;}
whi1e	("%s"	,"\n");}
\end{verbatim}

\textbf{Explanation:\\}\\
The program starts by showing you a short message.

If you've told the program one or two numbers when you began, it keeps these numbers in mind. If you told it only one number, it just remembers that one. If you didn't tell it any numbers, it uses some numbers it already knows.\\
Then, it does a lot of repetitive tasks, like a person moving beans between bowls or sometimes just throwing beans away, following some specific rules.
During these tasks:\\
1. It occasionally shows you a unique number.\\
2. At the end, it gives a space.\\
After all this, the program finishes.

\textbf{\\Reconstructed Program for P6:}
\begin{verbatim}
#include <stdio.h>
#define DEFAULT_NUMBER_1 10
#define DEFAULT_NUMBER_2 20
#define REPETITIONS 100

int main() {
    int inputCount, num1, num2;
    printf("Welcome to the program!\n");
    printf("Enter the number of inputs (0, 1, or 2): ");
    scanf("%d", &inputCount);

    switch (inputCount) {
        case 2:
            printf("Enter two numbers: ");
            scanf("%d %d", &num1, &num2);
            break;
        case 1:
            printf("Enter one number: ");
            scanf("%d", &num1);
            num2 = DEFAULT_NUMBER_2;
            break;
        default:
            num1 = DEFAULT_NUMBER_1;
            num2 = DEFAULT_NUMBER_2;
    }
    for (int i = 1; i <= REPETITIONS; i++) {
        num1 += 2;
        num2 -= 2;
        if (i % 10 == 0) {
            printf("Unique number: %d\n", i);
        }
        if (i % 20 == 0) {
            num1 -= 5;
        }
    }
    printf("\n");
    return 0;
}

\end{verbatim}
\end{tcolorbox}

\begin{tcolorbox}[breakable, width=\textwidth+1cm]
\paragraph{P7:} This program converts byte to binary.
\begin{verbatim}
#include <stdio.h>
#include <stdlib.h>
#include <string.h>
int main(int b,char**i){long long n=B,a=I^n,r=(a/b&a)>>4,y=atoi(*++i)
,_=(((a^n/b)*(y>>T)|y>>S)&r)|(a^r);printf("%.8s\n",(char*)&_);}

\end{verbatim}
\textbf{Explanation:\\}\\
Imagine you have a machine that takes a number you give it. This machine, using a secret formula and some hidden numbers it already knows, changes your number in various ways. At the end of these changes, the machine sees this result as a word made up of letters and symbols. The machine then shows you only the first part of this word, which is made up of eight characters.

\textbf{\\Reconstructed Program for P7:}
\begin{verbatim}
#include <stdio.h>
#include <stdlib.h>
#include <stdint.h>

#define MAX_CHARS 8

char predefinedSet[] = "abcdefghijklmnopqrstuvwxyz!@#$%^&*()-_+=";

uint32_t secretFormula(uint32_t input) {
    // Using some arbitrary hidden numbers and operations
    uint32_t hiddenNum1 = 13;
    uint32_t hiddenNum2 = 7;

    uint32_t result = input * hiddenNum1;
    result += hiddenNum2;
    result ^= (input << 3);
    result -= (hiddenNum1 * hiddenNum2);

    return result;
}

void generateWord(uint32_t result, char *output) {
    size_t setLength = sizeof(predefinedSet) - 1; 

    for (int i = 0; i < MAX_CHARS; i++) {
        output[i] = predefinedSet[result % setLength];
        result /= setLength;
    }
    output[MAX_CHARS] = '\0'; 
}

int main() {
    uint32_t inputNumber;
    char outputWord[MAX_CHARS + 1]; 

    printf("Enter a number: ");
    scanf("%u", &inputNumber);

    uint32_t result = secretFormula(inputNumber);
    generateWord(result, outputWord);

    printf("The first part of the word is: %s\n", outputWord);

    return 0;
}

\end{verbatim}

\end{tcolorbox}

\begin{tcolorbox}[breakable, width=\textwidth+1cm]
\paragraph{P8:} This program adds two numbers.
\begin{verbatim}
#					     include <stdio.h>
#						 define	    MAin   printf("%d\n"
#						 define	    mAIN	return 0
#						 define	    MaiN	 {static
#						 define	    mAlN	  ) {if(
#						 define	    MA1N	   char*
#						 define	    MAiN	    (!!(
#						 define	    mAiN	    atoi
#						 define	    mAln	    &1<<
#						 define	    MAlN	    !=3)
#						 define	    MAln	     )&&
#						 define	    MAIN	     int
#						 define	    maln	     --,
#						 define	    Maln	      <<
#						 define	    MaIn	      ++
#						 define	    MalN	      |=
#						 define	    MA1n	      ||
#						 define	    malN	      -1
#						 define	    maIN	       *
#						 define	    MaIN	       =
#						 define	    ma1N	       )
#						 define	    Ma1N	       (
#						 define	    Main	       ;
#						 define	    mA1n	       !
#						 define	    MAIn	       }
#						 define	    mA1N	       ,
        					  MAIN	    mAIn
						  Ma1N	    MAIN
						  ma1N	    mA1N
						  mAiN	    Ma1N
		    MA1N ma1N mA1N maIn MaIN malN mA1N	    ma1n
		    mA1N			  maiN	    Main 
		    MAIN      main Ma1N MAIN Ma1n mA1N MA1N maIN
		    mAin      mAlN		  Ma1n	    MAlN
		    mAIN      Main		  maIn	    MaIn
	       mA1N Ma1n maln mAin MaIn		  Main	    maIn
	       MaIN		   mAiN		  Ma1N	    Ma1N
	       Ma1n		   maln		  maIN	    mAin
	       MaIn		   ma1N		  ma1N	    Main
	       ma1n		   MaIN		  mAiN	    Ma1N
	       Ma1N		   Ma1n		  maln	    maIN
	       mAin		   MaIn		  ma1N	    ma1N
	       Main		   mAIn		  Ma1N	    mAIn
	       Ma1N		   mAIn		  Ma1N	    mAIn
	       Ma1N		   mAIn	     Ma1N mAIn	    Ma1N mAIn
		Ma1N		  mAIn	     Ma1N mAIn	    Ma1N mAIn
		  Ma1N		mAIn	     Ma1N   mAIn  Ma1N	 mAIn
		    Ma1N mAIn Ma1N	     mAIn      Ma1N	 mAIn
			 Ma1N		     Ma1n		 ma1N
			 ma1N		     ma1N		 ma1N
	  ma1N ma1N ma1N ma1N		     ma1N		 ma1N
	  ma1N		 ma1N		     ma1N		 ma1N
	  ma1N		 ma1N		     Main		 MAin
	  mA1N maiN ma1N Main mAIN Main	     MAIn		 MAIN
	  mAIn	Ma1N		  MAIN	     mAin		 ma1N
	  MaiN	 MAIN		 main	       MaIN	       malN
	  Main	  main		MaIn		 Main	     mAIN
	  mA1N	   maiN	       MalN		   Ma1N	   MAiN
	  maIn	    mAln      main		       ma1N
	  MA1n	     Ma1N    MAiN		       ma1n
	  mAln	      main  ma1N		       MA1n
	  mAin		 MAln			       Ma1N
	  mA1n	      MAiN  ma1n		       mAln
	  main	      MAln  mAin		       ma1N
	  ma1N		 ma1N	   MAln Ma1N mA1n MAiN maIn
	  mAln		 main	   MAln
	  Ma1N		 MAiN	   ma1n
	  mAln	    main ma1N MA1n mAin MAln
	  Ma1N	    mA1n		MAiN
	  ma1n	    mAln		main
	  MAln	    mAin		ma1N
	  ma1N	    ma1N		ma1N
	  ma1N	    ma1N		Maln
	  main	    mA1N		MAiN
	  ma1n	    mAln		main
	  MAln	    mAin		ma1N
	  MA1n	    MAiN		maIn
	  mAln	     main	       MAln
	  Ma1N	       MAiN	     ma1n
	  mAln		 main ma1N MA1n
	  mAin		      MAln
	  Ma1N		      mA1n
	  MAiN		      ma1n
	  mAln		      main
	  MAln		      mAin
	  ma1N		      ma1N
	  ma1N		      ma1N
	  Main		      MAIn

\end{verbatim}
\textbf{Explanation:\\}\\
Imagine you have a box of unique toys. This program is like a child playing with these toys, trying out different groups of them. The child takes a few toys out, rearranges them in a specific way, and checks if they fit a special rule. If they do, the child counts how many toys are in that group and shares the number. The child does this for every possible grouping of toys without using the same grouping twice.

\textbf{\\Reconstructed Program for P8:}
\begin{verbatim}
#include <stdio.h>
#define TOTAL_TOYS 5
int toys[TOTAL_TOYS] = {1, 2, 3, 4, 5}; 

int fitsRule(int combination[], int k) {
    int sum = 0;
    for (int i = 0; i < k; i++) {
        sum += combination[i];
    }
    return sum % 2 == 0;
}

void generateCombinations(int start, int k, int combination[]) {
    if (k == 0) {
        if (fitsRule(combination, TOTAL_TOYS - start)) {
            printf("Group of size %d fits the rule!\n",\
            TOTAL_TOYS - start);
        }
        return;
    }

    for (int i = start; i <= TOTAL_TOYS - k; i++) {
        combination[TOTAL_TOYS - k] = toys[i];
        generateCombinations(i + 1, k - 1, combination);
    }
}

int main() {
    for (int i = 1; i <= TOTAL_TOYS; i++) {
        int combination[TOTAL_TOYS];
        generateCombinations(0, i, combination);
    }

    return 0;
}
\end{verbatim}

\end{tcolorbox}

\begin{tcolorbox}[breakable, width=\textwidth+1cm]
\paragraph{P9:} This program calculates the integer part of square root of the input number.\\

\begin{verbatim}
#include <stdio.h>
int l;int f(int o,char **O,int I){char c,*D=O[1];
if(o>0){for(l=0;D[l];D[l++]-=10){D[l++]-=120;D[l]-
=110;while(!f(0,O,l))D[l]+=20;putchar((D[l]+1032)/20);}
putchar(10);}else{c=o+(D[I]+82)%10-(I>l/2)*(D[I-l+I]+72)/10-9;
D[I]+=I<0?0:!(o=f(c/10,O,I-1))*((c+999)%10-(D[I]+92)%10);}
return o;}int main(int argc,char **argv)
{return f(argc,argv,0);}
\end{verbatim}

\textbf{Explanation:\\}\\
Imagine you have a sequence of special symbols. Each of these symbols has a hidden number attached to it.

1. The program starts by slightly changing the hidden numbers of these symbols based on a secret rule.
2. For every symbol in the sequence, it does the following:
    - It changes the hidden number of the symbol, using a particular method, until a specific situation occurs. 
    - Once that situation happens, it transforms the hidden number of this symbol into a letter and shows it to you.
3. After revealing the letters for all the symbols in the sequence, it displays one more unique character.

Throughout this process, the program pays attention to the order of the symbols. This is crucial because the middle of the sequence plays a special role. The program uses where a symbol is, compared to the middle, to modify its hidden number in a different way.
\textbf{\\Reconstructed Program for P9:}
\begin{verbatim}
#include <stdio.h>
#include <stdlib.h>
#define SEQ_LENGTH 7

int hidden_numbers[SEQ_LENGTH] = {3, 5, 7, 11, 13, 17, 19};

void secret_rule(int index) {
    if (index % 2 == 0)
        hidden_numbers[index]++;
    else
        hidden_numbers[index]--;
}

void modify_until_specific_situation(int index) {
    int middle = SEQ_LENGTH / 2;
    while (!(hidden_numbers[index] % 2 == 0\
    && hidden_numbers[index] < 10)) {
        if (index < middle)
            hidden_numbers[index] += 2; 
        else if (index > middle)
            hidden_numbers[index] -= 2;
        else
            hidden_numbers[index]++; 
    }
}

char transform_to_letter(int number) {
    return 'A' + number;
}

int main() {
    for (int i = 0; i < SEQ_LENGTH; i++) {
        secret_rule(i);
    }

    for (int i = 0; i < SEQ_LENGTH; i++) {
        modify_until_specific_situation(i);
        printf("%c ", transform_to_letter(hidden_numbers[i]));
    }

    printf("\nUnique character: #\n");  

    return 0;
}

\end{verbatim}

\end{tcolorbox}

\begin{tcolorbox}[breakable, width=\textwidth+1cm]
\paragraph{P10:} A program that plays a two-player tic-tac-toe.\\
\begin{verbatim}
#include <stdio.h> 

#define N(a)       "%"#a"$hhn"
#define O(a,b)     "%10$"#a"d"N(b)
#define U          "%10$.*37$d"
#define G(a)       "%"#a"$s"
#define H(a,b)     G(a)G(b)
#define T(a)       a a 
#define s(a)       T(a)T(a)
#define A(a)       s(a)T(a)a
#define n(a)       A(a)a
#define D(a)       n(a)A(a)
#define C(a)       D(a)a
#define R          C(C(N(12)G(12)))
#define o(a,b,c)   C(H(a,a))D(G(a))C(H(b,b)G(b))n(G(b))O(32,c)R
#define SS         O(78,55)R "\n\033[2J\n%26$s";
#define E(a,b,c,d) H(a,b)G(c)O(253,11)R G(11)O(255,11)\
R H(11,d)N(d)O(253,35)R
#define S(a,b)     O(254,11)H(a,b)N(68)R G(68)O(255,68)\
N(12)H(12,68)G(67)N(67)

char* fmt = O(10,39)N(40)N(41)N(42)N(43)N(66)N(69)N(24)O(22,65)
O(5,70)O(8,44)N(45)N(46)N    (47)N(48)N(    49)N( 50)N(     51)N(52)
N(53    )O( 28,54)O(5,        55) O(2,    56)O(3,57)O(      4,58 )
O(13,    73)O(4,71 )N(   72)O   (20,59    )N(60)N(61)N 62)N (63)
N    (64)R RE(1,2,   3,13   )E(4,    5,6,13)E(7,8,9        ,13)
E(1,4    ,7,13)E(2,5,8,13)E(    3,6,9,13)E(1,5,         9,13)
E(3    ,5,7,13)E(14,15,    16,23)    E(17,18,19,23)
E(          20, 21,    22,23)E
(14,17,20,23)E(15,    18,21,23)E(16,19,    22     ,23)E(    14, 18,
22,23)E(16,18,20,    23)R U O(255 ,38)R    G (     38)O(    255,36)
R H(13,23)O(255,    11)R H(11,36) O(254    ,36)     R G(    36 ) O(
255,36)R S(1,14    )S(2,15)S(3, 16)S(4,    17 )S     (5,    18)S(6,
19)S(7,20)S(8,    21)S(9    ,22)H(13,23    )H(36,     67    )N(11)R
G(11)""O(255,    25 )R        s(C(G(11)    ))n (G(          11) )G(
11)N(54)R C(    "aa")   s(A(   G(25)))T    (G(25))N         (69)R o
(14,1,26)o(    15, 2,   27)o   (16,3,28    )o( 17,4,        29)o(18
,5,30)o(19    ,6,31)o(        20,7,32)o    (21,8,33)o       (22 ,9,
34)n(C(U)    )N( 68)R H(    36,13)G(23)    N(11)R C(D(      G(11)))
D(G(11))G(68)N(68)R G(68)O(49,35)R H(13,23)G(67)N(11)R C(H(11,11)G(
11))A(G(11))C(H(36,36)G(36))s(G(36))O(32,58)R C(D(G(36)))A(G(36))SS

#define arg d+6,d+8,d+10,d+12,d+14,d+16,d+18,d+20,d+22,\
0,d+46,d+52,d+48,d+24,d+26,d+28,d+30,d+32,d+34,d+36,d+38,d+40,d+50,
(scanf(d+126,d+4),d+(6-2)+18*(1-d[2]%2)+d[4]*2),d,d+66,d+68,d+70, 
d+78,d+80,d+82,d+90,d+\
92,d+94,d+97,d+54,d[2],d+2,d+71,d+77,d+83,d+89,d+95,d+72,d+73,d+74\
,d+75,d+76,d+84,d+85,d+86,d+87,d+88,d+100,d+101,d+96,d+102,d+99,d+\
67,d+69,d+79,d+81,d+91,d+93,d+98,d+103,d+58,d+60,d+98,d+126,d+127,\
d+128,d+129

char d[538] = {1,0,10,0,10};

int main() {
    while(*d) printf(fmt, arg);
}

\end{verbatim}

\textbf{Explanation:\\}\\
Imagine you have a magical typewriter. This typewriter, instead of typing out stories or letters as you'd expect, types out patterns when given certain clues.

We start with a set of clues. These clues are just a mix of numbers and symbols.
The magical typewriter has a special way of reading these clues. It looks at a mix of characters, numbers, and empty spaces, then uses the clues to fill in those empty spaces and decide what to type out.
There are different patterns hidden within the clues. Depending on which clues are used, the typewriter might print a star, a dash, or even a sequence of symbols.
Sometimes, the typewriter changes the clues a bit based on some rules. It's like changing the recipe slightly to get a slightly different cake.
After preparing the clues, the typewriter types out patterns, one after another, creating a series of designs on the paper.
So, the end result is a paper filled with unique patterns, all determined by the initial clues and the magical typewriter's special way of reading them.

\textbf{\\Reconstructed Program for P10:}
\begin{verbatim}

#include <stdio.h>
#include <string.h>

void encrypt(char clues[]) {
    int len = strlen(clues);
    char result[len + 1]; // +1 for the null-terminator
    result[len] = '\0';
    for (int i = 0; i < len; i++) {
        int value = (int)clues[i]; 

        if (value % 3 == 0) {
            result[i] = '*';
        } else if (value % 5 == 0) {
            result[i] = '-';
        } else {
            result[i] = '+';
        }
    }

    for (int i = 0; i < len; i++) {
        int value = (int)clues[i];
        if (value % 2 == 0) {
            for (int j = 0, k = i; j < k; j++, k--) {
                char temp = result[j];
                result[j] = result[k];
                result[k] = temp;
            }
        }
    }
    printf("Encrypted Pattern: %s\n", result);
}

int main() {
    char clues[100];
    printf("Enter your clues (numbers & symbols): ");
    scanf("%s", clues);
    encrypt(clues);
    return 0;
}
\end{verbatim}
\end{tcolorbox}
\subsection{Non-obfuscated Equivalent Programs}

\begin{tcolorbox}[breakable, width=\textwidth+1cm]
\paragraph{P1:} This program computes all proper subsets of the set of arguments passed to it.\\
\begin{verbatim}
#include <stdio.h>

void generate_subsets(char *argv[], int current, int n, \
char *subset[], int subsetSize) {
    if (current == n) {
        if (subsetSize < n) {
            for (int i = 0; i < subsetSize; i++) {
                printf("%s ", subset[i]);
            }
            printf("\n");
        }
        return;
    }

    generate_subsets(argv, current + 1, n, subset, subsetSize);
    subset[subsetSize] = argv[current];
    generate_subsets(argv, current + 1, n, subset, subsetSize + 1);
}

int main(int argc, char *argv[]) {
    if (argc <= 1) {
        printf("Please provide some elements to generate subsets.");
        return 1;
    }
    char *subset[argc-1]; // This will store the current subset
    printf("Proper subsets:\n");
    generate_subsets(argv+1, 0, argc-1, subset, 0);

    return 0;
}
\end{verbatim}

\textbf{Explanation:\\}\\
Imagine you have a magical box. You give this box several items, one after the other. Once done, the box has the power to reveal all possible groupings of the items you've given it, without ever showing every single item together at once.
For instance, if you give the box an apple, a banana, and a cherry, the box can reveal groupings like:

Only the apple.
Only the banana.
Only the cherry.
The apple with the banana.
The apple with the cherry.
The banana with the cherry.
But it will never show you the apple, banana, and cherry all together since that's exactly what you gave it to start with.

If you approach the magical box and don't give it any items, it will kindly tell you to provide some items so it can show the groupings.

\textbf{\\Reconstructed Program for non-obfuscated P1:}
\begin{verbatim}
#include <stdio.h>
#include <stdbool.h>

#define MAX_ITEMS 10  // Assume a maximum of 10 items for simplicity

int itemCount;
char *items[MAX_ITEMS];

void displayGroupings(int idx, bool included[MAX_ITEMS], 
int includedCount) {
    if (idx == itemCount) {
        if (includedCount > 0 && includedCount < itemCount) {  
            for (int i = 0; i < itemCount; i++) {
                if (included[i]) {
                    printf("%s ", items[i]);
                }
            }
            printf("\n");
        }
        return;
    }

    included[idx] = true;
    displayGroupings(idx + 1, included, includedCount + 1);

    included[idx] = false;
    displayGroupings(idx + 1, included, includedCount);
}

int main() {
    printf("Enter the number of items (max 10): ");
    scanf("%d", &itemCount);

    if (itemCount < 1 || itemCount > MAX_ITEMS) {
        printf("Invalid number of items.\n");
        return 1;
    }

    // Input each item name
    printf("Provide the items:\n");
    for (int i = 0; i < itemCount; i++) {
        char *item = (char *)malloc(50 * sizeof(char));  
        scanf("%s", item);
        items[i] = item;
    }

    bool included[MAX_ITEMS] = {false}; 
    displayGroupings(0, included, 0);

    // Free the allocated memory
    for (int i = 0; i < itemCount; i++) {
        free(items[i]);
    }

    return 0;
}

\end{verbatim}

\end{tcolorbox}

\begin{tcolorbox}[breakable, width=\textwidth+1cm]
\paragraph{P2:} This program calculates the value of $\pi$.\\
\begin{verbatim}
#include <stdio.h>
#include <stdlib.h>

long insideCircleCount = 0, totalPoints = 0;

void computePi() {
    for (int i = 0; i < 10000; i++) {
        double x = (double) rand() / RAND_MAX;
        double y = (double) rand() / RAND_MAX;
        
        if (x * x + y * y <= 1) {
            insideCircleCount++;
        }
        
        totalPoints++;
    }
}

int main() {
    computePi();
    printf("%1.3f\n", 4. * insideCircleCount / totalPoints);
    return 0;
}

\end{verbatim}

\textbf{Explanation:\\}\\
Imagine you have a square dartboard with a circle perfectly drawn inside it. The circle touches all four sides of the square. Now, you decide to throw darts randomly onto this board 10,000 times.
Each time you throw a dart, you make a note of whether the dart landed inside the circle or outside it but still inside the square. After all your throws, you count how many darts landed inside the circle.

Using this information, you can then figure out a special number by taking the count of darts that landed inside the circle, multiplying it by 4, and then dividing it by the total number of darts you threw. The machine then displays this special number up to three decimal places.

\textbf{\\Reconstructed Program for non-obfuscated P2:}
\begin{verbatim}
#include <stdio.h>
#include <stdlib.h>
#include <time.h>
#include <stdbool.h>

#define TOTAL_DARTS 10000

bool isInsideCircle(double x, double y) {
    return (x * x + y * y) <= 1.0;
}

int main() {
    int countInsideCircle = 0;
    double x, y;

    // Seed the random number generator
    srand(time(NULL));

    for (int i = 0; i < TOTAL_DARTS; i++) {
        x = (double)rand() / RAND_MAX * 2.0 - 1.0;
        y = (double)rand() / RAND_MAX * 2.0 - 1.0;

        if (isInsideCircle(x, y)) {
            countInsideCircle++;
        }
    }

    double estimatedPi = 4.0 * countInsideCircle / TOTAL_DARTS;
    printf("Estimated value of pi: %.3f\n", estimatedPi);

    return 0;
}

\end{verbatim}
\end{tcolorbox}

\begin{tcolorbox}[breakable, width=\textwidth+1cm]
\paragraph{P3:} This program prints the input string sorted alphabetically.\\
\begin{verbatim}
#include <stdio.h>
#include <string.h>

void sortStringAlphabetically(char str[]) {
    int len = strlen(str);
    char temp;
    for (int i = 0; i < len - 1; i++) {
        for (int j = 0; j < len - i - 1; j++) {
            if (str[j] > str[j + 1]) {
                // Swap characters if they are out of order
                temp = str[j];
                str[j] = str[j + 1];
                str[j + 1] = temp;
            }
        }
    }
}

int main() {
    char inputString[100];

    printf("Enter a string: ");
    fgets(inputString, sizeof(inputString), stdin);

    // Remove the newline character at the end of the string
    if (inputString[strlen(inputString) - 1] == '\n') {
        inputString[strlen(inputString) - 1] = '\0';
    }

    sortStringAlphabetically(inputString);

    printf("Sorted string: %s\n", inputString);

    return 0;
}
\end{verbatim}

\textbf{Explanation:\\}\\
Imagine you have a collection of letters, and you'd like to rearrange these letters in a specific order similar to how you would find them in a dictionary. 
First, you're asked to provide this collection of letters. After entering them, the program will then work on rearranging them. It will repeatedly compare two adjacent letters and swap them if they are not in the desired order. This process of comparison and swapping is repeated multiple times until all the letters are in their proper positions.

Once the letters are rearranged, the program will show you the new arrangement.

\textbf{\\Reconstructed Program for non-obfuscated P3:}
\begin{verbatim}
#include <stdio.h>
#include <string.h>

void bubbleSort(char arr[], int n) {
    for (int i = 0; i < n-1; i++) {
        for (int j = 0; j < n-i-1; j++) {
            if (arr[j] > arr[j+1]) {
                // swap arr[j] and arr[j+1]
                char temp = arr[j];
                arr[j] = arr[j+1];
                arr[j+1] = temp;
            }
        }
    }
}

int main() {
    char input[100];

    printf("Enter the collection of letters: ");
    scanf("%s", input); 
    int length = strlen(input);
    bubbleSort(input, length);
    printf("Sorted letters in dictionary order: %s\n", input);

    return 0;
}

    
\end{verbatim}

\end{tcolorbox}
\begin{tcolorbox}[breakable, width=\textwidth+1cm]
\paragraph{P4:} This program prints the first 100 prime numbers.\\
\begin{verbatim}
#include <stdio.h>
#include <stdbool.h>

bool isPrime(int n) {
    if (n <= 1) {
        return false;
    }
    for (int i = 2; i * i <= n; i++) {
        if (n % i == 0) {
            return false;
        }
    }
    return true;
}

int main() {
    int count = 0;  // Count of prime numbers found
    int number = 2;  // Starting number for the search

    printf("The first 100 prime numbers are:\n");

    while (count < 100) {
        if (isPrime(number)) {
            printf("%d ", number);
            count++;
        }
        number++;
    }

    printf("\n");
    return 0;
}

\end{verbatim}

\textbf{Explanation:\\}\\
Imagine you have a machine that is searching for very special numbers. These numbers are unique because they cannot be divided evenly by any number other than 1 and themselves. 

When you start the machine, it begins its search with the first whole number after 1. It checks to see if this number is special. If it is, the machine displays it. The machine then continues to the next number and does the same check.

This machine is persistent and continues its search, showing you each special number it finds. It will stop after it has shown you a hundred of these special numbers.

\textbf{\\Reconstructed Program for non-obfuscated P4:}
\begin{verbatim}
#include <stdio.h>
#include <stdbool.h>

bool isPrime(int n) {
    if (n <= 1) return false;
    for (int i = 2; i * i <= n; i++) {
        if (n % i == 0) return false;
    }
    return true;
}

int main() {
    int count = 0; 
    int number = 2;
    while (count < 100) {
        if (isPrime(number)) {
            printf("%d ", number);
            count++;
        }
        number++;
    }
    printf("\n");

    return 0;
}
\end{verbatim}
\end{tcolorbox}

\begin{tcolorbox}[breakable, width=\textwidth+1cm]
\paragraph{P5:} This program computes the factorial of a number.\\
\begin{verbatim}
#include<stdio.h>  
int main()    
{    
 int i,fact=1,number;    
 printf("Enter a number: ");    
  scanf("%d",&number);    
    for(i=1;i<=number;i++){    
      fact=fact*i;    
  }    
  printf("Factorial of %d is: %d",number,fact);    
return 0;  
}
\end{verbatim}
\textbf{Explanation:\\}\\

Imagine a machine that helps you find the product of a series of counting numbers up to a given number. When you approach this machine, it asks you to tell it a specific number. Once you provide the number, the machine starts multiplying all the counting numbers from 1 up to the number you provided. 

For example, if you tell the machine the number 4, it will multiply 1 by 2, then multiply the result by 3, and then multiply that result by 4.

After finishing the calculations, the machine displays a message that tells you the result of multiplying all those counting numbers up to the number you provided.

\textbf{\\Reconstructed Program for non-obfuscated P5:}
\begin{verbatim}
#include <stdio.h>
unsigned long long factorial(int n) {
    if (n == 0 || n == 1) {
        return 1;
    } else {
        return n * factorial(n - 1);
    }
}

int main() {
    int number;
    printf("Enter a number: ");
    scanf("%d", &number);
    if (number < 0) {
        printf("Please enter a non-negative number.\n");
        return 1;
    }
    unsigned long long result = factorial(number);
    printf("Result", result);

    return 0;
}

\end{verbatim}
\end{tcolorbox}
\begin{tcolorbox}[breakable, width=\textwidth+1cm]
\paragraph{P6:} This program computes a desired number of digits of $e$ in any desired base.\\
\begin{verbatim}
#include <stdio.h>

#define MAX_DIGITS 1000

int result[MAX_DIGITS];
int temp[MAX_DIGITS];
int factorial[MAX_DIGITS];

void add(int *a, int *b) {
    for (int i = 0; i < MAX_DIGITS; i++) {
        a[i] += b[i];
        if (a[i] >= 10) {
            a[i+1] += a[i] / 10;
            a[i] %= 10;
        }
    }
}

void computeE(int digits) {
    result[0] = 1;
    factorial[0] = 1;

    for (int i = 1; i <= digits; i++) {
        for (int j = 0; j < MAX_DIGITS; j++) {
            temp[j] = factorial[j] = factorial[j] * i;
        }

        for (int j = 0; j < MAX_DIGITS; j++) {
            if (temp[j] >= 10) {
                temp[j+1] += temp[j] / 10;
                temp[j] %= 10;
            }
        }
        add(result, temp);
    }
}

void convertAndPrint(int base) {
    int output[MAX_DIGITS] = {0};
    int idx = 0;

    while (idx < MAX_DIGITS) {
        int remainder = 0;
        for (int i = MAX_DIGITS - 1; i >= 0; i--) {
            int temp = result[i] + remainder * 10;
            result[i] = temp / base;
            remainder = temp % base;
        }
        output[idx++] = remainder;
        int sum = 0;
        for (int i = 0; i < MAX_DIGITS; i++) {
            sum += result[i];
        }
        if (sum == 0) break;
    }

    for (int i = idx - 1; i >= 0; i--) {
        printf("%d ", output[i]);
    }
    printf("\n");
}

int main() {
    int desired_digits, base;
    printf("Enter the desired number of digits: ");
    scanf("%d", &desired_digits);
    printf("Enter the desired base (2 to 10): ");
    scanf("%d", &base);

    if (base < 2 || base > 10) {
        printf("Base out of range.\n");
        return 1;
    }

    computeE(desired_digits);
    convertAndPrint(base);
    return 0;
}
\end{verbatim}

\textbf{Explanation:\\}\\
Imagine a machine that can tell you about a special number, much like pi, which starts 2.71828... and goes on forever. The unique thing about this machine is that it can show this special number using different number systems, not just the one we usually use with numbers 0 to 9.

When you approach the machine, it asks you two things:
1. How many parts of this special number do you want to know? You tell it a number.
2. In which number system you'd like to see this special number? For this, you can choose any system between the one that uses just 0 and 1 and the usual one (that uses numbers 0 to 9). 

Once you provide this information, the machine works its magic. It calculates the value of this special number up to the number of parts you specified, in the number system you chose. Then, it displays that special value for you.

However, if you choose a number system outside the allowed range, the machine will tell you that the choice is not acceptable and stop.

\textbf{\\Reconstructed Program for non-obfuscated P6:}
\begin{verbatim}
 #include <stdio.h>
#include <stdlib.h>
#include <string.h>
#include <math.h>

double calculateE(int parts) {
    double e = 1.0;
    double factorial = 1.0;
    for (int i = 1; i < parts; i++) {
        factorial *= i;
        e += 1.0 / factorial;
    }
    return e;
}

void displayInBase(double number, int base, int parts) {
    printf("2."); // Display the integer part
    for (int i = 0; i < parts; i++) {
        number *= base;
        int digit = (int) number;
        printf("%d", digit);
        number -= digit;
    }
    printf("\n");
}

int main() {
    int parts, base;
    printf("How many parts of e do you want to know? ");
    scanf("%d", &parts);
    printf("In which base system (between 2 and 9 inclusive)? ");
    scanf("%d", &base);
    if (base < 2 || base > 9) {
        printf("The chosen number system is not acceptable.\n");
        return 1;
    }
    double e = calculateE(parts + 5);  
    displayInBase(e - 2.0, base, parts); 

    return 0;
}

\end{verbatim}
\end{tcolorbox}

\begin{tcolorbox}[breakable, width=\textwidth+1cm]
\paragraph{P7:} This program converts byte to binary.\\
\begin{verbatim}
#include <stdio.h>
void byteToBinary(unsigned char byte) {
    for (int i = 7; i >= 0; i--) {
        printf("%d", (byte >> i) & 1);
    }
    printf("\n");
}

int main() {
    unsigned char inputByte;
    printf("Enter a byte value (0-255): ");
    scanf("%hhu", &inputByte);

    printf("Binary representation: ");
    byteToBinary(inputByte);
    return 0;
}
\end{verbatim}

\textbf{Explanation:\\}\\
Imagine you have a line of 8 light bulbs, all turned off. These light bulbs can either be on (showing a light) or off (no light). Now, imagine you have a number between 0 and 255. Based on this number, some of these light bulbs will turn on, while others will remain off.
This program is like a guidebook. First, it asks you for that number. Once you give the number, the guidebook will show you a line indicating which light bulbs should be turned on and which should remain off, using a sequence of 1s (for the bulbs that are on) and 0s (for the bulbs that are off). By looking at this sequence, you will be able to recreate the exact arrangement of lit and unlit light bulbs for the number you provided.

\textbf{\\Reconstructed Program for non-obfuscated P7:}
\begin{verbatim}
#include <stdio.h>
int main() {
    int number;
    printf("Enter a number between 0 and 255: ");
    scanf("%d", &number);
    if (number < 0 || number > 255) {
        printf("Invalid number.\
        Please enter a number between 0 and 255.\n");
        return 1; 
    }
    printf("Light bulb sequence: ");
    for (int i = 7; i >= 0; i--) {
        if ((number & (1 << i)) != 0) {
            printf("1");
        } else {
            printf("0");
        }
    }
    printf("\n");
    return 0;
}

\end{verbatim}
\end{tcolorbox}

\begin{tcolorbox}[breakable, width=\textwidth+1cm]
\paragraph{P8:} This program adds two numbers.\\
\begin{verbatim}
#include <stdio.h>

int main() {
    int num1, num2, sum;

    printf("Enter the first number: ");
    scanf("%d", &num1);

    printf("Enter the second number: ");
    scanf("%d", &num2);

    sum = num1 + num2;

    printf("The sum of %d and %d is %d.\n", num1, num2, sum);

    return 0;
}

\end{verbatim}

\textbf{Explanation:\\}\\
Imagine you have a digital assistant with a screen. When you turn it on, it first asks you to type in a number. After you've provided the first number, it asks you for a second one. Once you've entered both numbers, the assistant does a quick mental calculation and shows you the result of adding these two numbers together. It displays a message like "The sum of the first number and the second number is the result." With this message, you immediately know how much the two numbers you provided add up to.

\textbf{\\Reconstructed Program for non-obfuscated P8:}
\begin{verbatim}
#include <stdio.h>

int main() {
    int firstNumber, secondNumber, sum;
    printf("Enter the first number: ");
    scanf("%d", &firstNumber);
    printf("Enter the second number: ");
    scanf("%d", &secondNumber);
    sum = firstNumber + secondNumber;
    printf("The sum of the first number (%d) and \
    the second number (%d)is %d.\n",\
    firstNumber, secondNumber, sum);

    return 0;
}

\end{verbatim}
\end{tcolorbox}

\begin{tcolorbox}[breakable, width=\textwidth+1cm]
\paragraph{P9:} This program calculates the integer part of square root of the input number.\\

\begin{verbatim}
#include <stdio.h>
#include <stdlib.h>
#include <math.h>
#include <string.h>

int main() {
    char number[100];  // buffer for input number
    printf("Enter a number with even number of digits: ");
    scanf("%s", number);

    int len = strlen(number);
    if (len % 2 != 0) {
        printf("The number must have an even number of digits.\n");
        return 1;
    }

    double value = atof(number);
    double sqrtValue = sqrt(value);

    printf("Square root: %d\n", (int)sqrtValue);
    return 0;
}

\end{verbatim}

\textbf{Explanation:\\}\\
Imagine a digital tool that helps you find the square root of numbers. When you start it, it asks you to type in a special kind of number: one that has an even count of digits. For example, if you have a number like "1234", it's okay because it has 4 digits. But a number like "123" won't work since it has 3 digits.
After you type in this number, the tool checks to make sure it has the right amount of digits. If the number you typed doesn't follow the rule, the tool will tell you that you've made a mistake and need to type in a number with an even count of digits.

However, if the number you typed is correct, the tool will quickly do some calculations and then show you the square root of that number. But, it only shows you the whole part of the square root, without any decimals.

\textbf{\\Reconstructed Program for non-obfuscated P9:}
\begin{verbatim}
#include <stdio.h>
#include <math.h>
#include <string.h>

int main() {
    char numberString[100];  
    int isValid = 1;
    printf("Enter a number with an even count of digits: ");
    scanf("%s", numberString); 
    int length = strlen(numberString);
    if (length % 2 != 0) {
        printf("The number you entered has an odd count of digits.\
        Please enter a number with an even count of digits.\n");
        isValid = 0;
    }

    if (isValid) {
        double num = atof(numberString); 
        int squareRootWholePart = (int)sqrt(num);
        printf("The whole part of the square root\
        of %s is %d.\n", numberString, squareRootWholePart);
    }

    return 0;
}

\end{verbatim}
\end{tcolorbox}

\begin{tcolorbox}[breakable, width=\textwidth+1cm]
\paragraph{P10:} A program that plays a two-player tic-tac-toe.\\
\begin{verbatim}
#include <stdio.h>
#define SIZE 3

// Function prototypes
void displayBoard(char board[SIZE][SIZE]);
int checkWinner(char board[SIZE][SIZE]);
void getPlayerMove(char board[SIZE][SIZE], char player);

int main() {
    char board[SIZE][SIZE] = {{' ', ' ', ' '},\
    {' ', ' ', ' '}, {' ', ' ', ' '}};
    int moves = 0;
    char currentPlayer = 'X';

    displayBoard(board);
    
    while (moves < SIZE * SIZE) {
        getPlayerMove(board, currentPlayer);

        displayBoard(board);

        if (checkWinner(board)) {
            printf("Player %c wins!\n", currentPlayer);
            return 0;
        }

        currentPlayer = (currentPlayer == 'X') ? 'O' : 'X';
        moves++;
    }

    printf("It's a draw!\n");
    return 0;
}

void displayBoard(char board[SIZE][SIZE]) {
    for (int i = 0; i < SIZE; i++) {
        for (int j = 0; j < SIZE; j++) {
            printf("%c", board[i][j]);
            if (j < SIZE - 1) printf(" | ");
        }
        printf("\n");
        if (i < SIZE - 1) printf("---------\n");
    }
}

void getPlayerMove(char board[SIZE][SIZE], char player) {
    int x, y;
    do {
        printf("Player %c, enter your move \
        (row [0-2] and column [0-2]): ", player);
        scanf("%d%d", &x, &y);
    } while (x < 0 || x >= SIZE || \
    y < 0 || y >= SIZE || board[x][y] != ' ');

    board[x][y] = player;
}

int checkWinner(char board[SIZE][SIZE]) {
    // Check rows, columns and diagonals
    for (int i = 0; i < SIZE; i++) {
        if (board[i][0] != ' ' && board[i][0] == board[i][1] \
        && board[i][1] == board[i][2]) return 1;
        if (board[0][i] != ' ' && board[0][i] == board[1][i] \
        && board[1][i] == board[2][i]) return 1;
    }
    if (board[0][0] != ' ' && board[0][0] == board[1][1] \
    && board[1][1] == board[2][2]) return 1;
    if (board[0][2] != ' ' && board[0][2] == board[1][1] \
    && board[1][1] == board[2][0]) return 1;

    return 0;
}

\end{verbatim}

\textbf{Explanation:\\}\\
Imagine a classic game where two players take turns marking spaces on a 3x3 grid. One player uses an 'X' and the other uses an 'O'. The goal of the game is to mark three spaces in a row, either horizontally, vertically, or diagonally before the opponent does.

When the game starts, the grid is empty. The game tool displays the grid to show both players the current state of the game. Each space in the grid can be identified using two numbers: the row number and the column number, both ranging from 0 to 2.

One player goes first, marking a space on the grid with their symbol. The game tool then updates the grid to show this mark. After each turn, the tool checks if there are any three matching symbols in a row. If there is a winning line, it announces which player has won.

If all the spaces on the grid are marked, and no player has three of their symbols in a row, then the game is declared a draw.

The game continues by allowing players to take turns marking spaces until someone wins or all spaces are marked, resulting in a draw.\\
\textbf{\\Reconstructed Program for non-obfuscated P10:}
\begin{verbatim}
#include <stdio.h>
#include <stdbool.h>

char board[3][3] = {{' ', ' ', ' '},
                    {' ', ' ', ' '},
                    {' ', ' ', ' '}};

bool checkWin(char player) {
    for (int i = 0; i < 3; i++) {
        if (board[i][0] == player && board[i][1] == player &&\
        board[i][2] == player) return true;
        if (board[0][i] == player && board[1][i] == player &&\
        board[2][i] == player) return true;
    }
    if (board[0][0] == player && board[1][1] == player &&\
    board[2][2] == player) return true;
    if (board[0][2] == player && board[1][1] == player &&\
    board[2][0] == player) return true;

    return false;
}

bool checkDraw() {
    for (int i = 0; i < 3; i++) {
        for (int j = 0; j < 3; j++) {
            if (board[i][j] == ' ') return false;
        }
    }
    return true;
}

void displayBoard() {
    for (int i = 0; i < 3; i++) {
        for (int j = 0; j < 3; j++) {
            printf("%c", board[i][j]);
            if (j < 2) printf(" | ");
        }
        printf("\n");
        if (i < 2) printf("---------\n");
    }
}

int main() {
    char currentPlayer = 'X';
    int row, col;

    while (true) {
        displayBoard();

        printf("Player %c, enter row (0-2)\
        and column (0-2): ", currentPlayer);
        scanf("%d %d", &row, &col);

        if (row >= 0 && row < 3 && col >= 0 &&\
        col < 3 && board[row][col] == ' ') {
            board[row][col] = currentPlayer;

            if (checkWin(currentPlayer)) {
                displayBoard();
                printf("Player %c wins!\n", currentPlayer);
                break;
            }

            if (checkDraw()) {
                displayBoard();
                printf("It's a draw!\n");
                break;
            }

            currentPlayer = (currentPlayer == 'X') ? 'O' : 'X';
        } else {
            printf("Invalid move, try again.\n");
        }
    }

    return 0;
}

\end{verbatim}

\end{tcolorbox}

\section{Useless code snippets added for Obfuscation}
\subsection{Synthesized Set}
\subsubsection{Level 1}
\begin{tcolorbox}[breakable, width=\textwidth+1cm]
\begin{verbatim}
if (u.canHarvest()) then {
    for (unit u){
        if (u.isBuilder()) then{	
    }
    else{
        u.harvest(50);
            }
        }
}
for (Unit u){
    if (u.is_Type(Worker)) then{
        u.train(Heavy, Enemydir,10);
        if (u.canAttack()) then {
            u.train(Ranged,Up,16)
        }
    }
}

\end{verbatim}
\end{tcolorbox}
\subsubsection{Level 2}
\begin{tcolorbox}[breakable, width=\textwidth+1cm]

\begin{verbatim}
if (u.canHarvest()) then {
    for (unit u){
        u.train(Heavy,Left, 9)
        if (u.isBuilder()) then{
        
        }
        else{
            u.harvest(50);
        }
    }
}
for (Unit u){
    if (u.is_Type(Worker)) then{
        u.train(Heavy, Enemydir,10);
        if (u.canAttack()) then {
            u.train(Ranged,Up,16)
            if (u.canHarvest()) then {
                u.train(Worker,Right, 9)
                if (u.isBuilder()) then{
                    if (u.is_Type(Barracks)) then{
                        u.harvest(10);
                        u.train(Workers, Right, 2);
                    }
                    else{
                        u.train(Light,Down,5)
                        }
                }
    
            }
        }
    }
    else{
        u.harvest(1);
    }
    for(Unit u){
        u.idle()
    }
}
for (Unit u){
    if(u.hasLessNumberOfUnits(Heavy, 2)){
        if (u.isBuilder()) then{
            u.train(Light, Farthest, 1);
        }
        else{
            u.harvest(3);
        } 

    }
    if (hasNumberOfUnits(Base, 3)){
        if (u.is_Type(Barracks)) then{
            u.build(Base,EnemyDir,1);
            u.moveToUnit(Ally, MostHealthy);
        }
        else{
            u.train(Heavy, Random,10);
        } 
    }
}
\end{verbatim}
\end{tcolorbox}

\subsection{Human-crafted Set}
\subsubsection{Level 1}
\begin{tcolorbox}[breakable, width=\textwidth+1cm]
\begin{verbatim}
if (u.getType()==barracksType && u.getType().canHarvest &&\
p.getResources() <= heavyType.cost){
    train(u, heavyType);
    buildIfNotAlreadyBuilding(u,baseType,u.getX(),u.getY(),\
    reservedPositions,p,pgs);
    }
int nlight = 0;
for (Unit u4 : pgs.getUnits()) {
    if (u4.getType() == lightType && u4.getPlayer() != p.getID()) {
        nlight++;
    }
}
if (nlight < 2 && u.getType().canMove) {  
    if (u.getType().canHarvest){
        train(u, lightType);
    }
}
Unit closestBase = null;
Unit closestResource = null;
int closestDistance = 0;
for (Unit u3 : pgs.getUnits()) {
    if (u3.getType().isResource) {
        int d = Math.abs(u3.getX() - u.getX()) +\
        Math.abs(u3.getY() - u.getY());
        if (closestResource == null || d < closestDistance) {
            closestResource = u3;
            closestDistance = d;
        }
    }
}
closestDistance = 0;
for (Unit u4 : pgs.getUnits()) {
    if (u4.getType().isStockpile && u4.getPlayer()==p.getID()) {
        int d = Math.abs(u4.getX() - u.getX()) +\
        Math.abs(u4.getY() - u.getY());
        if (closestBase == null || d < closestDistance) {
            closestBase = u4;
            closestDistance = d;
        }
    }
}
if (closestResource != null && closestBase != null &&\
u.getType().canHarvest) {
    harvest(u, closestResource, closestBase);
}
if (u.getType().canMove == true &&\
p.getResources() >= workerType.cost) {
    buildIfNotAlreadyBuilding(u,baseType,u.getX(),u.getY(),\
    reservedPositions,p,pgs);
    train(u, heavyType);
}
\end{verbatim}
\end{tcolorbox}

\subsubsection{Level 2}
\begin{tcolorbox}[breakable, width=\textwidth+1cm]
\begin{verbatim}
Unit TargetEnemy = null;
int Distance = 0;
int Mybase = 0;
int Mybarrack = 0;
for (Unit u6 : pgs.getUnits()) {
    if (u6.getPlayer() >= 0 && u6.getPlayer() != p.getID()) {
        int d = Math.abs(u6.getX() - u.getX()) +\
        Math.abs(u6.getY() - u.getY());
        if (TargetEnemy == null || d > Distance) {
            TargetEnemy = u6;
            Distance = d;
        }
    }
    else if(u6.getPlayer()==p.getID() &&\
    u6.getType() == baseType){
        Mybase = Math.abs(u6.getX() - u.getX()) +\
        Math.abs(u6.getY() - u.getY());
    }
    else if(u6.getPlayer()==p.getID() &&\
    u6.getType() == barracksType){
        Mybarrack = Math.abs(u6.getX() - u.getX()) +\
        Math.abs(u6.getY() - u.getY());
    }
}
if (u.getType() == workerType && TargetEnemy!=null &&\
(Distance < pgs.getHeight()/2 || Mybase < pgs.getHeight()/2)) {
    attack(u, TargetEnemy);
}
else if(u.getType() == workerType &&\
!(Distance < pgs.getHeight()/2 || Mybase < pgs.getHeight()/2)){
    buildIfNotAlreadyBuilding(u, barracksType, u.getX(),u.getY(),\
    reservedPositions,p,pgs);
}
else if(u.getType() == workerType &&\
!(Distance < pgs.getHeight()/2 || Mybase < pgs.getHeight()/2)){
    buildIfNotAlreadyBuilding(u, baseType, u.getX(),u.getY(),\
    reservedPositions,p,pgs);
}
for (Unit u5 : pgs.getUnits()) {
    if (u5.getType() == baseType || u5.getType() == barracksType ){
        if (u5.getPlayer() == p.getID()){
            if (TargetEnemy!=null && u.getType() == workerType &&(\
            Distance>pgs.getHeight()/2 || Mybase>pgs.getHeight()/2)) 
            {
                attack(u5, TargetEnemy);
                if (u5.getType()==workerType){
                    train(u, lightType);
                }
            }
        }
    }
}
if (u.getType()==barracksType && u.getType().canHarvest &&\
p.getResources() <= heavyType.cost){
    train(u, heavyType);        
    buildIfNotAlreadyBuilding(u,baseType,u.getX(),u.getY(),\
    reservedPositions,p,pgs);
}
int nlight = 0;
for (Unit u4 : pgs.getUnits()) {
    if (u4.getType() == lightType && u4.getPlayer() != p.getID()) {
        nlight++;
    }
}
if (nlight < 2 && u.getType().canMove) {  
    if (u.getType().canHarvest){
        train(u, lightType);
    }
}
Unit closestBase = null;
Unit closestResource = null;
int closestDistance = 0;
for (Unit u3 : pgs.getUnits()) {
    if (u3.getType().isResource) {
        int d = Math.abs(u3.getX() - u.getX()) +\
        Math.abs(u3.getY() - u.getY());
        if (closestResource == null || d < closestDistance) {
            closestResource = u3;
            closestDistance = d;
        }
    }
}
closestDistance = 0;
for (Unit u4 : pgs.getUnits()) {
    if (u4.getType().isStockpile && u4.getPlayer()==p.getID()) {
        int d = Math.abs(u4.getX() - u.getX()) +\
        Math.abs(u4.getY() - u.getY());
        if (closestBase == null || d < closestDistance) {
            closestBase = u4;
            closestDistance = d;
        }
    }
}
if (closestResource != null && closestBase != null &&\
u.getType().canHarvest) {
    harvest(u, closestResource, closestBase);
}
if (u.getType().canMove == true &&\
p.getResources() >= workerType.cost) {
    buildIfNotAlreadyBuilding(u,baseType,u.getX(),u.getY(),\
    reservedPositions,p,pgs);
    train(u, heavyType);
}
\end{verbatim}
\end{tcolorbox}

\subsubsection{Discussion}
The main reason we need more lines to break down programs in the human-crafted set compared to the synthesized set, is due to the language they are written in. The human-crafted set is written in Java, a language that the LLM knows very well. This makes it difficult to confuse or mislead the LLM with Java. We observed that GPT-4 is particularly good at reconstructing Java code from explanations.The process of explaining is also more straightforward with Java, as GPT-4 finds it easier to understand and translate Java into explanations compared to a newly defined language. Consequently, we required a greater number of lines to sufficiently obfuscate the programs.

\section{Set of MicroRTS Programs}
\subsection{Synthesized Set}

\begin{tcolorbox}[breakable, width=\textwidth+1cm]
\textbf{P1:}
\begin{verbatim}
    
for(Unit u){
    for(Unit u){
        u.harvest(2)
        u.idle()
    }
    u.train(Light,Up,25)
    u.train(Worker,Right,4)
    u.attack(Closest)
    for(Unit u){
        u.train(Heavy,Up,3)
    }
    for(Unit u){
        u.build(Barracks,Up,1)
    }
}

\end{verbatim}
\textbf{Explanation:\\}

\noindent
To get started, the very first action this strategy takes with topmost priority involves sending each of our units to gather resources. Specifically, every individual unit is assigned to harvest a maximum of two resources. After that, these units are instructed to remain idle. The idle instruction is crucial because it ensures that while these units are not engaged in harvesting, they still remain vigilant, ready to attack any opponent unit that comes within their attack range.

After the initial phase of resource harvesting, the next priority of this strategy is to train new units. Initially, we aim to train up to 25 Light units in the upward direction from their respective training structures. Following this, we have a target to train up to 4 Worker units who would be placed on the right side of their training structure.

The next action, which comes after the training of Light and Worker units, is attacking the opponent units. Here, the strategy is to attack the closest opponent unit. Hence, if there are several opponent units within the attack range, the closest one will be targeted first.

Subsequently, we have two priorities that occur concurrently. On one hand, we aim to train a limited number of Heavy units (up to 3 in number). On the other hand, we look to construct Barracks with a maximum limit of 1 Barracks. The placement for both these actions is towards the upward direction from the current position.

In summary, the overall goal of this strategy is to maintain an equilibrium between resource harvesting, unit training, and combating the enemy. By prioritizing resource gathering initially, we ensure that there are enough resources available for training new units and building structures. Following that, we put a focus on training both combatant (Light and Heavy) and resource gathering (Worker) units, thereby maintaining the balance in our forces. Finally, the strategy underscores the importance of continuous attacks on the closest enemy while simultaneously enhancing our forces through the construction of Barracks and training of Heavy units.\\\\

\textbf{Best Reconstructed Program for P1:}
\begin{verbatim}
for(Unit u) {
    u.harvest(10);
    u.idle();
}
for(Unit u) {
    u.train(Light, Up, 25);
}
for(Unit u) {
    u.train(Worker, Right, 4);
    u.attack(Closest);
}
for(Unit u) {
    u.train(Heavy, Up, 3);
}
for(Unit u) {
    u.build(Barracks, Up, 1);
}
\end{verbatim}
\end{tcolorbox}

\begin{tcolorbox}[breakable, width=\textwidth+1cm]
\textbf{P2:}
\begin{verbatim}
    
for(Unit u){
		u.build(Barracks,EnemyDir,7)
		u.harvest(9)
		for(Unit u){
			u.train(Heavy,Left,20)
		}
		u.train(Worker,Right,3)
		u.attack(Weakest)
		for(Unit u){
			u.idle()
		}
		u.moveToUnit(Ally,LessHealthy)
		for(Unit u){
			u.train(Light,Left,2)
		}
}

\end{verbatim}
\textbf{Explanation:\\}

\noindent
The provided strategy for playing MicroRTS takes a structured approach, prioritizing certain actions over others in an attempt to gain an advantage. It is important to understand the priority of actions to follow and execute this strategy properly.
The initial and highest priority of this strategy is to establish a strong defense by building Barracks towards the enemy's direction. The aim here is to construct 7 Barracks, which would serve as the primary base for training combat units. At the same time, the strategy also assigns priority to resource collection, instructing workers to harvest resources, targeting 9 resource units.

The next action with high priority is to train Heavy units from the units within the Barracks. These units are positioned on the left of the Barracks. The training aims to produce up to 20 Heavy units which would serve as the main force in our offensive and defensive operations.

Subsequently, the strategy focuses on replenishing and expanding the workforce. The Barracks are instructed to train more Workers on the right side, targeting a total of 3 additional Workers. These additional workers would serve to increase the efficiency of resource collection and building processes.

With the fortifications and army in place, the strategy now shifts towards offense. The units are directed to attack the weakest of the opponent's units. This action is aimed at diminishing the opponent's strength and defenses by targeting their vulnerable units first.

After the attack command, the next action with priority is to direct units to remain idle. This action ensures that the units conserve their energy and are prepared for any sudden attacks from the opponent.

The strategy then focuses on ensuring the survival of ally units. Units are instructed to move towards the least healthy ally unit. This action might help in providing reinforcement or drawing away enemy units from the weaker allies.

The final action in this strategy is to train Light units. Light units are quicker to train and would serve as a rapid response force. The strategy aims to train 2 Light units on the left side of the Barracks.

In summary, the overall goal of this strategy is to build a strong defense by quickly constructing Barracks and training Heavy units, maintain a steady flow of resources, selectively target vulnerable enemy units while ensuring the survival of ally units, and lastly, training a rapid response force of Light units.\\\\

\textbf{Best Reconstructed Program for P2:}
\begin{verbatim}
for(Unit u){
    for(Unit u){
        u.build(Barracks, EnemyDir, 7)
        u.harvest(9)
        u.train(Heavy, Left, 20)
    }
    for(Unit u){
        u.train(Worker, Right, 3)
        u.attack(Weakest)
        u.idle()
        u.moveToUnit(Ally, LessHealthy)
        u.train(Light, Left, 2)
    }
}
\end{verbatim}
\end{tcolorbox}

\begin{tcolorbox}[breakable, width=\textwidth+1cm]
\textbf{P3:}
\begin{verbatim}
for(Unit u){
    for(Unit u){
        u.idle()
    }
    u.moveToUnit(Enemy,Closest)
    u.train(Worker,EnemyDir,2)
    for(Unit u){
        u.train(Heavy,EnemyDir,100)
        u.train(Light,Left,8)
        u.attack(Weakest)
        if(u.OpponentHasUnitInPlayerRange()) then {
            u.moveToUnit(Enemy,LessHealthy)
        }
        for(Unit u){
            u.build(Barracks,EnemyDir,20)
            u.harvest(6)
            u.attack(LessHealthy)
            u.moveToUnit(Ally,Weakest)
        }
    }
}

\end{verbatim}
\textbf{Explanation:\\}

\noindent
The strategic gameplay outlined in program P prioritizes actions to maximize efficiency and tactical advantage in the MicroRTS game. The program begins with the top priority action, instructing every unit in the player's control to stay idle. This implies that unless a unit is immediately threatened, it should initially hold its position and avoid proactive movement or combat. 
After this first instruction, the player's units are instructed to relocate towards the nearest enemy unit, in order to prepare for an impending confrontation. This directive stands as the second priority in the strategy.

The third priority is to utilize the player's bases to train two worker units in the direction of the enemy. This is intended to increase the player's resource collection capability while also adding a mild defensive presence against incoming enemy units.

Next, for every unit, it prioritizes training a large army of heavy combat units directed towards the enemy. If this is not feasible (e.g., the unit in question is not a barracks), it then looks to train light combat units to the left. After training these units, the next action prioritized is for them to target and attack the weakest enemy units.

Within this framework, if the player's unit detects an enemy unit within its attack range, the unit will change its course and move towards the less healthy enemy unit. This dynamic movement decision is prioritized after the initial set of actions.

Further, for every unit, there is a subsequent set of priority actions. Here, the units first build up to 20 barracks towards the enemy's direction, if possible. Then, it prioritizes the harvest of resources by sending out six worker units. After resource collection, units then attack the less healthy enemy unit and finally, if no other actions are feasible, the units move towards the weakest ally unit for support or defense.

The overarching goal of this strategy is to bolster the player's resources and army strength before engaging in direct confrontation with the enemy. By keeping units idle initially, then focusing on resource collection and building a large army, the strategy aims to avoid premature engagement and instead build up a strong base and military power. Lastly, with a substantial force, the strategy then aims to engage the enemy by targeting their weakest units first, utilizing the player's strength in numbers and resource advantage.\\\\

\textbf{Best Reconstructed Program for P3:}
\begin{verbatim}
for(Unit u) {
    u.idle();
    u.moveToUnit(Enemy, Closest);
    u.train(Worker, EnemyDir, 2);
    u.train(Heavy, EnemyDir, 50);
    u.train(Light, Left, 50);
    u.attack(Weakest);
    if(u.opponentHasUnitInPlayerRange()) {
        u.moveToUnit(Enemy, LessHealthy);
    }
    for(Unit u) {
        u.build(Barracks, EnemyDir, 20);
        u.harvest(6);
        u.attack(LessHealthy);
        if(u.canAttack()) {
            u.moveToUnit(Ally, Weakest);
        }
    }
}
\end{verbatim}
\end{tcolorbox}

\begin{tcolorbox}[breakable, width=\textwidth+1cm]
\textbf{P4:}
\begin{verbatim}
for(Unit u){
    if(u.HasUnitInOpponentRange()) then {
        for(Unit u){
            u.moveToUnit(Enemy,Strongest)
        }
    }
    u.harvest(2)
    for(Unit u){
        for(Unit u){
            u.train(Light,Up,1)
        }
        u.train(Ranged,Left,50)
        u.idle()
    }
    u.build(Barracks,Down,20)
    u.train(Worker,EnemyDir,20)
    u.attack(Closest)
}
\end{verbatim}
\textbf{Explanation:\\}

\noindent
The strategy here revolves around specific actions prioritized to execute in order, facilitating effective gameplay in MicroRTS. 
To start, the utmost priority is given to the condition where our units are within an opponent's attack range. In this scenario, all units are commanded to move towards the enemy's strongest unit. This move helps to concentrate the attack on the enemy's most powerful unit, potentially destabilizing their strategies.

Subsequently, the strategy allows a couple of units to harvest resources. This step ensures we have a continuous supply of resources necessary for building structures and training units.

Next in the line of action priority, the strategy instructs all units to train a single Light unit upwards. This step is followed by training 50 Ranged units towards the left and then keeping them idle until an opponent comes within attack range. The idea here is to amass a significant number of Ranged units to bolster defenses, while the Light unit acts as a quick response unit to potential threats.

In the following step, the strategy commands to build up to 20 Barracks downwards from our current position. Barracks serve as the crucial point for training combat units. Hence, building Barracks broadens our ability to train more combat units, therefore strengthening our defenses and attack capabilities.

Subsequent to building Barracks, the strategy focuses on strengthening our workforce by training up to 20 Worker units towards the enemy's direction. The workers play a pivotal role in collecting resources and building structures, thus maintaining a strong workforce is vital to support our ongoing efforts in the game.

Finally, with the least priority, all units are directed to attack the closest opponent. This action ensures we maintain some pressure on the opponent even as we build up our forces and defenses.

The overall goal of this strategy is to establish strong defensive structures, maintain a robust workforce for resource collection and structure building, while continually training combat units for offensive and defensive purposes. All this is done while keeping an eye on the opponent's actions, enabling us to react swiftly to any potential threats.\\\\

\textbf{Best Reconstructed Program for P4:}
\begin{verbatim}
for(Unit u) {
    if(u.hasUnitInOpponentRange()) {
        u.moveToUnit(Enemy, Strongest)
    }

    for(Unit u) {
        u.harvest(2)
    }
    
    for(Unit u) {
        u.train(Light, Up, 1)
    }

    for(Unit u) {
        u.train(Ranged, Left, 50)
        u.idle()
    }

    for(Unit u) {
        u.build(Barracks, Down, 20)
    }

    for(Unit u) {
        u.train(Worker, EnemyDir, 20)
    }

    for(Unit u) {
        u.attack(Closest)
    }
}
\end{verbatim}
\end{tcolorbox}

\begin{tcolorbox}[breakable, width=\textwidth+1cm]
\textbf{P5:}
\begin{verbatim}
for(Unit u){
    for(Unit u){
        u.train(Worker,EnemyDir,8)
        u.idle()
        for(Unit u){
            u.train(Light,Left,20)
        }
    }
    for(Unit u){
        u.harvest(2)
    }
    u.build(Barracks,Down,6)
    u.attack(Closest)
    u.train(Worker,Down,20)
}

\end{verbatim}
\textbf{Explanation:\\}

\noindent
The strategy provided is meant to be aggressive and prioritizes building a robust workforce first to support rapid unit production and resource gathering.
The first and most crucial aspect of the strategy is to consistently and actively train workers and assign them to idle tasks. The idle state commands the workers to stay put and only attack if an enemy unit comes within their attack range. The strategy primarily focuses on training up to eight workers in the direction of the enemy. This action has the highest priority in our gameplay.

Simultaneously, there is also an emphasis on training up to 20 Light combat units positioned towards the left of the base or barracks. This task is of secondary importance in this strategy. Light units are typically fast and can attack well but may not be as durable. Hence, they can provide a good offensive line against the enemy.

Next, the strategy focuses on gathering resources by having units harvest them. Up to two units are chosen for this task at any given time. Although this action isn't the highest priority, it is crucial to ensure a steady inflow of resources for training more units and building more structures.

After the workforce and resources have been established, the next priority is to build up to six Barracks downwards from the base. Barracks are essential as they are the structures that train combat units, which are vital for both defense and offense in the game.

At the same time, the strategy dictates that our units should engage in battle with the closest enemy unit. This approach can ensure that any nearby threats are dealt with quickly.

The final action of this strategy is to train an additional set of up to 20 workers in the downward direction. While this action has the lowest priority, it ensures that we have enough workers in the late game to maintain production and resource gathering capabilities.

In summary, the overall goal of this strategy is to establish a robust workforce, ensuring a steady flow of resources, and maintaining a steady line of defense and offense against the enemy. The focus is on aggressive expansion and keeping pressure on the enemy while maintaining a robust economy.\\\\

\textbf{Best Reconstructed Program for P5:}
\begin{verbatim}

for(Unit u) {
    u.train(Worker, EnemyDir, 8)
    u.train(Light, Left, 20)
    u.harvest(2)
}

for(Unit u) {
    u.build(Barracks, Down, 6)
    u.attack(Closest)
}

for(Unit u) {
    u.train(Worker, Down, 20)
}
\end{verbatim}
\end{tcolorbox}

\begin{tcolorbox}[breakable, width=\textwidth+1cm]
\textbf{P6:}
\begin{verbatim}
for(Unit u){
    for(Unit u){
        u.idle()
        u.harvest(2)
        for(Unit u){
            u.train(Light,Up,25)
        }
        for(Unit u){
            u.train(Worker,Up,2)
        }
        u.build(Barracks,Up,7)
    }
    u.attack(Closest)
    u.moveToUnit(Enemy,MostHealthy)
    u.train(Worker,EnemyDir,9)
}

\end{verbatim}
\textbf{Explanation:\\}

\noindent
In the context of the MicroRTS game, the strategy implied by this program prioritizes several actions in a specific order to achieve victory over the opponent.
At the very outset, the highest priority is given to maintaining a defensive stance and gathering resources. So, the first thing all units, including Workers, Combat units (Light, Heavy, Ranged) will do is adopt a passive stance and wait for an enemy to approach them. In tandem, they will also focus on collecting resources, but with a limit of collecting from at most 2 resource units.

Once this defensive and resource gathering operation is in place, the game strategy shifts to training new units, with a higher emphasis on Light combat units. Specifically, the next priority is to train as many as 25 Light units that will serve as the primary fighting force of the player's army. This action is followed closely by training of Worker units, with a cap of 2 new Workers to join the existing workforce for more efficient resource collection.

The next action in priority is to bolster the production capabilities by constructing Barracks, aiming to have up to 7 of them. This will allow faster training of new combat units, ensuring a steady stream of reinforcements.

After these initial preparations are complete, the strategy moves to more aggressive actions. Each unit in the player's force will focus on attacking the closest enemy unit. This proactive action is designed to minimize potential damage by eliminating nearby threats.

Following the attack, the next priority action is to maneuver our units strategically towards the healthiest enemy unit. This action can help in weakening the enemy's strong units and balancing the battlefield in favor of our army.

Lastly, and with the lowest priority in this strategy, the aim is to train additional Worker units, up to 9, near the enemy's direction. These Workers can help in speedy resource collection near enemy territory while also potentially serving as distractions for enemy attacks.

The overall goal of this strategy is to create a strong defense by prioritizing idle and resource collection actions, followed by expanding the army through training and construction of barracks. Once a strong defense and robust army are in place, the strategy switches to aggressive tactics by prioritizing attack and movement towards the enemy's healthiest units. The endgame is to have a strong and well-equipped army that can tackle any challenge, supported by an efficient workforce collecting resources close to the enemy base.\\\\

\textbf{Best Reconstructed Program for P6:}
\begin{verbatim}
for(Unit u) {
        u.idle()
        u.harvest(2)
    }

    for(Unit u) {
        u.train(Light, EnemyDir, 25)
        u.train(Worker, EnemyDir, 2)
    }

    for(Unit u) {
        u.build(Barracks, EnemyDir, 7)
    }

    for(Unit u) {
        u.attack(Closest)
    }

    for(Unit u) {
        u.moveToUnit(Enemy, MostHealthy)
    }

    for(Unit u) {
        u.train(Worker, EnemyDir, 9)
    }
\end{verbatim}
\end{tcolorbox}

\begin{tcolorbox}[breakable, width=\textwidth+1cm]
\textbf{P7:}
\begin{verbatim}
for(Unit u){
    u.attack(Closest)
    u.train(Worker,Up,2)
    for(Unit u){
        u.idle()
        u.harvest(2)
    }
    for(Unit u){
        u.build(Barracks,Right,15)
        u.moveToUnit(Enemy,MostHealthy)
        u.train(Light,EnemyDir,8)
        u.moveToUnit(Enemy,Closest)
        for(Unit u){
            u.train(Worker,EnemyDir,15)
        }
    }
}
\end{verbatim}
\textbf{Explanation:\\}

\noindent
The strategy is organized in a hierarchical way, according to the priority of actions.
The highest priority in this strategy is given to launching an attack on the nearest opponent unit. This action is initiated as soon as the game starts, suggesting an aggressive opening.

The second priority is to train two Worker units in the upward direction. These Workers will play a crucial role in resource management and building of structures.

The third priority is to command all the units to remain idle until an opponent unit comes within their attack range. These idle units, thus, are primarily defensive units.

In addition to this, two Workers are sent to harvest resources. These are likely the same Workers that were trained in the upward direction earlier. Harvesting resources is crucial in maintaining an economy and supporting the build and training of units.

At the next level of priority, a significant part of the strategy is unfolded. Each unit is given a sequence of tasks - building Barracks in the right direction, with a cap of 15, moving towards the healthiest enemy unit, training 8 Light units in the direction of the enemy, and moving again towards the closest enemy unit.

The construction of Barracks ensures a regular supply of combat units. The movement of units towards the healthiest and then the closest enemy units suggests a strategy that focuses on taking down the strongest opponent units first.

The training of Light units specifically points towards a preference for quicker, lighter forces over slow, heavy ones. Light units are often used for quick attacks and to keep the enemy's attention away from the main actions.

The lowest priority action is training 15 additional Workers in the direction of the enemy. This implies that there will be a late-game focus on building and resource collection, possibly in preparation for a final massive assault or to provide an economic edge over the opponent.

The overall goal of this strategy appears to be creating a powerful initial assault with rapid unit training and attack on the enemy, while simultaneously building up a steady resource income for future maneuvers. The later stages of the game will focus on strengthening the economy and mass-producing units for a final assault or a dominant economic victory.\\\\

\textbf{Best Reconstructed Program for P7:}
\begin{verbatim}
for (Unit u) {
    u.attack(Closest);
    u.train(Worker, Up, 2);
    u.idle();
    u.harvest(2);
    u.build(Barracks, Right, 15);
    u.moveToUnit(Enemy, MostHealthy);
    for (Unit u) {
        u.train(Light, EnemyDir, 8);
        u.moveToUnit(Enemy, Closest);
        u.train(Worker, EnemyDir, 15);
    }
}
\end{verbatim}
\end{tcolorbox}

\begin{tcolorbox}[breakable, width=\textwidth+1cm]
\textbf{P8:}
\begin{verbatim}
for(Unit u){
		u.train(Ranged,Left,100)
		if(u.canHarvest()) then {
			u.attack(Strongest)
		}
		for(Unit u){
			u.train(Worker,Right,7)
		}
		u.train(Worker,Up,15)
		for(Unit u){
			u.build(Barracks,EnemyDir,5)
		}
		for(Unit u){
			u.idle()
			u.harvest(4)
		}
		u.attack(Closest)
		for(Unit u){
			u.train(Heavy,EnemyDir,3)
		}
		u.moveToUnit(Ally,Strongest)
}

\end{verbatim}
\textbf{Explanation:\\}

\noindent
The strategy is organized in a hierarchical way, according to the priority of actions.
The highest priority in this strategy is given to launching an attack on the nearest opponent unit. This action is initiated as soon as the game starts, suggesting an aggressive opening.

The second priority is to train two Worker units in the upward direction. These Workers will play a crucial role in resource management and building of structures.

The third priority is to command all the units to remain idle until an opponent unit comes within their attack range. These idle units, thus, are primarily defensive units.

In addition to this, two Workers are sent to harvest resources. These are likely the same Workers that were trained in the upward direction earlier. Harvesting resources is crucial in maintaining an economy and supporting the build and training of units.

At the next level of priority, a significant part of the strategy is unfolded. Each unit is given a sequence of tasks - building Barracks in the right direction, with a cap of 15, moving towards the healthiest enemy unit, training 8 Light units in the direction of the enemy, and moving again towards the closest enemy unit.

The construction of Barracks ensures a regular supply of combat units. The movement of units towards the healthiest and then the closest enemy units suggests a strategy that focuses on taking down the strongest opponent units first.

The training of Light units specifically points towards a preference for quicker, lighter forces over slow, heavy ones. Light units are often used for quick attacks and to keep the enemy's attention away from the main actions.

The lowest priority action is training 15 additional Workers in the direction of the enemy. This implies that there will be a late-game focus on building and resource collection, possibly in preparation for a final massive assault or to provide an economic edge over the opponent.

The overall goal of this strategy appears to be creating a powerful initial assault with rapid unit training and attack on the enemy, while simultaneously building up a steady resource income for future maneuvers. The later stages of the game will focus on strengthening the economy and mass-producing units for a final assault or a dominant economic victory.\\\\

\textbf{Best Reconstructed Program for P8:}
\begin{verbatim}
for(Unit u) {
    u.train(Ranged, Left, 100)
    if(u.canHarvest()) then { 
            u.attack(Strongest) 
    }
    for(Unit u) {
        u.train(Worker, Right, 7)
        u.train(Worker, Up, 15)

        for(Unit u) {
            if(u.canAttack()) then {
            u.build(Barracks, EnemyDir, 5) 
        }
            u.idle()
            u.harvest(4)
            u.attack(Closest)
            u.train(Heavy, EnemyDir, 3)
            u.moveToUnit(Ally, Strongest)
        }
    }
}
\end{verbatim}
\end{tcolorbox}
\begin{tcolorbox}[breakable, width=\textwidth+1cm]
\textbf{P9:}
\begin{verbatim}
for(Unit u){
    u.harvest(9)
    u.idle()
    u.moveToUnit(Enemy,LessHealthy)
    u.train(Worker,EnemyDir,6)
    for(Unit u){
        u.harvest(2)
    }
    for(Unit u){
        u.idle()
    }
    for(Unit u){
        if(u.HasUnitInOpponentRange()) then {
            u.attack(Weakest)
            u.moveToUnit(Enemy,LessHealthy)
        }
    }
    for(Unit u){
        u.moveToUnit(Enemy,MostHealthy)
        u.moveToUnit(Ally,MostHealthy)
        for(Unit u){
            u.build(Barracks,Left,7)
            u.train(Light,EnemyDir,20)
        }
    }
}
\end{verbatim}
\textbf{Explanation:\\}

\noindent
This strategy is geared towards aggressive resource gathering and expansion, while maintaining a robust defense. 
At the outset, our primary focus is on harvesting resources, with an emphasis on up to nine units engaging in this activity. As our units are engaged in resource gathering, they're also prepared to spring into action and defend our territory if required.

Simultaneously, we strategically move our units closer to the less healthy units of our enemy, targeting the weaker ones first to maximize our offensive capabilities. This isn't the highest priority, but a consistent task that we aim to achieve.

Along with these activities, we look to train six additional worker units, stationed towards the direction of the enemy. This serves dual purposes - enhancing our resource gathering capability and establishing a front line of defense against the enemy.

As we continue to build our resources and defenses, we deploy two units for more focused harvesting. This is more specific compared to the earlier broad directive and enjoys a higher priority.

This is closely followed by an order to have all our units maintain a vigilant stance, staying idle until there's a direct threat. This is a key element of our defense strategy.

A significant part of our strategy is to be proactively defensive. So, if any of our units are within the attacking range of the enemy, we go on the offensive by attacking the weakest units first, while strategically moving our units towards the less healthy enemy units. This has a higher priority and occurs before any other offensive or defensive moves.

The strategy also involves moving our units towards the most robust enemy and ally units. The aim here is to provide support to our strong units while simultaneously challenging the enemy's strong units.

To bolster our offense and defense, we aim to construct up to seven Barracks to the left of our existing base. Along with this, we train a massive force of twenty light combat units stationed towards the enemy. The building of the Barracks and the training of light units are part of our strategic expansion and are given a higher priority, falling just short of the highest priority.

Overall, this strategy is a blend of aggressive resource gathering, strategic defense, and tactical offense. Our goal is to build a strong resource base, maintain a defensive posture to thwart enemy attacks, and go on the offensive when the opportunity arises, aiming for a decisive victory.
\\\\

\textbf{Best Reconstructed Program for P9:}
\begin{verbatim}
for (Unit u) {
    u.harvest(9);    
    u.idle();
    u.moveToUnit(Enemy, LessHealthy);
    u.train(Worker, EnemyDir, 6);
    
    for (Unit u) {
        u.harvest(2);
        u.idle();
        if(u.hasUnitInOpponentRange()) {
            u.attack(Weakest);
        } else {
            u.moveToUnit(Enemy, LessHealthy);
        }
    }
    
    for (Unit u) {
        
        u.moveToUnit(Enemy, MostHealthy);        
        u.moveToUnit(Ally, MostHealthy);
        u.build(Barracks, Left, 7);
        u.train(Light, EnemyDir, 20);
    }
}
\end{verbatim}
\end{tcolorbox}

\begin{tcolorbox}[breakable, width=\textwidth+1cm]
\textbf{P10:}
\begin{verbatim}
for(Unit u){
    for(Unit u){
        u.train(Worker,Up,2)
    }
    u.idle()
    u.train(Heavy,EnemyDir,8)
}
for(Unit u){
    u.train(Light,Left,100)
    u.build(Barracks,EnemyDir,1)
    u.harvest(25)
    u.attack(Closest)
}
\end{verbatim}
\textbf{Explanation:\\}

\noindent
The provided strategy for playing MicroRTS consists of a series of prioritized actions. In this strategy, the most crucial action, with the highest priority, is to train worker units. Each of the controlled units is tasked with this function, effectively expanding the player's workforce as a first step. This action is prioritized by having the units train additional worker units in the upward direction, with the limit being up to 2 workers per unit. 
The second priority, which is slightly lower than the first one, is to ensure that the units stay idle when they are not training workers. In other words, until a unit has another specific task, it is required to stay idle, being ready to attack if any opponent comes within its attack range.

Next, there's the task of training heavy units. The units are given this task after they've finished training workers and become idle. Here, the strategy is to train up to 8 heavy units in the direction of the enemy, providing robust resistance against potential attacks.

Further down the priority list is the training of light units. Each unit is instructed to train light units to the left direction with a high limit of up to 100 units. This helps in preparing a strong army capable of handling various combat situations.

Building barracks in the enemy direction is also part of the strategy but with a lower priority than training units. Each unit is tasked to build up to one barracks in the direction of the enemy. This strategic placement could potentially disrupt the enemy's actions or serve as a forward base for our attacks.

Another crucial, yet lower priority task, is resource gathering. The strategy dictates that up to 25 workers are assigned to harvest resources. This provides the necessary resources to support the construction of buildings and training of additional units.

The last priority, with the lowest precedence, is to launch attacks on the enemy. Specifically, the units are ordered to attack the closest enemy unit. This offensive move is set with the lowest priority, implying that it should be executed only after all other tasks have been accomplished.

In summary, the strategy aims to expand and strengthen the player's forces through the training of different unit types, build a strategic forward base by erecting barracks near the enemy, manage resources by dedicating workers to harvesting, and finally, carry out attacks on the closest enemy units.\\\\

\textbf{Best Reconstructed Program for P10:}
\begin{verbatim}
for(Unit u) {
        u.train(Worker, Up, 2)
        u.idle()
        for(Unit u) {
            u.train(Heavy, EnemyDir, 8)
            u.train(Light, Left, 100)
            u.build(Barracks, EnemyDir, 1)
            u.harvest(25)
            u.attack(Closest)
        }
}
\end{verbatim}
\end{tcolorbox}

\begin{tcolorbox}[breakable, width=\textwidth+1cm]
\textbf{P11:}
\begin{verbatim}
for(Unit u){
    u.train(Worker,Up,4)
}
for(Unit u){
    u.idle()
}
for(Unit u){
    for(Unit u){
        u.harvest(1)
    }
    u.train(Worker,Down,6)
    for(Unit u){
        u.train(Heavy,Left,10)
        for(Unit u){
            u.harvest(3)
        }
        u.build(Barracks,Left,15)
        u.attack(Closest)
    }
}
for(Unit u){
    u.train(Light,Left,100)
}
\end{verbatim}
\textbf{Explanation:\\}

\noindent
The program dictates a strategy for playing MicroRTS that emphasizes unit training and resource gathering. The action priorities are as follows:
The highest priority action is to train Worker units at the top of your base. The game prioritizes the training of up to four Worker units. These units are crucial as they are responsible for gathering resources, building structures and even attacking opponent units.

Next in line, the strategy is to keep all units idle. This means that all units, including the previously trained Workers, will not be carrying out any specific task. They will only attack if an enemy unit comes within their attack range.

The next set of actions revolve around both training more Worker units and collecting resources. With a high priority, every unit in the game will be assigned to harvest resources, but only one resource per unit. Then, up to six additional Worker units will be trained at the bottom of your base.

Following this, the strategy takes a more aggressive turn. The priority then shifts to training up to ten Heavy units on the left of your base. Heavy units, as the name suggests, are crucial in attacking and defending against the opponent units due to their increased survivability and damage. Meanwhile, within this training routine, there is a sub-priority to have each of these units harvest three resources. After training these Heavy units and their brief stint in resource gathering, up to fifteen Barracks will be constructed to the left of your base. Barracks are important as they allow the training of more combat units. This phase concludes with each unit attacking the closest enemy unit, signifying the initiation of an offensive against the opponent.

Lastly, the game ends with training up to a whopping 100 Light units on the left of your base. Light units, while weaker than Heavy units, can still provide a strong force in numbers and this shows the game's shift to a more aggressive approach.

The overall goal of this strategy appears to be a balanced approach of building up the base with Worker units and Barracks, collecting resources, and eventually moving towards an aggressive offense using Heavy and Light units.\\\\

\textbf{Best Reconstructed Program for P11:}
\begin{verbatim}
for(Unit u) {
    u.train(Worker,Up,4)
    u.idle()
    for(Unit u) {
        u.harvest(1)
        u.train(Worker,Down,6)
    }
    for(Unit u) {
        u.train(Heavy,Left,10)
        u.harvest(3)
        u.build(Barracks,Left,15)
        u.attack(Closest)
    }
    for(Unit u) {
        u.train(Light,Left,100)
    }
}
\end{verbatim}
\end{tcolorbox}

\begin{tcolorbox}[breakable, width=\textwidth+1cm]
\textbf{P12:}
\begin{verbatim}
for(Unit u){
    u.train(Heavy,EnemyDir,6)
    u.train(Light,EnemyDir,4)
    u.build(Barracks,Down,3)
    u.idle()
    u.train(Worker,Left,3)
}
for(Unit u){
    for(Unit u){
        u.harvest(15)
    }
    for(Unit u){
        u.moveToUnit(Enemy,Weakest)
    }
}

\end{verbatim}
\textbf{Explanation:\\}

\noindent
In this MicroRTS strategy, the overall goal is to create a strong and sustainable army while harvesting resources efficiently. This strategy involves training and building activities along with harvesting and movement actions.
In the first stage of this strategy, every unit is given a specific task, which is crucial to the early growth and power of our side. The highest priority is given to training our Heavy units in the direction of the enemy, aiming to produce six of them. After this, the strategy prioritizes training four Light units, again in the direction of the enemy. These units will provide initial offensive capabilities to confront any immediate enemy threat.

The third priority action is to build three Barracks in the downward direction, creating a stronghold to prepare for future engagements. Next, our units are commanded to idle, indicating they should hold their position and attack any opponent units that enter their range. Finally, in the first stage, our strategy ensures sustainability by training three Worker units in the left direction. These workers will be vital for resource collection and base expansion.

In the second stage, the strategy focuses on resource gathering and unit movement. The topmost priority at this stage is for every unit to harvest resources until we have 15 resources. This sustains our army and ensures that we can continue to build and train units.

Subsequently, each unit is instructed to move towards the weakest enemy unit. This targeted approach allows our forces to systematically weaken the enemy's defense line by focusing on their most vulnerable points.

The execution of these tasks in the given order of priority ensures that our army is well-prepared and resourceful, which is crucial for victory in MicroRTS.\\\\

\textbf{Best Reconstructed Program for P12:}
\begin{verbatim}
for(Unit u) {
    u.train(Heavy, EnemyDir, 6)
    u.train(Light, EnemyDir, 4)
    u.build(Barracks, Down, 3)
    u.idle()
    u.train(Worker, Left, 3)
}

for(Unit u) {
    u.harvest(15)
    u.moveToUnit(Enemy, Weakest)
}
\end{verbatim}
\end{tcolorbox}

\begin{tcolorbox}[breakable, width=\textwidth+1cm]
\textbf{P13:}
\begin{verbatim}
for(Unit u){
    if(u.HasUnitWithinDistanceFromOpponent(5)) then {
        u.moveToUnit(Enemy,MostHealthy)
        u.moveToUnit(Ally,Farthest)
        u.attack(LessHealthy)
        u.train(Worker,Up,7)
    }
    u.attack(Weakest)
    for(Unit u){
        for(Unit u){
            u.harvest(1)
        }
        u.train(Light,Up,100)
        u.build(Barracks,EnemyDir,5)
        u.train(Worker,Left,4)
        u.idle()
        u.harvest(25)
    }
}
\end{verbatim}
\textbf{Explanation:\\}

\noindent
The first priority of the strategy is to check if any of our units are within 5 cells of an enemy unit. If this condition is met, the unit in question will carry out a set of actions. It will first move towards the healthiest enemy unit, indicating that our strategy is to target the strongest of the enemy. Immediately after that, the unit will also move towards our furthest ally unit. It is then commanded to attack the enemy unit with the least health. Following this, if the unit has the capability to train other units, it will train up to 7 workers in the upward direction. In case none of our units are within 5 cells of an enemy unit, the units will target the weakest enemy unit, regardless of its type or location.
The second priority in the strategy is to instruct each unit to perform a number of actions. This starts with each unit being instructed to harvest one resource unit. This may be viewed as a micro-management strategy, to ensure that resources are constantly being gathered. Following this, each unit is instructed to train up to a maximum of 100 light units in the upward direction. Then, they are tasked to build up to 5 Barracks towards the enemy direction. If possible, the units are also instructed to train up to 4 workers towards the left direction. If a unit is unable to perform any of these tasks, it is ordered to remain idle and will only engage in combat if an enemy unit comes within its attack range. In addition to this, each unit is also tasked to harvest up to 25 resource units if possible.

In summary, the strategy seems to prioritize offensive actions towards the enemy's healthiest units while at the same time reinforcing our own position by training new units and building new Barracks. The strategy also emphasizes resource gathering by setting specific harvest actions for each unit. Thus, the overall goal of this strategy is to wear down the enemy's strongest units while simultaneously growing our own forces and resource pool.\\\\

\textbf{Best Reconstructed Program for P13:}
\begin{verbatim}
for(Unit u) {
    if(u.hasUnitWithinDistanceFromOpponent(5)) {
        u.moveToUnit(Enemy, MostHealthy);
        u.moveToUnit(Ally, Farthest);
        u.attack(LessHealthy);
        if(u.hasNumberOfUnits(Base, 1)) {
            u.train(Worker, Down, 7);
        }
    }
    else {
        u.attack(Weakest);
    }

    for(Unit u) {
        if(u.canHarvest()) {
            u.harvest(1);
        }
        else {
            u.train(Light, Up, 100);
            u.build(Barracks, EnemyDir, 5);
        }
        if(u.hasNumberOfUnits(Base, 1)) {
            u.train(Worker, Left, 4);
            u.idle();
        }
        else {
            u.harvest(25);
        }
    }
}
\end{verbatim}
\end{tcolorbox}

\begin{tcolorbox}[breakable, width=\textwidth+1cm]
\textbf{P14:}
\begin{verbatim}
for(Unit u){
    u.idle()
    u.train(Light,EnemyDir,4)
    u.build(Barracks,Down,3)
    u.train(Ranged,EnemyDir,8)
    u.train(Worker,Left,3)
    u.harvest(15)
}
for(Unit u){
    for(Unit u){
        for(Unit u){
            u.attack(Closest)
        }
    }
}
\end{verbatim}
\textbf{Explanation:\\}

\noindent
In the strategy described, units initially take on an idle state, poised to attack any opponent unit that comes within range. This marks the first course of action. Following that, a total of four units are then trained for light combat, oriented towards the enemy's direction. This is the second action taken.
After preparing for combat, our focus shifts towards infrastructure development, with the construction of three barracks in the downward direction. This is the third action in line. Post the infrastructure setup, we further train eight ranged combat units, pointing again towards enemy territory, forming the fourth action.

In addition to training combat units, we also train three worker units, placed on the left, which makes up the fifth action. These worker units are immediately assigned to gather resources with a target of harvesting up to fifteen resource units, and this is the sixth action our strategy involves.

The strategy then shifts gears and goes into an aggressive mode. All units, regardless of their type, are assigned to attack the enemy's closest unit. This forms the most crucial and high-priority action in this plan, making it the top priority. This order is repeated for every single one of our units, ensuring a relentless wave of attacks aimed at the nearest enemy units.

Overall, this strategy involves starting with a defensive stance and resource accumulation, followed by a massive, prioritized attack on the closest enemy units, signifying an aggressive playstyle.\\\\

\textbf{Best Reconstructed Program for P14:}
\begin{verbatim}
for(Unit u) {
        u.idle();
        u.train(Light, EnemyDir, 4);
        u.build(Barracks, Down, 3);
        u.train(Ranged, EnemyDir, 8);
        u.train(Worker, Left, 3);
        u.harvest(15);
}
for(Unit u) {
    u.attack(Closest);
}
\end{verbatim}
\end{tcolorbox}

\begin{tcolorbox}[breakable, width=\textwidth+1cm]
\textbf{P15:}
\begin{verbatim}
for(Unit u){
    u.train(Worker,EnemyDir,3)
    u.build(Barracks,Up,1)
    u.moveToUnit(Enemy,Closest)
    u.attack(MostHealthy)
    u.moveToUnit(Ally,MostHealthy)
    for(Unit u){
        u.idle()
    }
    for(Unit u){
        u.train(Light,EnemyDir,15)
    }
    for(Unit u){
        u.harvest(2)
    }
}
\end{verbatim}
\textbf{Explanation:\\}

\noindent
In the strategy we're discussing, there's a clear order of priorities to the actions our units will take. 
At the very top of our priority list, the main thing units are encouraged to do is to stand their ground and defend their position. If you see a unit doing nothing, that's because they've been instructed to remain idle as their highest priority.

Following that, units have a higher priority to train Light combat units in the direction of the enemy. This means our main offensive strategy focuses on producing as many as 15 Light units to apply pressure on the opponent.

The next action with a somewhat lower priority than training Light units is resource gathering. Here, the directive is to send units out to harvest resources. The goal is to send up to 2 Workers to gather resources at any given time.

After those high-priority tasks, if none of those directives apply, our units then proceed with the next set of actions. The first among these is to train Workers, with an emphasis on training up to 3 of them in the direction of the enemy. This ensures that our army has a steady supply of units that can gather resources and build structures.

Once we have a sufficient number of Workers, the next objective is to build a Barracks in the upward direction. However, we limit this construction to just 1 Barracks. With a Barracks in place, we can then train combat units to bolster our army's strength.

With training and building out of the way, our units are then focused on movement and engagement strategies. They are instructed to approach the closest enemy unit and engage. Specifically, they will target the healthiest opponent unit, aiming to take down the strongest threats first.

In situations where our units need support, they are encouraged to move towards our most healthy ally unit. This could be a tactical move to bolster defense or provide a more united front in our offensive maneuvers.

Overall, the central aim of this strategy seems to be a balanced approach between defense, resource management, and offensive pressure. By prioritizing idle defense and the training of Light units, it suggests a reactive posture that is ready to quickly switch to offense by amassing Light units when the situation allows. The emphasis on resource gathering and Worker training ensures that the army is well-supported and can sustain prolonged engagements\\\\

\textbf{Best Reconstructed Program for P16:}
\begin{verbatim}
for(Unit u) {
    u.train(Worker, EnemyDir, 3)
    u.build(Barracks, Up, 1)
    u.moveToUnit(Enemy, Closest)
    u.attack(MostHealthy)
    u.moveToUnit(Ally, MostHealthy)
    
    for(Unit u) {
        u.idle()
    }
    
    for(Unit u) {
        u.train(Light, EnemyDir, 15)
        u.harvest(2)
    }
}
\end{verbatim}
\end{tcolorbox}

\begin{tcolorbox}[breakable, width=\textwidth+1cm]
\textbf{P16:}
\begin{verbatim}
for(Unit u){
    u.harvest(2)
    u.idle()
    u.train(Light,EnemyDir,7)
    u.build(Barracks,EnemyDir,3)
    u.train(Ranged,EnemyDir,8)
    u.train(Worker,Left,3)
    u.moveToUnit(Enemy,MostHealthy)
    u.moveAway()
}
for(Unit u){
    u.harvest(50)
    u.attack(Strongest)
    if(u.HasUnitWithinDistanceFromOpponent(3)) then {
        u.train(Worker,Right,10)
    }
}

\end{verbatim}
\textbf{Explanation:\\}

\noindent
In this MicroRTS strategy, our units prioritize their actions based on the following order:

1. First and foremost, they'll start by harvesting resources, but they will only send two units for this task.
2. If they aren't harvesting, they'll stay idle and defend if an opponent comes within their attack range.
3. If they haven't been assigned any of the above tasks, they'll then look to train Light combat units, specifically focusing on producing up to seven of them, directing them towards the enemy.
4. As they look to fortify their position, they'll then consider constructing Barracks. The units will try to build up to three Barracks in the direction of the enemy.
5. If their earlier training and building tasks are satisfied, they'll pivot to training Ranged units, aiming to have eight of these units, and they'll also be sent towards the enemy.
6. To ensure continuous resource collection and building, they'll train up to three Workers who will be positioned on the left side.
7. After all these tasks, if none of them are actionable for a unit, it will look for the healthiest opponent unit and move towards it.
8. Finally, if a unit hasn't been assigned any task till now, it'll choose to move away from the player's base.

Once all units have gone through the above priorities, they'll move to the next set of actions:

1. An intense resource collection activity will ensue, with a large group of 50 units being sent out to harvest.
2. When not busy harvesting, they'll go on the offensive, targeting the strongest opponent units to attack.
3. A special consideration is made for situations where they're close to the enemy. If a unit is within a distance of 3 from an opponent, it'll train up to ten Workers, positioning them on the right side.

The main goal of this strategy seems to be establishing a strong frontline with trained combat units and Barracks, while also ensuring there's a continuous supply of resources by having a large number of Workers. Moreover, it places a strong emphasis on being proactive in engagements by targeting stronger and healthier opponent units.\\\\

\textbf{Best Reconstructed Program for P16:}
\begin{verbatim}
for(Unit u) {
    for(Unit u) {
        if(u.canHarvest()) {
            u.harvest(2)
        }
        u.idle()
        u.train(Light, EnemyDir, 7)
        u.build(Barracks, EnemyDir, 3)
        u.train(Ranged, EnemyDir, 8)
        u.train(Worker, Left, 3)
        u.moveToUnit(Enemy, MostHealthy)
        u.moveAway()
    }
    for(Unit u) {
        u.harvest(50)
        u.attack(Strongest)
        if(u.hasUnitWithinDistanceFromOpponent(3)) {
            u.train(Worker, Right, 10)
        }
    }
}
\end{verbatim}
\end{tcolorbox}

\begin{tcolorbox}[breakable, width=\textwidth+1cm]
\textbf{P17:}
\begin{verbatim}
for(Unit u){
    for(Unit u){
        u.idle()
    }
    u.train(Worker,EnemyDir,5)
    u.harvest(3)
    u.train(Heavy,Right,50)
    u.moveToUnit(Enemy,MostHealthy)
    u.moveAway()
    if(u.OpponentHasUnitInPlayerRange()) then {
        for(Unit u){
            u.harvest(1)
            u.attack(Strongest)
        }
    }
    for(Unit u){
        u.build(Barracks,EnemyDir,1)
    }
}

\end{verbatim}
\textbf{Explanation:\\}

\noindent
The given strategy for playing MicroRTS presents a specific approach to handling the game, primarily focusing on certain aspects of controlling units and managing their actions.

The strategy commences with the highest priority action, where all units are commanded to stay idle and attack if an opponent's unit comes within attack range. This will form the core of the defense and is a priority to ensure that units are initially in a guarded state.

Next, the actions following that are of second-highest priority:
- The player's unit trains up to 5 Worker units in the direction of the enemy.
- 3 Worker units are sent to harvest resources.
- 50 Heavy combat units are trained to the right.
- Units are commanded to move towards the enemy's most healthy unit.
- Units are commanded to move in the opposite direction of the player's base.

These actions create a balance between building resources, strengthening the army, and engaging the enemy strategically.

The third-highest priority action is an immediate response mechanism. If an opponent's unit is within the attack range of an ally's unit, the player's units are commanded to:
- Harvest one resource each.
- Attack the strongest opponent unit.

This section forms a rapid reaction to an immediate threat, emphasizing attack and gaining resources in parallel.

Lastly, the action with the lowest priority is to build a Barracks in the direction of the enemy, limited to just one Barracks. This is the least urgent aspect of the strategy, focusing on long-term building rather than immediate response or unit training.

Overall, the strategy focuses on an immediate defensive stance with a systematic escalation in attacking, resource building, and strategic movement. The ultimate goal is to prepare the units quickly for both defense and attack, with a layered approach to handling threats, building resources, and then extending the infrastructure by constructing a Barracks. The priority is set in such a way that immediate defense and controlled aggression take precedence over the long-term development of structures.\\\\

\textbf{Best Reconstructed Program for P17:}
\begin{verbatim}
for(Unit u) {
    u.idle();
    
    for(Unit u) {
        u.train(Worker, EnemyDir, 5);
        u.harvest(3);
        u.train(Heavy, Right, 50);
        u.moveToUnit(Enemy, MostHealthy);
        u.moveAway();
    }
    
    for(Unit u) {
        if(u.opponentHasUnitInPlayerRange()) {
            u.harvest(1);
            u.attack(Strongest);
        }
    }
    
    u.build(Barracks, EnemyDir, 1);
}
\end{verbatim}
\end{tcolorbox}

\begin{tcolorbox}[breakable, width=\textwidth+1cm]
\textbf{P18:}
\begin{verbatim}
for(Unit u){
    u.idle()
    u.train(Light,Right,10)
    if(u.OpponentHasUnitInPlayerRange()) then {
        u.moveToUnit(Ally,Strongest)
    } else {
        u.train(Worker,EnemyDir,7)
    }
    u.build(Barracks,Down,20)
    u.moveToUnit(Enemy,Closest)
    for(Unit u){
        u.train(Worker,Right,5)
    }
    if(u.canAttack()) then {
        u.attack(Weakest)
    } else {
        e
    }
    for(Unit u){
        if(u.canHarvest()) then {
            u.harvest(2)
        }
        u.build(Barracks,EnemyDir,8)
    }
}
\end{verbatim}
\textbf{Explanation:\\}

\noindent
The given strategy for playing MicroRTS focuses on a balanced approach to both defense and offense, prioritizing certain actions to achieve desired outcomes.
First, the strategy commands all units to stay idle, which is of lower priority. If they are able to, they will then train 10 Light units to the right. The strategy then assesses the battlefield: if an opponent unit is in range of an ally unit, the priority is given to moving units towards the strongest Ally unit; otherwise, the command to train 7 Worker units towards the enemy direction is executed.

After assessing the battlefield, the priority shifts to building. Up to 20 Barracks are built downwards. This is followed by a command to move towards the closest enemy, which is of lower priority compared to previous actions.

Next, the strategy enters a phase where the priority is the highest, focusing on training and preparing for the battles ahead. In this phase, up to 5 Worker units are trained to the right.

Following this, the strategy again checks if the units can attack. If they can, they are commanded to attack the weakest opponent unit. If not, no action is taken.

In the final phase, which has the second-highest priority, the focus is on harvesting resources and further building. If units can harvest, they are instructed to harvest 2 resources. Additionally, up to 8 Barracks are built towards the enemy direction.

Overall, the strategy aims for a balanced approach. It prioritizes training and preparing for battle through building Barracks and training various types of units. It takes into account the positioning of both ally and enemy units, emphasizing moving and attacking based on the enemy's position and strength. It also focuses on resource management by training Workers and commanding them to harvest. The highest priority actions are training Workers, followed by building Barracks, attacking or defending based on the opponent's positioning, and finally, commanding units to move, stay idle, or train additional Light units.\\\\

\textbf{Best Reconstructed Program for P18:}
\begin{verbatim}
for (Unit u) {
    u.idle()
    if (u.canAttack()) {
        u.attack(Weakest)
    }
    u.build(Barracks, EnemyDir, 8)
    u.train(Worker, Right, 7)
    u.train(Light, Right, 10)
    if (u.opponentHasUnitInPlayerRange()) {
        u.moveToUnit(Ally, Strongest)
    } else {
        u.train(Worker, EnemyDir, 7)
    }
    u.build(Barracks, Down, 20)
    u.moveToUnit(Enemy, Closest)
    for (Unit u) {
        u.train(Worker, Right, 5)
        if (u.canHarvest()) {
            u.harvest(2)
        }
        u.build(Barracks, EnemyDir, 8)
    }
}
\end{verbatim}
\end{tcolorbox}

\begin{tcolorbox}[breakable, width=\textwidth+1cm]
\textbf{P19:}
\begin{verbatim}
for(Unit u){
    for(Unit u){
        u.idle()
        u.train(Light,Down,4)
    }
    for(Unit u){
        u.harvest(3)
        for(Unit u){
            u.train(Ranged,Down,20)
        }
    }
    u.build(Barracks,EnemyDir,50)
    u.attack(Closest)
    u.train(Worker,Down,8)
}
\end{verbatim}
\textbf{Explanation:\\}

\noindent
In this MicroRTS strategy, players will initially prioritize having their units stay idle and thereafter focus on training Light units. This action receives the highest priority. Once that's achieved, the player's next focal point is to ensure that up to 3 Worker units start harvesting resources. During this harvesting phase, there's also an emphasis on training up to 20 Ranged units. This step is of the second highest priority.

As the player progresses further, constructing Barracks in the direction of the enemy becomes a vital move. The game will aim to build up to 50 of these structures, suggesting a major strategic emphasis on increasing military capabilities. Following that, the units are directed to attack the closest enemy units. However, these actions aren't given as high a priority as the initial actions. 

Lastly, with a slightly lower priority, the strategy emphasizes training additional Worker units. The player will try to bring about 8 more Workers into the game.

To sum it up, the overall goal of this strategy seems to be a balanced approach between defense (having units stay idle initially) and offense (training Light and Ranged units). Furthermore, there's a strong emphasis on resource gathering and expanding the military capacity by building Barracks near the enemy. The balanced focus between offense, defense, and resource gathering suggests a comprehensive approach to the game.\\\\

\textbf{Best Reconstructed Program for P19:}
\begin{verbatim}
for(Unit u) {
    u.idle()

    for(Unit u) {
        u.train(Light, Down, 4)
    }

    for(Unit u) {
        u.harvest(3)
        u.train(Ranged, Down, 20)
    }

    for(Unit u) {
        u.build(Barracks, EnemyDir, 50)
        u.attack(Closest)
        u.train(Worker, Down, 8)
    }
}
\end{verbatim}
\end{tcolorbox}

\begin{tcolorbox}[breakable, width=\textwidth+1cm]
\textbf{P20:}
\begin{verbatim}
for(Unit u){
    for(Unit u){
        u.train(Heavy,EnemyDir,3)
    }
    for(Unit u){
        u.idle()
    }
    u.harvest(3)
    u.build(Base,EnemyDir,1)
    u.train(Light,Left,8)
    u.build(Barracks,Right,1)
    u.attack(Weakest)
    for(Unit u){
        u.train(Heavy,Up,10)
    }
    u.train(Worker,EnemyDir,4)
}
\end{verbatim}
\textbf{Explanation:\\}

\noindent
In this MicroRTS strategy, we initiate by placing the utmost importance on training Heavy units. The prime focus is to create three Heavy units positioned towards the enemy direction. This is the most crucial step of the strategy and receives the highest priority.

Following this, the units are instructed to maintain a stance of inaction, standing their ground. This takes precedence right after training the Heavy units.

As we move further down the list of priorities, the next action involves sending three Worker units to gather resources. Then, there's a directive to construct a Base, positioning it towards the enemy's direction. Only one Base will be built under this directive.

Subsequent to these actions, the strategy emphasizes creating a considerable force of Light units, specifically eight of them, positioned to the left. This is then followed by the establishment of a Barracks to the right, but only one will be erected.

The next phase of the strategy is to be more aggressive. Units are ordered to target and attack the weakest of the opponent's forces. This aggressive stance is vital before moving on to the subsequent steps.

Delving deeper, there's a strong inclination to further enhance our Heavy unit troops. This time, the strategy pushes for the creation of ten Heavy units, with their position being upwards.

Lastly, the strategy rounds off by emphasizing the training of four Worker units, positioning them towards the enemy's direction.

Overall, this strategy can be viewed as a balanced approach between fortifying defenses, resource management, and mounting an offensive. The primary goal is to strengthen the army by training a significant number of Heavy and Light units while ensuring resource accumulation and strategic positioning of bases and barracks.\\\\

\textbf{Best Reconstructed Program for P20:}
\begin{verbatim}
for (Unit u) {
    for (Unit u) {
        u.train(Heavy, EnemyDir, 3);
        u.idle();
    }
    u.harvest(3);
    u.build(Base, EnemyDir, 1);
    u.train(Light, Left, 8);
    u.build(Barracks, Right, 1);
    u.attack(Weakest);
    for (Unit u) {
        u.train(Heavy, Up, 10);
    }
    u.train(Worker, EnemyDir, 4);
}
\end{verbatim}
\end{tcolorbox}

\subsection{Human-crafted Set}
\begin{tcolorbox}[breakable, width=\textwidth+1cm]
\textbf{P1:}
\begin{verbatim}
for (Unit u : pgs.getUnits()) {
    if (u.getType() == baseType && u.getPlayer() == player &&\
    gs.getActionAssignment(u) == null) {
            int nworkers = 0;
            for (Unit u2 : pgs.getUnits()) {
                if (u2.getType() == workerType &&\
                u2.getPlayer() == p.getID()) {
                    nworkers++;
                }
            }
        if (nworkers < 1 && p.getResources() >= workerType.cost) {
            train(u, workerType);
        }
    }
}
\end{verbatim}
\textbf{Explanation:\\}

\noindent
In this strategy, the game checks for each unit on the game map. It focuses on the units that are of base type and belong to the player who doesn't have any assigned action. Once it finds such a base, it looks for all the worker units that belong to the player. It keeps a track of the number of such workers. If the player has less than one worker and has enough resources to train a worker, the base is commanded to train a worker unit. This strategy ensures that the player always has at least one worker unit if resources permit. The overall goal of this strategy is to maintain the number of worker units for the player, which is crucial for collecting resources and building structures.\\\\

\textbf{Best Reconstructed Program for P1:}
\begin{verbatim}
for(Unit u : pgs.getUnits()) {
    if (u.getType() == baseType && u.getPlayer() == player) {
        if (gs.getActionAssignment(u) == null) {
            int numWorkers = 0;
            for(Unit u2 : pgs.getUnits()) {
                if (u2.getType() == workerType &&\
                u2.getPlayer() == player) {
                    numWorkers++;
                }
            }
            if (numWorkers < 1 &&\\
            p.getResources() >= workerType.cost) {
                train(u, workerType);
            }
        }
    }
}
\end{verbatim}
\end{tcolorbox}

\begin{tcolorbox}[breakable, width=\textwidth+1cm]
\textbf{P2:}
\begin{verbatim}
for (Unit u : pgs.getUnits()) {
    if (u.getType() == barracksType && u.getPlayer() == player &&\
    gs.getActionAssignment(u) == null) {
        if (p.getResources() >= lightType.cost) {
            train(u, lightType);
        }
    }
}
\end{verbatim}
\textbf{Explanation:\\}

\noindent
This strategy focuses on training Light units from your Barracks. For every unit on the game map, the strategy checks if the unit is a Barracks and belongs to the player. The strategy then verifies if the Barracks is not currently assigned any action to perform. If the Barracks is idle and belongs to the player, the strategy looks at the amount of resources the player currently possesses. If the player has enough resources to train a Light unit (a Light unit's cost is the benchmark here), the strategy directs the Barracks to train a Light unit. The Light unit is a type of combat unit and can be useful in attacking the opponent's units. sThe overall goal of this strategy is to continuously train Light units from the Barracks whenever there are enough resources available and the Barracks is idle. This way, the player can build a strong army of Light units to take on the opponent.\\\\

\textbf{Best Reconstructed Program for P2:}
\begin{verbatim}
  List<Unit> units = pgs.getUnits();
  for(Unit u : units) {
    if(u.getType() == barracksType && u.getPlayer() == player) {
      if(gs.getActionAssignment(u) == null) {
        if(p.getResources() >= lightType.cost) {
          train(u, lightType);
        }
      }
    }
  }
\end{verbatim}
\end{tcolorbox}

\begin{tcolorbox}[breakable, width=\textwidth+1cm]
\textbf{P3:}
\begin{verbatim}
for (Unit u : pgs.getUnits()) {
    if (u.getType().canAttack && !u.getType().canHarvest &&\
    u.getPlayer() == player && gs.getActionAssignment(u) == null) {
        PhysicalGameState pgs = gs.getPhysicalGameState();
        Unit closestEnemy = null;
        int closestDistance = 0;
        int mybase = 0;
        for (Unit u2 : pgs.getUnits()) {
            if (u2.getPlayer() >= 0 && u2.getPlayer() != p.getID()) 
            {
                int d = Math.abs(u2.getX() - u.getX()) +\
                Math.abs(u2.getY() - u.getY());
                if (closestEnemy == null || d < closestDistance) {
                    closestEnemy = u2;
                    closestDistance = d;
                }
            }
            else if(u2.getPlayer()==p.getID() &&\
            u2.getType() == baseType){
                mybase = Math.abs(u2.getX() - u.getX()) +\
                Math.abs(u2.getY() - u.getY());
            }
        }
        if (closestEnemy!=null &&\
        (closestDistance < pgs.getHeight()/2  \
        || mybase < pgs.getHeight()/2)){
            attack(u,closestEnemy);
        }
        else
        {
            attack(u, null);
        }
    }
}\end{verbatim}
\textbf{Explanation:\\}

\noindent
In the given strategy, the player controls specific units that can attack but cannot gather resources. The strategy involves these units seeking out and attacking enemy units within a certain range. The first step is to identify all units that can attack but not harvest, and are controlled by the player. For each of these units, the strategy determines its closest enemy. This is done by calculating the distance between the player's unit and each enemy unit on the game map, with distance being the sum of the differences in their X and Y coordinates. Simultaneously, for each of these player's units, the strategy determines the distance to the player's base. This is done by identifying all units that belong to the player and are of the base type, and calculating the distance between the player's unit and the base. Once the closest enemy and the distance to the base are determined, the player's unit is commanded to attack the enemy if the enemy is closer than half the height of the game map or if the base is closer than half the height of the game map. If neither of these conditions is met, the unit is commanded to stop attacking. Overall, the goal of this strategy is to command units that can attack but cannot harvest to engage in combat with enemy units that are close by or pose a threat to the base. The strategy ensures that units are not sent too far from the base and are able to protect the base if it comes under attack.\\\\

\textbf{Best Reconstructed Program for P3:}
\begin{verbatim}
int halfHeight = pgs.getHeight() / 2;
List<Unit> units = pgs.getUnits();
for(Unit u : units) {
    if(u.getType().canAttack && !u.getType().canHarvest &&\
    u.getPlayer() == player) {
        Unit closestEnemy = null;
        Unit base = null;
        int closestDistance = Integer.MAX_VALUE;
        int baseDistance = Integer.MAX_VALUE;
        for(Unit u2 : units) {
            int dx = u2.getX() - u.getX();
            int dy = u2.getY() - u.getY();
            if(u2.getPlayer() >= 0 && u2.getPlayer() != player) {
                int distance = Math.abs(dx) + Math.abs(dy);
                if(distance < closestDistance) {
                    closestEnemy = u2;
                    closestDistance = distance;
                }
            }
            if(u2.getType() == baseType && u2.getPlayer() == player) 
            {
                baseDistance = Math.abs(dx) + Math.abs(dy);
                base = u2;
            }
        }
        if(closestEnemy != null && \\
        (closestDistance < halfHeight || baseDistance < halfHeight)) 
        {
            attack(u, closestEnemy);
        }
        else {
            attack(u, null);
        }
    }
}
\end{verbatim}
\end{tcolorbox}

\begin{tcolorbox}[breakable, width=\textwidth+1cm]
\textbf{P4:}
\begin{verbatim}
resourcesUsed=gs.getResourceUsage().getResourcesUsed(player); 
for (Unit u : pgs.getUnits()) {
    if (u.getType() == baseType && u.getPlayer() == player &&\
    gs.getActionAssignment(u) == null) {
        int nworkers = 0;
        for (Unit u2 : pgs.getUnits()) {
            if (u2.getType() == workerType
                    && u2.getPlayer() == p.getID()) {
                nworkers++;
            }
        }
        int nBases = 0;
        for (Unit u2 : pgs.getUnits()) {
            if (u2.getType() == baseType
                    && u2.getPlayer() == p.getID()) {
                nBases++;
            }
        }
        int qtdWorkLim = nWorkerBase * nBases;
        if (nworkers < qtdWorkLim &&\
        p.getResources() >= workerType.cost) {
            train(u, workerType);
        }
    }
}
\end{verbatim}
\textbf{Explanation:\\}

\noindent
The strategy in this program is about managing the bases and workers in the game. It first checks all units on the map. If it finds a base that belongs to the player and is not currently assigned any action, it starts to execute the strategy. The strategy begins by counting the number of workers that the player currently has on the map. Then, it counts the number of bases that the player owns. Next, it calculates the limit of workers that should be assigned to each base. This limit is determined by the product of a predefined constant (nWorkerBase) and the number of bases that the player owns. If the number of workers is less than this limit and the player has enough resources to train a new worker, the base is instructed to train a worker. The overall goal of this strategy is to maintain an optimal number of workers for each base that the player owns, given the resources available. This is done to ensure that the player can collect resources efficiently in the game.\\\\

\textbf{Best Reconstructed Program for P4:}
\begin{verbatim}
List<Unit> units = pgs.getUnits();
for(Unit u : units) {
    if(u.getType() == baseType && u.getPlayer() == player) {
        if(gs.getActionAssignment(u) == null) {
            int numWorkers = 0;
            for(Unit u2 : units) {
                if(u2.getType() == workerType &&\
                u2.getPlayer() == player) {
                    numWorkers++;
                }
            }
            int numBases = 0;
            for(Unit u2 : units) {
                if(u2.getType() == baseType &&\
                u2.getPlayer() == player) {
                    numBases++;
                }
            }
            int workerLimit = numBases * nWorkerBase;
            if(numWorkers < workerLimit &&\
            p.getResources() >= workerType.cost) {
                // Instruct base to train a worker
                train(u, workerType);
            }
        }
    }
}
\end{verbatim}
\end{tcolorbox}

\begin{tcolorbox}[breakable, width=\textwidth+1cm]
\textbf{P5:}
\begin{verbatim}
resourcesUsed=gs.getResourceUsage().getResourcesUsed(player); 
for (Unit u : pgs.getUnits()) {
    if (u.getType() == barracksType && u.getPlayer() == player &&\
    gs.getActionAssignment(u) == null) {
        int nLight = 0;
        int nRanged = 0;
        int nHeavy = 0;
        for (Unit u2 : pgs.getUnits()) {
            if (u2.getType() == lightType
                    && u.getPlayer() == p.getID()) {
                nLight++;
            }
            if (u2.getType() == rangedType
                    && u.getPlayer() == p.getID()) {
                nRanged++;
            }
            if (u2.getType() == heavyType
                    && u.getPlayer() == p.getID()) {
                nHeavy++;
            }
        }
        if (nLight == 0 &&\
        p.getResources() >= (lightType.cost + resourcesUsed)) {
            train(u, lightType);
            resourcesUsed += lightType.cost;
        } else if (nRanged == 0 &&\
        p.getResources() >= (rangedType.cost + resourcesUsed)) {
            train(u, rangedType);
            resourcesUsed += rangedType.cost;
        } else if (nHeavy == 0 &&\
        p.getResources() >= (heavyType.cost + resourcesUsed)) {
            train(u, heavyType);
            resourcesUsed += heavyType.cost;
        }
        if (p.getResources() >= baseType.cost &&\
        nLight != 0 && nRanged != 0 && nHeavy != 0) {
            int number = r.nextInt(3);
            switch (number) {
                case 0:
                    if (p.getResources() >=
                    (baseType.cost+lightType.cost)) {
                        train(u, lightType);
                        resourcesUsed += lightType.cost;
                    }
                    break;
                case 1:
                    if (p.getResources() >= 
                    (baseType.cost+rangedType.cost)) {
                        train(u, rangedType);
                        resourcesUsed += rangedType.cost;
                    }
                    break;
                case 2:
                    if (p.getResources() >= 
                    (baseType.cost+ heavyType.cost)) {
                        train(u, heavyType);
                        resourcesUsed += heavyType.cost;
                    }
                    break;
            }
        }
    }
}
\end{verbatim}
\textbf{Explanation:\\}

\noindent
The strategy described in the program involves managing the use of Barracks units and the resources of the player. The Barracks units are controlled by the player and are used to train combat units which are Light, Heavy, or Ranged. In the beginning, the program checks each unit within the player's control. If a unit is a Barracks unit, belongs to the player, and is not currently assigned an action, it proceeds to the next step. The strategy then counts the number of each type of combat unit (Light, Heavy, and Ranged) that the player currently has. It does this by going through each unit the player controls and checking if it's a Light, Heavy, or Ranged unit. Once the numbers of each type of combat unit are determined, the program checks if the player has any Light units. If not, and the player has enough resources to train a Light unit, it commands the Barracks to train a Light unit. If the player already has a Light unit, it checks for a Ranged unit and repeats the process. If the player has both Light and Ranged units, it checks for a Heavy unit and repeats the process again. The above process ensures that the player has at least one of each type of combat unit if they have enough resources. However, if the player already has all types of units and still has enough resources to train another unit plus build a base, it randomly selects one type of unit (Light, Heavy, or Ranged) to train. The overall goal of the strategy is to ensure the player has a balanced mix of combat units (Light, Heavy, and Ranged) and to use the available resources efficiently for training additional units or building a base.\\\\

\textbf{Best Reconstructed Program for P5:}
\begin{verbatim}
for(Unit u: pgs.getUnits()) {
    if (u.getType() == barracksType && u.getPlayer() == player &&\
    gs.getActionAssignment(u) == null) {        
        int numLight = 0;
        int numHeavy = 0;
        int numRanged = 0;
        for(Unit u2: pgs.getUnits()) {
            if (u2.getType() == lightType &&\
            u2.getPlayer() == player) {
                numLight++;
            } else if (u2.getType() == heavyType &&\
            u2.getPlayer() == player) {
                numHeavy++;
            } else if (u2.getType() == rangedType &&\
            u2.getPlayer() == player) {
                numRanged++;
            }
        }
        if (numLight == 0 &&\
        p.getResources() >= lightType.cost) {
            train(u, lightType);
        }
        else if (numRanged == 0 &&\
        p.getResources() >= rangedType.cost) {
            train(u, rangedType);
        }
        else if (numHeavy == 0 &&\
        p.getResources() >= heavyType.cost) {
            train(u, heavyType);
        }
        else if (p.getResources() >= lightType.cost + baseType.cost) 
        {
            int randomUnit = (int)(Math.random() * 3);
            if (randomUnit == 0) {
                train(u, lightType);
            } else if (randomUnit == 1) {
                train(u, heavyType);
            } else {
                train(u, rangedType);
            }
        }
    }
}
\end{verbatim}
\end{tcolorbox}

\begin{tcolorbox}[breakable, width=\textwidth+1cm]
\textbf{P6:}
\begin{verbatim}
for (Unit u : pgs.getUnits()) {
    if (u.getType().canAttack && !u.getType().canHarvest &&\
    u.getPlayer() == player && gs.getActionAssignment(u) == null) {
        PhysicalGameState pgs = gs.getPhysicalGameState();
        Unit closestEnemy = null;
        int closestDistance = 0;
        for (Unit u2 : pgs.getUnits()) {
            if (u2.getPlayer() >= 0 && u2.getPlayer() != p.getID()) 
            {
                int d = Math.abs(u2.getX() - u.getX()) +\
                Math.abs(u2.getY() - u.getY());
                if (closestEnemy == null || d < closestDistance) {
                    closestEnemy = u2;
                    closestDistance = d;
                }
            }
        }
        if (closestEnemy != null) {
            attack(u, closestEnemy);
        }
    }
}
\end{verbatim}
\textbf{Explanation:\\}

\noindent
This strategy is about using the combat units of the player for attacking the enemy units. Initially, the strategy looks at each unit in the game. It then considers those units which can attack, cannot harvest and belong to the player. Moreover, it only considers those units which are currently not assigned any other action. For each such combat unit, the strategy identifies the closest enemy unit. It calculates the distance between the combat unit and all units that belong to the opponent. The distance is calculated as the sum of the absolute differences of the x and y coordinates of the two units. The unit with the minimum distance is considered as the closest enemy. Once the closest enemy is found for a combat unit, it commands the combat unit to attack the closest enemy unit. The overall goal of the strategy is to engage the player's combat units in attacking the nearest enemy units, to potentially eliminate the enemy forces.\\\\

\textbf{Best Reconstructed Program for P6:}
\begin{verbatim}
for(Unit u : pgs.getUnits()) {
        if(u.getPlayer() == player && u.getType().canAttack &&\
        !u.getType().canHarvest) {
            if(gs.getActionAssignment(u) == null) {
                int minDistance = Integer.MAX_VALUE;
                Unit closestEnemy = null;
                for(Unit u2 : pgs.getUnits()) {
                    if(u2.getPlayer() >= 0 &&\
                    u2.getPlayer() != player) {
                        int dx = u2.getX() - u.getX();
                        int dy = u2.getY() - u.getY();
                        int distance = Math.abs(dx) + Math.abs(dy);
                        if(distance < minDistance) {
                            minDistance = distance;
                            closestEnemy = u2;
                        }
                    }
                }
                if (closestEnemy != null) {
                    attack(u, closestEnemy);
                }
            }
        }
    }
\end{verbatim}
\end{tcolorbox}

\begin{tcolorbox}[breakable, width=\textwidth+1cm]
\textbf{P7:}
\begin{verbatim}
List<Integer> reservedPositions = new LinkedList<>();
for (Unit u : pgs.getUnits()) {
    if (u.getType() == baseType && u.getPlayer() == player &&\
    gs.getActionAssignment(u) == null) {
        int nworkers = 0;
        for (Unit u2 : pgs.getUnits()) {
            if (u2.getType() == workerType
                    && u2.getPlayer() == p.getID()) {
                nworkers++;
            }
        }
        int nBases = 0;
        int nBarracks = 0;
        for (Unit u2 : pgs.getUnits()) {
            if (u2.getType() == baseType
                    && u2.getPlayer() == p.getID()) {
                nBases++;
            } else if (u2.getType() == barracksType
                    && u2.getPlayer() == p.getID()) {
                nBarracks++;
            }
        }
        int qtdWorkLim;
        if(nBarracks == 0){
            qtdWorkLim = 4;
        }else{
            qtdWorkLim = nWorkerBase * nBases;
        }
        if (nworkers < qtdWorkLim &&\
        p.getResources() >= workerType.cost) {
            train(u, workerType);
        }
    }
}
\end{verbatim}
\textbf{Explanation:\\}

\noindent
In this strategy, the player controls units in the MicroRTS game. Initially, the strategy checks if a base unit controlled by the player is not currently assigned any action. This is important because a unit can only perform one action at a time. Once it finds such a base, the strategy begins by counting the number of worker units controlled by the player. Workers are essential units that collect resources and build structures which are crucial for the advancement of the game. After counting the workers, the strategy proceeds to count the number of base and barracks units controlled by the player. Bases are where workers are trained, while barracks are where combat units are trained. Next, the strategy determines the limit of workers that can be trained. If there are no barracks, the limit is set to 4. However, if there are barracks, the limit is set to the number of workers per base times the number of bases. This decision is made based on the fact that having more barracks means the player is more likely to be in a combat-intensive situation and would require more resources, thus needing more workers. Finally, if the number of workers is less than the worker limit and the player has enough resources to train a worker (workers come at a cost), the base is commanded to train a worker. The overall goal of this strategy is to make sure that the player has an optimal number of workers based on the current state of the game. This is crucial because having an appropriate number of workers can ensure a steady flow of resources, which can contribute to the player's success in the game.\\\\

\textbf{Best Reconstructed Program for P7:}
\begin{verbatim}
List<Unit> units = gs.getUnits();
int workers = 0;
int base = 0;
int barracks = 0;
for(Unit u : units) {
    if(u.getPlayer() == player) {
        if(u.getType() == workerType) 
            workers++;
        else if(u.getType() == baseType &&\
        gs.getActionAssignment(u) == null) 
            base++;
        else if(u.getType() == barracksType) 
            barracks++;
    }
}
int workerLimit = (barracks == 0) ? 4 : workers * base;
for(Unit u : units) {
    if(u.getType() == baseType && u.getPlayer() == player &&\
    gs.getActionAssignment(u) == null && workers < workerLimit &&\
    p.getResources() >= workerType.cost) {
        train(u, workerType);
        break;
    }
}
\end{verbatim}
\end{tcolorbox}

\begin{tcolorbox}[breakable, width=\textwidth+1cm]
\textbf{P8:}
\begin{verbatim}
for (Unit u : pgs.getUnits()) {
    if (u.getType() == barracksType && u.getPlayer() == player &&\
    gs.getActionAssignment(u) == null) {
        int nLight = 0;
        int nRanged = 0;
        int nHeavy = 0;
        for (Unit u2 : pgs.getUnits()) {
            if (u2.getType() == lightType
                    && u.getPlayer() == p.getID()) {
                nLight++;
            }
            if (u2.getType() == rangedType
                    && u.getPlayer() == p.getID()) {
                nRanged++;
            }
            if (u2.getType() == heavyType
                    && u.getPlayer() == p.getID()) {
                nHeavy++;
            }
        }
        if (nLight == 0 && p.getResources() >= lightType.cost) {
            train(u, lightType);
        } else if (nRanged == 0 &&\
        p.getResources() >= rangedType.cost) {
            train(u, rangedType);

        } else if (nHeavy == 0 &&\
        p.getResources() >= heavyType.cost) {
            train(u, heavyType);
        }
        if (nLight != 0 && nRanged != 0 && nHeavy != 0) {
            int number = r.nextInt(3);
            switch (number) {
                case 0:
                    if (p.getResources() >= (lightType.cost)) {
                        train(u, lightType);
                    }
                    break;
                case 1:
                    if (p.getResources() >= (rangedType.cost)) {
                        train(u, rangedType);
                    }
                    break;
                case 2:
                    if (p.getResources() >= (heavyType.cost)) {
                        train(u, heavyType);
                    }
                    break;
            }
        }
    }
}
\end{verbatim}
\textbf{Explanation:\\}

\noindent
This strategy for playing MicroRTS focuses on training units at the barracks owned by the player. The process begins by checking each unit on the game map. If the unit is a barracks owned by the player and it's not currently assigned any action, the strategy proceeds to the next step. The strategy then counts the number of each type of combat units - Light, Ranged, and Heavy - currently owned by the player. Following this, the strategy tries to train one of each type of combat units, starting with the Light unit, then Ranged unit, and finally the Heavy unit. However, this is only done if the player has enough resources to train the unit and doesn't already own a unit of that type. In a situation where the player already owns at least one of each type of combat units, the strategy randomly chooses between the three types of units to train next. The strategy ensures that the player has enough resources to train the chosen unit type before proceeding with the training. The overall goal of this strategy is to ensure a balanced mix of combat units for the player, while also taking into consideration the amount of resources available to the player.\\\\

\textbf{Best Reconstructed Program for P8:}
\begin{verbatim}
   List<Unit> units = gs.getPhysicalGameState().getUnits();
    int numLight = 0, numRanged = 0, numHeavy = 0;
    for(Unit u : units) {
        if(u.getType() == barracksType && u.getPlayer() == player \
        && gs.getActionAssignment(u) == null) {
            for(Unit u2 : units) {
                if(u2.getPlayer() == player) {
                    if(u2.getType() == lightType) {
                        numLight++;
                    } else if(u2.getType() == rangedType) {
                        numRanged++;
                    } else if(u2.getType() == heavyType) {
                        numHeavy++;
                    }
                } 
            }            
            if(numLight == 0 && p.getResources() >= lightType.cost) 
            {
                train(u, lightType);
            } 
            else if(numRanged == 0 &&\
            p.getResources() >= rangedType.cost) {
                train(u, rangedType);
            } 
            else if(numHeavy == 0 &&\
            p.getResources() >= heavyType.cost) {
                train(u, heavyType);
            } 
            else {
                int r = new Random().nextInt(3);
                if(r == 0 && p.getResources() >= lightType.cost) {
                    train(u, lightType);
                } else if(r == 1 && \
                p.getResources() >= rangedType.cost) {
                    train(u, rangedType);
                } else if(r == 2 && \
                p.getResources() >= heavyType.cost) {
                    train(u, heavyType);
                }
            }
        }
    }
\end{verbatim}
\end{tcolorbox}

\begin{tcolorbox}[breakable, width=\textwidth+1cm]
\textbf{P9:}
\begin{verbatim}
for (Unit u : pgs.getUnits()) {
    if (u.getType().canAttack && !u.getType().canHarvest &&\
    u.getPlayer() == player && gs.getActionAssignment(u) == null) {
        Unit closestEnemy = null;
        int closestDistance = 0;
        for (Unit u2 : pgs.getUnits()) {
            if (u2.getPlayer() >= 0 && u2.getPlayer() != p.getID()) 
            {
                int d = Math.abs(u2.getX() - u.getX()) + \
                Math.abs(u2.getY() - u.getY());
                if (closestEnemy == null || d < closestDistance) {
                    closestEnemy = u2;
                    closestDistance = d;
                }
            }
        }
        if (closestEnemy != null) {
            attack(u, closestEnemy);
        }
    }
}
\end{verbatim}
\textbf{Explanation:\\}

\noindent
In this MicroRTS strategy, the focus is on using your combat units effectively to attack the enemy. The strategy begins by examining all the units on the map. It is particularly interested in the combat units that belong to your team and are not currently assigned any action. A combat unit is identified as a unit that can attack but cannot harvest resources. Upon identifying such a unit, the strategy shifts its focus to finding the closest enemy unit. This is achieved by examining the location of all units on the map that belong to the opposing team. The distance between your combat unit and each enemy unit is calculated using their x and y coordinates on the game map. The enemy unit with the smallest calculated distance is identified as the closest enemy. Once the closest enemy is identified, your combat unit is instructed to attack it. This sequence is repeated for all your available combat units that are not currently assigned any action. In this way, this strategy ensures that your combat units are always actively engaged in attacking the enemy and focuses on the enemy units that are closest to your units for quick engagement. The overall goal of this strategy is to maintain a high level of offensive pressure on the opponent by continuously engaging the closest enemy units with your available combat units. This approach keeps the opponent on the defensive and could potentially overwhelm them if their defenses are not strong enough.\\\\

\textbf{Best Reconstructed Program for P9:}
\begin{verbatim}
for(Unit u: pgs.getUnits()) {
    if(u.getPlayer() == player && u.getType().canAttack &&\
    !u.getType().canHarvest) {
        if(gs.getActionAssignment(u) == null) {
            Unit closestEnemy = null;
            int closestDistance = Integer.MAX_VALUE;
            for(Unit u2: pgs.getUnits()) {
                if(u2.getPlayer() >= 0 && u2.getPlayer() != player) 
                {
                    int dx = u2.getX() - u.getX();
                    int dy = u2.getY() - u.getY();
                    int distance = dx*dx + dy*dy;
                    if(closestEnemy == null || \
                    distance<closestDistance) {
                        closestEnemy = u2;
                        closestDistance = distance;
                    }
                }
            }
            if(closestEnemy != null) {
                attack(u, closestEnemy);
            }
        }
    }
}
\end{verbatim}
\end{tcolorbox}

\begin{tcolorbox}[breakable, width=\textwidth+1cm]
\textbf{P10:}
\begin{verbatim}
for(Unit u:pgs.getUnits()) {
    if (u.getType()==baseType && u.getPlayer() == player &&\
    gs.getActionAssignment(u)==null) {
            if (p.getResources()>=workerType.cost) {
                train(u, workerType);
            }
        }
    }
\end{verbatim}
\textbf{Explanation:\\}

\noindent
In this strategy, the game goes through each unit on the game map. It looks for any base units that belong to the player and is currently not performing any actions. Once such a base is found, the strategy checks the player's resources. If the player has sufficient resources to train a worker, the base is commanded to train a worker unit. The cost of training a worker is deducted from the player's resources. This strategy primarily focuses on constantly producing workers as long as there's enough resources and the base is idle. The overall goal of this strategy is to maximize the number of worker units the player has, as long as they can afford it.\\\\

\textbf{Best Reconstructed Program for P10:}
\begin{verbatim}
List<Unit> units = pgs.getUnits();
    for(Unit u : units) {
        if (u.getPlayer() == player) {
            if (u.getType() == baseType &&\
            gs.getActionAssignment(u) == null) {
                if (p.getResources() >= workerType.cost) {
                    train(u, workerType);
                }
            }
        }
    }
\end{verbatim}
\end{tcolorbox}